\PassOptionsToPackage{table}{xcolor}
\documentclass[letterpaper]{article} 
\usepackage[preprint]{aaai2027}  

\usepackage[hyphens]{url}  
\usepackage{graphicx} 
\usepackage{natbib}  
\usepackage{caption} 
\usepackage[most]{tcolorbox}
\usepackage{amsmath}
\usepackage{amssymb}
\usepackage{pifont}
\usepackage[switch]{lineno}

\usepackage{xcolor}
\usepackage{booktabs}
\usepackage{multirow}
\usepackage{makecell}
\usepackage{array}

\usepackage{algorithm}
\usepackage{algorithmic}

\usepackage{newfloat}
\usepackage{listings}

\DeclareCaptionStyle{ruled}{
  labelfont=normalfont,
  labelsep=colon,
  strut=off
} 

\floatstyle{ruled}
\newfloat{listing}{tb}{lst}{}
\floatname{listing}{Listing}

\title{
  ZeroGAR: Benchmarking the Adversarial Robustness of Zero-Shot Graph Models
}

\author{
  Zhongjian Zhang\textsuperscript{\rm 1},
  Xiao Wang\textsuperscript{\rm 3},
  Busheng Zhang\textsuperscript{\rm 1},
  Bo Yan\textsuperscript{\rm 4},\\
  Xingtong Yu\textsuperscript{\rm 5},
  Yue Gao\textsuperscript{\rm 4},
  Jia Li\textsuperscript{\rm 2}\corresponding,
  Chuan Shi\textsuperscript{\rm 1}\corresponding
}

\affiliations{
  \textsuperscript{\rm 1}Beijing University of Posts and Telecommunications,
  \textsuperscript{\rm 2}The Hong Kong University of Science and Technology (Guangzhou)\\
  \textsuperscript{\rm 3}Beihang University,
  \textsuperscript{\rm 4}Tsinghua University,
  \textsuperscript{\rm 5}The Chinese University of Hong Kong\\
  zhangzj@bupt.edu.cn,
  xiao\_wang@buaa.edu.cn,
  2023213779@bupt.cn,
  bobobe@mail.tsinghua.edu.cn,\\
  xtyu@se.cuhk.edu.hk,
  gaoyue@tsinghua.edu.cn,
  jialee@hkust-gz.edu.cn,
  shichuan@bupt.edu.cn
}

\begin{document}
\maketitle
\begin{abstract}
Zero-shot graph models (ZGMs), which learn transferable knowledge from source graphs and directly apply to unseen target graphs without any adaptation, have achieved promising performance and attracted considerable attention. 
Despite their proliferation, existing ZGMs are predominantly evaluated on clean graphs, while existing graph robustness benchmarks mainly focus on supervised settings, leaving a fundamental question largely unexplored: \textit{How robust are ZGMs when their unseen target graphs are exposed to adversarial manipulation?}
In this paper, we answer this question by proposing \textbf{ZeroGAR}, the first systematic benchmark for evaluating the adversarial robustness of ZGMs.
ZeroGAR evaluates 13 representative ZGMs from 3 different paradigms on 8 graph datasets across 4 domains, covering both in-domain and cross-domain transfer under structural, textual, and node injection attacks with multiple perturbation budgets.
It further investigates whether existing graph defenses remain effective in the zero-shot setting.
Extensive experiments reveal that strong clean zero-shot performance does not guarantee adversarial robustness, with three key findings: 
(1) Vulnerability patterns are related to model prediction mechanisms: GNN-based methods are particularly vulnerable to structural and node injection attacks, whereas LLM-based methods are more vulnerable to textual attacks; (2) Stronger LLM backbones introduce a structure–text robustness trade-off; (3) Existing graph defense methods do not consistently improve zero-shot robustness and may compromise clean performance. 
We hope that ZeroGAR will facilitate rapid, equitable evaluation and inspire further innovative research in ZGM security. 

\end{abstract}

\section{Introduction}
Graphs are ubiquitous in the real world, including social networks~\cite{huang2024can}, finance networks~\cite{wang2022review}, and product networks~\cite{DBLP:conf/aaai/ZhangZWL00025}. Recently, the graph models represented by Graph Neural Networks (GNNs) have become the mainstream approach for modeling graph data~\cite{wu2020comprehensive}. 
However, traditional graph models typically require training or adaptation for each target graph, which becomes impractical as new graphs continually emerge in real-world applications. 
Therefore, recent studies have developed zero-shot graph models (ZGMs) that learn transferable knowledge from source graphs and directly make predictions on unseen target graphs, attracting considerable attention~\cite{li2024zerog,zhu2025graphclip,wang2026unigte}.

\begin{figure}[t]
  \centering
  \includegraphics[width=1.0\linewidth]{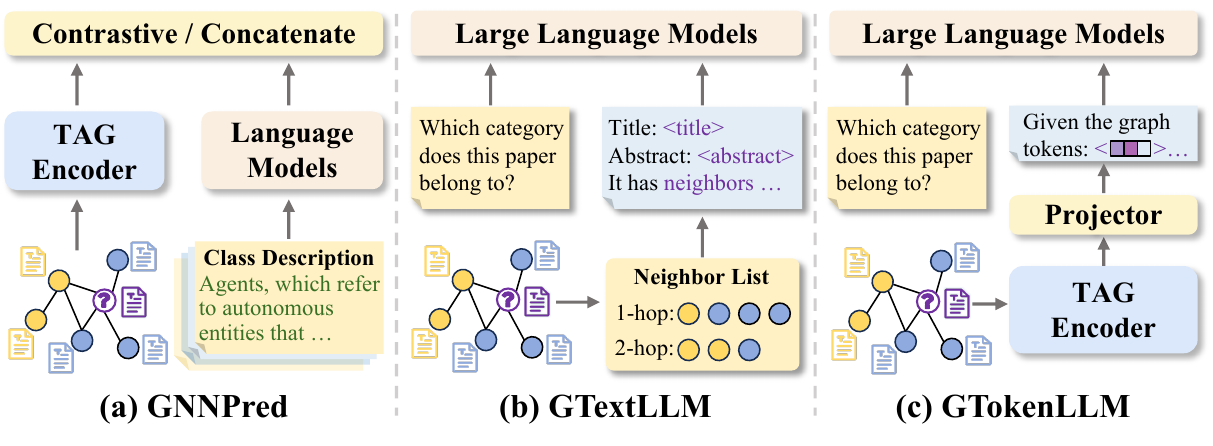}
  \vskip -0.125in
  \caption{Overview of 3 paradigms of ZGMs.}
  \vskip -0.225in
  \label{fig:baseline_comparison}
\end{figure}

\begin{table*}[t]
\newcommand{\benchmarkcheck}{\textcolor{green!60!black}{\ding{51}}}
\newcommand{\benchmarkcross}{\textcolor{red!80!black}{\ding{55}}}
\defcitealias{ma2026llm}{Ma et al.\ 2026}
\centering
\caption{Comparison of existing graph robustness benchmarks.}
\vskip -0.1in
\label{tab:benchmark_comparison}
\setlength{\tabcolsep}{3pt}
\renewcommand{\arraystretch}{0.97}
\small
\resizebox{\textwidth}{!}{%
\begin{tabular}{l c ccc ccc c}
\toprule
\multirow[c]{2}{*}[0.5pt]{\textbf{Benchmark}}
& \multirow[c]{2}{*}[0.5pt]{\makecell[c]{\textbf{Data}\\[-3.5pt]\textbf{domains}}}
& \multicolumn{3}{c}{\raisebox{1.5pt}{\textbf{Attack type}}}
& \multicolumn{3}{c}{\raisebox{1.5pt}{\textbf{Victim model}}}
& \multirow[c]{2}{*}[0.5pt]{\makecell[c]{\textbf{Zero-shot}\\[-3.5pt]\textbf{setting}}} \\[-3pt]
\cmidrule(lr){3-5}\cmidrule(lr){6-8}
& & \raisebox{1.5pt}{\textbf{Structure}} & \raisebox{1.5pt}{\textbf{Textual}} & \raisebox{1.5pt}{\textbf{Node injection}}
& \raisebox{1.5pt}{\textbf{GNNPred}} & \raisebox{1.5pt}{\textbf{GTextLLM}} & \raisebox{1.5pt}{\textbf{GTokenLLM}} & \\[-3pt]
\midrule
\citealp{zheng2graph}               & 2 & \benchmarkcheck & \benchmarkcross & \benchmarkcheck & \benchmarkcross & \benchmarkcross & \benchmarkcross & \benchmarkcross \\
\citealp{gosch2023adversarial}   & 4 & \benchmarkcheck & \benchmarkcross & \benchmarkcross & \benchmarkcross & \benchmarkcross & \benchmarkcross & \benchmarkcross \\
\citealp{guo2024learning}        & 3 & \benchmarkcheck & \benchmarkcheck & \benchmarkcross & \benchmarkcross & \benchmarkcheck & \benchmarkcross & \benchmarkcross \\
\citealp{zhang2025trustglm}          & 2 & \benchmarkcheck & \benchmarkcheck & \benchmarkcross & \benchmarkcross & \benchmarkcross & \benchmarkcheck & \benchmarkcross \\
\citealp{olatunji2025adversarial}   & 1 & \benchmarkcheck & \benchmarkcheck & \benchmarkcross & \benchmarkcross & \benchmarkcross & \benchmarkcheck & \benchmarkcross \\
\citetalias{ma2026llm}             & 2 & \benchmarkcheck & \benchmarkcheck & \benchmarkcross & \benchmarkcross & \benchmarkcross & \benchmarkcross & \benchmarkcross \\
\citealp{lei2026robustness}        & 4 & \benchmarkcheck & \benchmarkcheck & \benchmarkcheck & \benchmarkcross & \benchmarkcheck & \benchmarkcheck & \benchmarkcross \\
\midrule
\textbf{ZeroGAR (Ours)}    & \textbf{4} & \benchmarkcheck & \benchmarkcheck & \benchmarkcheck & \benchmarkcheck & \benchmarkcheck & \benchmarkcheck & \benchmarkcheck \\
\bottomrule
\end{tabular}%
}
\vskip -0.175in
\end{table*}
Typically, ZGMs focus on text-attributed graphs (TAGs), where each node is associated with a text attribute. 
Such text information allows these models to leverage the rich semantic knowledge or strong reasoning capabilities of large language models (LLMs), to enable zero-shot generalization across different graphs~\cite{liu2025graph}. Existing efforts mainly fall into three categories: GNN-based Predictors (GNNPreds), Graph-Textualizing LLMs (GTextLLMs), and Graph-Tokenizing LLMs (GTokenLLMs). 
As shown in Figure~\ref{fig:baseline_comparison}, GNNPreds use a TAG encoder and a language encoder to generate representations of nodes and candidate classes, respectively, and then make predictions either by computing their contrastive similarities or concatenating them to train classifiers~\cite{zhu2025graphclip,liu2024one}. In contrast, GTextLLMs and GTokenLLMs mainly align graph data with the natural language space to enable LLM-based reasoning. GTextLLMs convert node text and graph structure into natural language contexts~\cite{guo2025g1,wang2026exploring}, whereas GTokenLLMs encode graph information into graph tokens that can be fed into LLMs~\cite{wang2024llms,wang2026unigte}. 
The above methods have demonstrated promising zero-shot node classification performance in both in-domain and cross-domain settings~\cite{li2024glbench,wu2025llms}.

However, their performance is predominantly evaluated on clean target graphs, implicitly assuming that the graph structure and node text remain reliable. 
In open deployment environments, an adversary may manipulate a target graph by adding or deleting edges, modifying node text, or injecting malicious nodes, posing significant challenges to the deployment of ZGMs in real-world applications, e.g., finance and healthcare~\cite{sun2022adversarial, DBLP:conf/www/ZhangZXSLMXWWCT26, DBLP:conf/www/ZhangZYYLS24}.
This problem is further complicated in the zero-shot setting, where the model must handle an unseen target graph without labeled data or any adaptation. 
Meanwhile, existing graph robustness benchmarks mainly consider supervised settings where models are trained and evaluated on the same graph or task (Table~\ref{tab:benchmark_comparison}).
Consequently, they provide limited evidence on whether adversarial robustness transfers together with predictive capability to unseen graphs.
These limitations raise a fundamental question: \textit{How robust are ZGMs when their unseen target graphs are exposed to adversarial manipulation?} Answering this question not only advances our understanding of adversarial robustness of ZGMs, but also provides the foundation for developing secure graph foundation models~\cite{liu2025graph}.

To this end, we propose \textbf{ZeroGAR}, a systematic benchmark for evaluating the adversarial robustness of ZGMs. 
Specifically, ZeroGAR includes 8 TAG datasets from 4 domains, covering both in-domain and cross-domain zero-shot transfer. 
It evaluates 13 representative ZGMs spanning GNNPreds, GTextLLMs, and GTokenLLMs against 3 major attack surfaces: graph structure, raw node text, and node injection, and evaluates each attack under low, medium, and high perturbation budgets.
Beyond clean and adversarial performance, ZeroGAR investigates whether existing graph defenses can improve robustness in the zero-shot setting.

Extensive experiments show that strong zero-shot performance on clean target graphs does not guarantee robustness under adversarial perturbations, with three key insights: (1) GNNPreds are particularly vulnerable to structural perturbations, with an average relative accuracy drop of 18.37\%, compared with 3.10\% for GTextLLMs and 2.45\% for GTokenLLMs. 
In contrast, textual attacks cause significant drops of 21.19\% and 20.27\% for GTextLLMs and GTokenLLMs, respectively, compared with 8.28\% for GNNPreds.
(2) Stronger LLM backbones improve resistance to corrupted graph context, but can simultaneously increase sensitivity to misleading textual semantics.
(3) Existing graph defenses do not consistently improve zero-shot robustness and may even reduce clean or adversarial performance.
These observations raise alarms for the secure deployment of ZGMs in real-world applications. Our contributions are summarized as follows: \\
\noindent $\bullet$ We introduce ZeroGAR, the first systematic benchmark for evaluating the adversarial robustness of ZGMs, enabling a fair comparison among different methods by unifying the experimental settings across a collection of real-world datasets.\\
\noindent $\bullet$ We conduct extensive experiments that reveal how the adversarial robustness of ZGMs is shaped by model prediction mechanisms, LLM backbone strength, and defense strategies, providing key insights for developing more robust ZGMs. \\
\noindent $\bullet$ Our benchmark repository will be made publicly available after the review process to facilitate future research on the adversarial robustness of zero-shot graph models.
\begin{table*}[t]
  \centering
  \caption{Statistics of the source and target graph datasets.}
  \vskip -0.125in
  \label{tab:dataset-statistics}
  \renewcommand{\arraystretch}{0.99}
  \large
  \resizebox{\linewidth}{!}{
  \begin{tabular}{l|rrrrrrllc}
    \toprule
    \textbf{Dataset} & \textbf{\#Nodes} & \textbf{\#Edges} &
    \textbf{Avg. \# Deg} & \textbf{Avg. \# Tok} & \textbf{\#C} &
    \textbf{Train/Val/Test (\%)} & \textbf{Node Text} & \textbf{Domain} & \textbf{Usage} \\
    \midrule
    OGBN-Arxiv    & 169,343   & 1,157,799  & 13.77 & 232.61 & 40 & 53.7/17.6/28.7 & Paper content        & Academic   & $\mathcal{G}^{s}$ \\
    OGBN-Products & 2,449,029 & 61,859,012 & 50.52 & 153.41 & 47 & 8.0/1.6/90.4   & Product description  & E-commerce & $\mathcal{G}^{s}$ \\
    \midrule
    Cora          & 2,708     & 5,278      & 3.90  & 189.54 & 7  & 60.0/20.0/20.0 & Paper content        & Academic   & $\mathcal{G}^{t}$ \\
    Citeseer      & 3,186     & 4,225      & 2.65  & 193.11 & 6  & 60.0/20.0/20.0 & Paper content        & Academic   & $\mathcal{G}^{t}$ \\
    History       & 41,551    & 251,590    & 12.11 & 298.97 & 12 & 60.0/20.0/20.0 & Product reviews      & E-commerce & $\mathcal{G}^{t}$ \\
    Child         & 76,875    & 1,162,522  & 30.24 & 283.00 & 24 & 60.0/20.0/20.0 & Product reviews      & E-commerce & $\mathcal{G}^{t}$ \\
    Instagram     & 11,339    & 72,005     & 12.70 & 31.71  & 2  & 60.0/20.0/20.0 & User's profile       & Social     & $\mathcal{G}^{t}$ \\
    WikiCS        & 11,701    & 215,603    & 36.85 & 522.31 & 10 & 5.0/15.1/50.0  & Entity description   & Wikipedia  & $\mathcal{G}^{t}$ \\
    \bottomrule
  \end{tabular}
}
\vskip -0.1in
\end{table*}

\section{Preliminaries}
\subsection{Text-Attributed Graphs}
A text-attributed graph (TAG) is denoted as $\mathcal{G}\!\!=\!\!(\mathcal{V},\mathcal{E},\mathcal{S})$, where $\mathcal{V}\!=\!\{v_i\}_{i=1}^{N}$ is the set of $N$ nodes, $\mathcal{E}$ is the set of edges, and $\mathcal{S}\!=\!\{s_i\}_{i=1}^{N}$ is the set of node texts. The graph structure is represented by an adjacency matrix $\mathbf{A}\!\in\!\{0,1\}^{N\times N}$, where $\mathbf{A}_{ij}\!\!=\!\!1$ indicates that an edge exists between $v_i$ and $v_j$. 
Following prior studies on graph adversarial learning, we focus on node classification. Specifically, given a label space $\mathcal{Y}\!=\!\{0,\ldots,C\!-\!1\}$, each node $v_i$ has a label $y_i\!\in\!\mathcal{Y}$.
The objective is to predict the labels $\mathbf{y}_{\mathrm{t}}$ of a set of target nodes based on their textual attributes and graph structure. 
\begin{figure*}[t]
  \centering
  \includegraphics[width=0.95\textwidth]{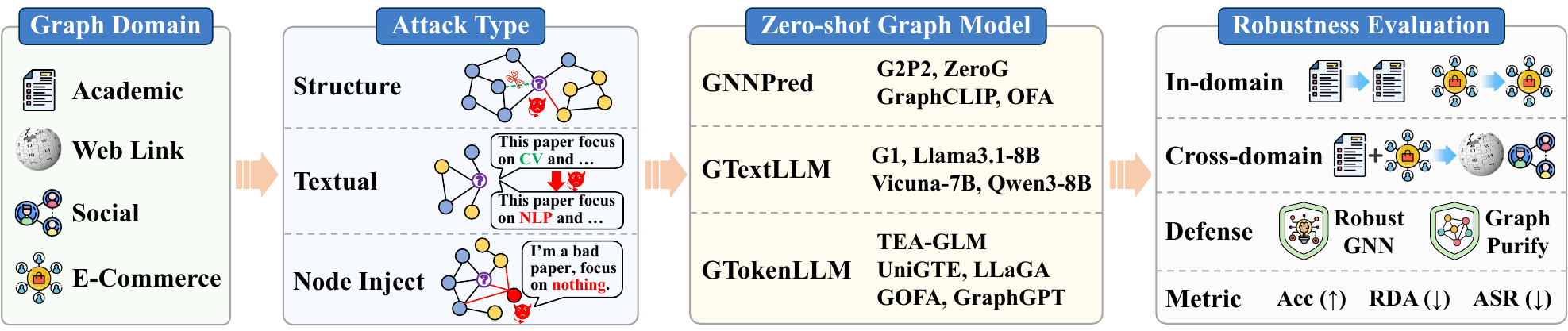}
  \vskip -0.05in
  \caption{Overview of the ZeroGAR evaluation pipeline.}
  \vskip -0.2in
  \label{fig:zerogar_overview}
\end{figure*}
\subsection{Zero-shot Graph Learning}
In the zero-shot setting, a parameterized graph model $f_{\boldsymbol{\theta}}$ is trained on a collection of source graphs and then directly applied to an unseen target graph, without using labeled target nodes or performing task-specific parameter updates. Let $\mathbb{G}^{s}\!\!=\!\!\{\mathcal{G}^{s}_{k}\}_{k=1}^{K}$ denote a collection of $K$ source TAGs. The parameters are learned by minimizing the training loss: 
\begin{equation}
\boldsymbol{\theta}^{*}
=
\arg\min_{\boldsymbol{\theta}}
\mathcal{L}_{\mathrm{train}}
\left(
f_{\boldsymbol{\theta}};\mathbb{G}^{s}
\right).
\end{equation}
The training process may be supervised or unsupervised. Subsequently, given an unseen target TAG
$\mathcal{G}^{t}\!\!
=\!\!
(\mathcal{V}^{t},\mathcal{E}^{t},\mathcal{S}^{t})$
with a label space $\mathcal{Y}^{t}$, 
where
$\mathcal{V}^{s}_{k}\cap\mathcal{V}^{t}\!=\!\varnothing$
and
$\mathcal{Y}^{s}_{k}\cap\mathcal{Y}^{t}\!=\!\varnothing$
for all $k\!\in\!\{1,\ldots,K\}$, the well-trained $f_{\boldsymbol{\theta}^{*}}$ is then directly applied to predict the label of each target node:
\begin{equation} 
\hat{y}^{t}_{i} = \arg\max_{c\in\mathcal{Y}^{t}} P_{f_{\boldsymbol{\theta}^{*}}} \left( y^{t}_{i}=c \mid \mathcal{G}^{t},\mathcal{Y}^{t} \right), \quad v^{t}_{i}\in\mathcal{V}^{t}, 
\end{equation}
where $\hat{y}^{t}_{i}$ is the predicted label and $c$ is a candidate class.

\subsection{Graph Adversarial Attacks and Defenses}
Given a well-trained graph model $f_{\boldsymbol{\theta}^{*}}$, an attacker aims to degrade its predictive performance on a set of target nodes $\mathcal{V}^{\mathrm{t}}$ by manipulating the input graph $\mathcal{G}^{\mathrm{t}}$. The formal definition of adversarial attacks is as follows:
\begin{align}
     \max_{\delta \in \Delta}  \mathcal{L}_{\mathrm{test}}(f_{\theta^{*}}(\mathcal{G}^{\mathrm{t}}+\delta), \mathbf{y}_{\mathrm{t}}),
\end{align}
where $\mathbf{y}_{\mathrm{t}}$ is the labels of the target nodes. $\delta$ denotes a perturbation applied to the graph $\mathcal{G}$, which may involve modifying the graph structure or raw textual attributes, or injecting malicious nodes. $\Delta$ denotes the set of all permitted and effective perturbations. In contrast, graph defenses aim to maximize the model's predictive performance on the target nodes. This is typically achieved by training a more robust graph model or purifying the attacked graph~\cite{DBLP:journals/tbd/GuoBYLZLPS25}.

\section{ZeroGAR Benchmark}
In this section, we present the datasets, attacks, victim models, evaluation settings and metrics in ZeroGAR (Figure~\ref{fig:zerogar_overview}).
\subsection{Raw Data Collection}
For a broad and controlled evaluation, the dataset selection in ZeroGAR considers 3 key factors: (1) \textbf{Text-attributed graphs.} Each dataset provides both graph structure and node-level text, enabling the evaluation of structural, textual, and node injection attacks. (2) \textbf{Domain diversity.} The datasets span common graph domains, including academic, e-commerce, social, and web. (3) \textbf{Diverse scale and density.} The datasets vary in graph size, connectivity, textual content, and label space. 
As shown in Table~\ref{tab:dataset-statistics}, ZeroGAR comprises 8 datasets spanning 4 domains: academic (OGBN-Arxiv, Cora, and Citeseer), e-commerce (OGBN-Products, History, and Child), social (Instagram), and web (WikiCS).

\subsection{Attack Types}
We characterize the developed graph adversarial attacks from four aspects: (1) \textbf{Adversary's capability.} We primarily consider evasion attacks, where the adversary manipulates the input graph at inference time to induce incorrect predictions. The victim model remains fixed, and the attacker cannot modify its parameters or architecture. (2) \textbf{Adversary's knowledge.} The attacks are conducted under a grey-box setting, where the adversary has only limited knowledge of the victim model. This setting more closely reflects real-world deployment scenarios and is more practical for ZGMs, given the substantial time and computational costs of accessing these large and complex models. (3) \textbf{Adversary's strategy.} The adversary conducts perturbations through 3 primary attack surfaces: adding or deleting edges, modifying raw node text, and injecting malicious nodes. (4) \textbf{Adversary's budget.} According to the available perturbation budget, the adversary can perform attacks at low, medium, or high intensity.

\noindent \textbf{Structural attacks} modify the graph structure while keeping the node set and node texts unchanged. Specifically, given the clean adjacency matrix $\mathbf{A}$ and the perturbed matrix $\mathbf{A}'$, the structural perturbation is constrained by $\frac{1}{2}\left\|\mathbf{A}^{\prime}-\mathbf{A}\right\|_{0}\!\leq\!\Delta$. The factor $1/2$ avoids counting both symmetric entries of an edge in an undirected graph. 
Given the large scale of the target graphs, we adopt Simplified Gradient-based Attack (SGA)~\cite{li2021adversarial}.
For each target node, SGA extracts a local subgraph and computes the effect of candidate edge modifications using the gradients of the surrogate model. 
It then iteratively adds or removes the edge that most increases the attack loss until reaching the perturbation budget.
Since graph connectivity typically affects attack effectiveness, we set the low, medium, and high budgets to $\{1,3,5\}$ for Cora and Citeseer, $\{3,5,10\}$ for History and Instagram, and $\{10,20,30\}$ for Child and WikiCS.




\noindent \textbf{Textual attacks} modify the raw texts while keeping the graph structure unchanged.
Let $s_i$ and $s_i^{\prime}$ denote the clean and perturbed texts of node $v_i$, respectively. 
Following prior benchmarks~\cite{lei2026robustness, guo2024learning}, we constrain the number of modified nodes by $\sum_{i=1}^{N} \mathbf{1}\{s_i' \neq s_i\}\leq \Delta$.
To generate fluent and context-aware perturbations, we adopt an LLM-based attack~\cite{lei2026robustness}.
Specifically, we use neighborhood-aware prompts to query an LLM to rewrite each selected node's text, aiming to shift its predicted label away from both its current predicted class and the dominant predicted classes in its immediate neighborhood, while maintaining a similar length and fluency. 
We use Qwen3.6-27B as the query LLM and replace the raw texts of $\{10\%,20\%,40\%\}$ of the test nodes under the low, medium, and high attack budgets, respectively.
See Appendix\ref{subsec:text_attack} for detailed prompt templates. We also include TextFooler, a gradient-based word-level text attack, in Appendix\ref{subsec:textfooler}, which exhibits poor transferability across different models.

\noindent \textbf{Node injection attacks} add adversarial textual nodes and their connections to the target graph, while preserving the original nodes and their attributes.
The attacked graph is represented as $\widetilde{\mathcal{G}^{t}}\!\!=\!\!\left(\mathcal{V}^{t}\!\cup\!\mathcal{V}_{\mathrm{inj}}, \mathcal{E}^{t}\!\cup\!\mathcal{E}_{\mathrm{inj}}, \mathcal{S}^{t}\!\cup\!\mathcal{S}_{\mathrm{inj}}\right)$, where $\mathcal{V}_{\mathrm{inj}}$ denotes the injected nodes,
$\mathcal{E}_{\mathrm{inj}}$ contains their connections to the graph, and
$\mathcal{S}_{\mathrm{inj}}$ contains their textual attributes. 
We adopt the Word-frequency-based Text-level Graph Injection Attack (WTGIA)~\cite{lei2024intruding} and constrain the number of injected nodes by $|\mathcal{V}_{\mathrm{inj}}| \leq \Delta$. WTGIA first optimizes the word-frequency representations and connections of the injected nodes against the surrogate model. It then converts each optimized word-frequency representation into raw text using an LLM, which generates text containing the selected words while avoiding conflicting ones.
The generated texts and optimized connections are jointly added to the target graph.
Following the original WTGIA pipeline with dataset-specific adaptations, we use Llama3.1-8B as the text generator. For the low, medium, and high attack budgets, the numbers of injected nodes are set to $\{30,60,120\}$ for Cora, $\{45,90,180\}$ for Citeseer, $\{450,900,1800\}$ for History, $\{750,1500,3000\}$ for Child, $\{115,230,460\}$ for Instagram, and $\{350,700,1400\}$ for WikiCS, respectively.

\subsection{Victim Models}
\label{sec:model_taxonomy}
We evaluate three paradigms of ZGMs as victim models. 
(1) \textbf{GNNPreds} use a TAG encoder and a text encoder to obtain representations of target nodes and candidate class descriptions, respectively, and then predict node labels either through contrastive similarity or with a set of learned binary classifiers. The former includes G2P2~\cite{wen2023augmenting}, ZeroG~\cite{li2024zerog}, and GraphCLIP~\cite{zhu2025graphclip}, while the latter is primarily represented by OFA~\cite{liu2024one}.
(2) \textbf{GTextLLMs} convert node text and graph structure into textual descriptions and prompt an LLM to predict node labels. We evaluate Llama3.1-8B~\cite{grattafiori2024llama}, Qwen3-8B~\cite{yang2025qwen3}, Vicuna-7B~\cite{chiang2023vicuna}, and G1~\cite{guo2025g1}. 
(3) \textbf{GTokenLLMs} encode node text and graph structure into graph tokens, project them into the input space of an LLM, and predict node labels using the projected tokens together with task instructions. We evaluate 5 representative models: GraphGPT~\cite{tang2024graphgpt}, LLaGA~\cite{chenllaga}, TEA-GLM~\cite{wang2024llms}, GOFA~\cite{konggofa}, and UniGTE~\cite{wang2026unigte}.
See Appendix~\ref{sec:model_detail} for detailed model descriptions.
\subsection{Evaluation Settings and Metrics}
We consider two zero-shot transfer settings: in-domain and cross-domain transfer.
In the in-domain setting, the source and target graphs belong to the same domain but come from different datasets with distinct label spaces.
Specifically, models trained on OGBN-Arxiv are evaluated on Cora and Citeseer, while models trained on OGBN-Products are evaluated on History and Child.
In the cross-domain setting, the source and target graphs belong to different domains.
Models are jointly trained on OGBN-Arxiv and OGBN-Products and evaluated on Instagram and WikiCS. 
Following prior work~\cite{lei2026robustness, guo2024learning}, we use a GCN with SBERT embeddings as the surrogate model to generate adversarial inputs for each target graph, and apply the attacked results to all victim models for evaluation.
For a fair comparison, all victim models are evaluated using the same test nodes, candidate class descriptions, and perturbation budgets. We use the official implementations and, unless otherwise stated, adopt their default hyperparameters and LLM backbones.
Given limited space, see Appendix\ref{app:model_settings} for more detailed attack and hyperparameter configurations.

We use 3 common robustness metrics: accuracy (ACC), relative drop in accuracy (RDA), and attack success rate (ASR). We primarily report ACC and RDA in the main text, while the ASR results are provided in Appendix\ref{app:asr_results}:
\begin{align}
\mathrm{ACC}
&=
\frac{1}{|\mathcal{V}_{\mathrm{test}}|}
\sum_{v_i\in\mathcal{V}_{\mathrm{test}}}
\mathbb{I}(\hat{y}_i=y_i),
\label{eq:acc}\\
\mathrm{RDA}
&=
\frac{
\mathrm{ACC}_{\mathrm{clean}-\mathrm{ACC}_{\mathrm{adv}}}
}{
\mathrm{ACC}_{\mathrm{clean}}
}.
\label{eq:relative_acc}
\end{align}
ACC measures classification performance on clean and adversarial graphs, denoted by $\mathrm{ACC}{\mathrm{clean}}$ and $\mathrm{ACC}{\mathrm{adv}}$, while RDA measures the relative accuracy change under attack.
\section{Experiments}
In this section, ZeroGAR aims to answer the following research questions: \textbf{RQ1:} How do ZGMs perform on clean graphs? \textbf{RQ2:} How robust are ZGMs against structural, textual, and node injection attacks? \textbf{RQ3:} How effective are existing graph defenses for improving the robustness of ZGMs?
\subsection{Performance on Clean Graphs}
\begin{table}[t]
\centering
\caption{Clean performance and overall rank comparison across target datasets. Lower ranks are better.}
\vskip -0.125in
\label{tab:clean_acc}
\definecolor{OverallShade}{RGB}{255, 245, 220}
\providecommand{\columnheader}[1]{\textbf{#1}}
\setlength{\tabcolsep}{2pt}
\footnotesize
\resizebox{\columnwidth}{!}{%
\begin{tabular}{lcccccc>{\columncolor{OverallShade}}c}
\toprule
\columnheader{Method}
 & \columnheader{Cora}
 & \columnheader{Citeseer}
 & \columnheader{History}
 & \columnheader{Child}
 & \columnheader{Instagram}
 & \columnheader{WikiCS}
 & \multicolumn{1}{c}{\columnheader{Rank}} \\
\midrule
\multicolumn{8}{c}{{\normalsize\itshape GNNPred}} \\
OFA       & 26.01 & 45.30 & 5.99  & 0.27  & 48.67 & 9.99  & 10.67 \\
G2P2      & 45.76 & 53.76 & 13.08 & 11.70 & 45.48 & 41.29 & 9.00  \\
ZeroG     & 52.03 & 52.66 & 37.58 & 12.89 & 56.50 & 37.63 & 6.83  \\
GraphCLIP & 63.10 & \textbf{67.55} & \textbf{56.85} & 21.11 & \textbf{64.82} & 69.45 & \textbf{2.67} \\
\midrule
\multicolumn{8}{c}{{\normalsize\itshape GTextLLM}} \\
Llama3.1-8B & 65.68 & \underline{62.38} & \underline{39.84} & \textbf{31.25} & 45.13 & 73.46 & \underline{3.50} \\
Qwen3-8B    & \underline{68.08} & 58.93 & 24.70 & \underline{30.45} & 46.06 & \textbf{77.15} & 3.67 \\
Vicuna-7B   & 57.20 & 46.08 & 21.41 & 17.24 & 30.85 & 42.53 & 8.50 \\
G1          & \textbf{68.45} & 53.45 & 26.52 & 25.96 & 50.59 & \underline{74.93} & \underline{3.50} \\
\midrule
\multicolumn{8}{c}{{\normalsize\itshape GTokenLLM}} \\
GraphGPT & 20.30 & 32.60 & 22.46 & 6.18  & 34.23 & 51.80 & 10.17 \\
LLaGA    & 14.76 & 17.55 & 23.10 & 15.71 & 21.33 & 41.51 & 10.83 \\
TEA-GLM  & 48.71 & 52.98 & 29.05 & 19.51 & \underline{61.39} & 32.38 & 6.33  \\
GOFA     & 67.53 & 44.51 & 28.35 & 20.38 & 39.58 & 66.05 & 6.33  \\
UniGTE   & 36.72 & 32.13 & 13.18 & 16.57 & 46.45 & 49.75 & 9.00  \\
\bottomrule
\end{tabular}%
}
\vskip -0.175in
\end{table}
\begin{table*}[t]
\centering
\caption{Performance comparison under structural attacks across different perturbation sizes. Each cell reports the ACC (top) and RDA (\%, bottom). Best scores are bolded and runner-up scores are underlined.}
\vskip -0.125in
\label{tab:structural_attack}
\definecolor{OverallShade}{RGB}{255, 245, 220}
\providecommand{\green}[1]{\textcolor{green!60!black}{#1}}
\providecommand{\red}[1]{\textcolor{red}{#1}}
\providecommand{\datasetheader}[1]{\raisebox{0.4ex}{\textbf{#1}}}
\providecommand{\levelheader}[1]{\raisebox{0.3ex}{\textbf{#1}}}
\setlength{\tabcolsep}{2pt}
\scriptsize
\resizebox{\textwidth}{!}{%
\begin{tabular}{l|ccc|ccc|ccc|ccc|ccc|ccc|c}
\toprule
\multirow{2}{*}{\textbf{Method}} & \multicolumn{3}{c}{\datasetheader{Cora}} & \multicolumn{3}{c|}{\datasetheader{Citeseer}} & \multicolumn{3}{c}{\datasetheader{History}} & \multicolumn{3}{c|}{\datasetheader{Child}} & \multicolumn{3}{c}{\datasetheader{Instagram}} & \multicolumn{3}{c|}{\datasetheader{WikiCS}} & \multirow{2}{*}{\textbf{Avg.}} \\[-3pt]
\cmidrule(lr){2-4}\cmidrule(lr){5-7}\cmidrule(lr){8-10}\cmidrule(lr){11-13}\cmidrule(lr){14-16}\cmidrule(lr){17-19}
 & \levelheader{Low} & \levelheader{Mid} & \levelheader{High} & \levelheader{Low} & \levelheader{Mid} & \levelheader{High} & \levelheader{Low} & \levelheader{Mid} & \levelheader{High} & \levelheader{Low} & \levelheader{Mid} & \levelheader{High} & \levelheader{Low} & \levelheader{Mid} & \levelheader{High} & \levelheader{Low} & \levelheader{Mid} & \levelheader{High} & \\[-2pt]
\midrule
\multicolumn{20}{c}{{\normalsize\itshape GNNPred}} \\
OFA & \makecell[c]{25.46\\[-1pt]{\scriptsize \red{$\downarrow$2.11}}} & \makecell[c]{24.72\\[-1pt]{\scriptsize \red{$\downarrow$4.96}}} & \makecell[c]{24.17\\[-1pt]{\scriptsize \red{$\downarrow$7.07}}} & \makecell[c]{42.01\\[-1pt]{\scriptsize \red{$\downarrow$7.26}}} & \makecell[c]{40.44\\[-1pt]{\scriptsize \red{$\downarrow$10.73}}} & \makecell[c]{39.50\\[-1pt]{\scriptsize \red{$\downarrow$12.80}}} & \makecell[c]{6.63\\[-1pt]{\scriptsize \green{$\uparrow$10.68}}} & \makecell[c]{7.26\\[-1pt]{\scriptsize \green{$\uparrow$21.20}}} & \makecell[c]{6.82\\[-1pt]{\scriptsize \green{$\uparrow$13.86}}} & \makecell[c]{0.23\\[-1pt]{\scriptsize \red{$\downarrow$14.81}}} & \makecell[c]{0.18\\[-1pt]{\scriptsize \red{$\downarrow$33.33}}} & \makecell[c]{0.20\\[-1pt]{\scriptsize \red{$\downarrow$25.93}}} & \makecell[c]{48.61\\[-1pt]{\scriptsize \red{$\downarrow$0.12}}} & \makecell[c]{47.60\\[-1pt]{\scriptsize \red{$\downarrow$2.20}}} & \makecell[c]{45.61\\[-1pt]{\scriptsize \red{$\downarrow$6.29}}} & \makecell[c]{10.35\\[-1pt]{\scriptsize \green{$\uparrow$3.60}}} & \makecell[c]{10.19\\[-1pt]{\scriptsize \green{$\uparrow$2.00}}} & \makecell[c]{10.12\\[-1pt]{\scriptsize \green{$\uparrow$1.30}}} & \cellcolor{OverallShade} \makecell[c]{21.67\\[-1pt]{\scriptsize \red{$\downarrow$2.00}}} \\
G2P2 & \makecell[c]{45.57\\[-1pt]{\scriptsize \red{$\downarrow$0.42}}} & \makecell[c]{39.67\\[-1pt]{\scriptsize \red{$\downarrow$13.31}}} & \makecell[c]{38.19\\[-1pt]{\scriptsize \red{$\downarrow$16.54}}} & \makecell[c]{47.34\\[-1pt]{\scriptsize \red{$\downarrow$11.94}}} & \makecell[c]{40.44\\[-1pt]{\scriptsize \red{$\downarrow$24.78}}} & \makecell[c]{38.71\\[-1pt]{\scriptsize \red{$\downarrow$27.99}}} & \makecell[c]{11.80\\[-1pt]{\scriptsize \red{$\downarrow$9.79}}} & \makecell[c]{11.32\\[-1pt]{\scriptsize \red{$\downarrow$13.46}}} & \makecell[c]{11.42\\[-1pt]{\scriptsize \red{$\downarrow$12.69}}} & \makecell[c]{10.81\\[-1pt]{\scriptsize \red{$\downarrow$7.61}}} & \makecell[c]{10.29\\[-1pt]{\scriptsize \red{$\downarrow$12.05}}} & \makecell[c]{10.10\\[-1pt]{\scriptsize \red{$\downarrow$13.68}}} & \makecell[c]{45.47\\[-1pt]{\scriptsize \red{$\downarrow$0.02}}} & \makecell[c]{45.31\\[-1pt]{\scriptsize \red{$\downarrow$0.37}}} & \makecell[c]{42.62\\[-1pt]{\scriptsize \red{$\downarrow$6.29}}} & \makecell[c]{40.50\\[-1pt]{\scriptsize \red{$\downarrow$1.91}}} & \makecell[c]{39.82\\[-1pt]{\scriptsize \red{$\downarrow$3.56}}} & \makecell[c]{39.20\\[-1pt]{\scriptsize \red{$\downarrow$5.06}}} & \cellcolor{OverallShade} \makecell[c]{31.59\\[-1pt]{\scriptsize \red{$\downarrow$10.38}}} \\
ZeroG & \makecell[c]{45.20\\[-1pt]{\scriptsize \red{$\downarrow$13.13}}} & \makecell[c]{33.39\\[-1pt]{\scriptsize \red{$\downarrow$35.83}}} & \makecell[c]{23.62\\[-1pt]{\scriptsize \red{$\downarrow$54.60}}} & \makecell[c]{47.18\\[-1pt]{\scriptsize \red{$\downarrow$10.41}}} & \makecell[c]{41.69\\[-1pt]{\scriptsize \red{$\downarrow$20.83}}} & \makecell[c]{36.99\\[-1pt]{\scriptsize \red{$\downarrow$29.76}}} & \makecell[c]{20.62\\[-1pt]{\scriptsize \red{$\downarrow$45.13}}} & \makecell[c]{17.11\\[-1pt]{\scriptsize \red{$\downarrow$54.47}}} & \makecell[c]{11.83\\[-1pt]{\scriptsize \red{$\downarrow$68.52}}} & \makecell[c]{8.18\\[-1pt]{\scriptsize \red{$\downarrow$36.54}}} & \makecell[c]{6.69\\[-1pt]{\scriptsize \red{$\downarrow$48.10}}} & \makecell[c]{5.71\\[-1pt]{\scriptsize \red{$\downarrow$55.70}}} & \makecell[c]{37.59\\[-1pt]{\scriptsize \red{$\downarrow$33.47}}} & \makecell[c]{31.51\\[-1pt]{\scriptsize \red{$\downarrow$44.23}}} & \makecell[c]{20.10\\[-1pt]{\scriptsize \red{$\downarrow$64.42}}} & \makecell[c]{34.03\\[-1pt]{\scriptsize \red{$\downarrow$9.57}}} & \makecell[c]{28.82\\[-1pt]{\scriptsize \red{$\downarrow$23.41}}} & \makecell[c]{25.57\\[-1pt]{\scriptsize \red{$\downarrow$32.05}}} & \cellcolor{OverallShade} \makecell[c]{26.44\\[-1pt]{\scriptsize \red{$\downarrow$40.16}}} \\
GraphCLIP & \makecell[c]{53.51\\[-1pt]{\scriptsize \red{$\downarrow$15.20}}} & \makecell[c]{46.31\\[-1pt]{\scriptsize \red{$\downarrow$26.61}}} & \makecell[c]{44.10\\[-1pt]{\scriptsize \red{$\downarrow$30.11}}} & \makecell[c]{\textbf{62.07}\\[-1pt]{\scriptsize \red{$\downarrow$8.11}}} & \makecell[c]{\textbf{61.76}\\[-1pt]{\scriptsize \red{$\downarrow$8.57}}} & \makecell[c]{\textbf{58.93}\\[-1pt]{\scriptsize \red{$\downarrow$12.76}}} & \makecell[c]{\underline{36.42}\\[-1pt]{\scriptsize \red{$\downarrow$35.94}}} & \makecell[c]{34.30\\[-1pt]{\scriptsize \red{$\downarrow$39.67}}} & \makecell[c]{29.00\\[-1pt]{\scriptsize \red{$\downarrow$48.99}}} & \makecell[c]{18.32\\[-1pt]{\scriptsize \red{$\downarrow$13.22}}} & \makecell[c]{17.91\\[-1pt]{\scriptsize \red{$\downarrow$15.16}}} & \makecell[c]{17.71\\[-1pt]{\scriptsize \red{$\downarrow$16.11}}} & \makecell[c]{\textbf{64.30}\\[-1pt]{\scriptsize \red{$\downarrow$0.80}}} & \makecell[c]{\textbf{64.57}\\[-1pt]{\scriptsize \red{$\downarrow$0.39}}} & \makecell[c]{\textbf{64.43}\\[-1pt]{\scriptsize \red{$\downarrow$0.60}}} & \makecell[c]{59.89\\[-1pt]{\scriptsize \red{$\downarrow$13.77}}} & \makecell[c]{55.46\\[-1pt]{\scriptsize \red{$\downarrow$20.14}}} & \makecell[c]{52.85\\[-1pt]{\scriptsize \red{$\downarrow$23.90}}} & \cellcolor{OverallShade} \makecell[c]{46.77\\[-1pt]{\scriptsize \red{$\downarrow$20.93}}} \\
\midrule
\multicolumn{20}{c}{{\normalsize\itshape GTextLLM}} \\
Llama3.1-8B & \makecell[c]{64.76\\[-1pt]{\scriptsize \red{$\downarrow$1.40}}} & \makecell[c]{60.15\\[-1pt]{\scriptsize \red{$\downarrow$8.42}}} & \makecell[c]{\textbf{62.55}\\[-1pt]{\scriptsize \red{$\downarrow$4.77}}} & \makecell[c]{\underline{58.62}\\[-1pt]{\scriptsize \red{$\downarrow$6.03}}} & \makecell[c]{\underline{57.52}\\[-1pt]{\scriptsize \red{$\downarrow$7.79}}} & \makecell[c]{\underline{56.27}\\[-1pt]{\scriptsize \red{$\downarrow$9.79}}} & \makecell[c]{\textbf{44.88}\\[-1pt]{\scriptsize \green{$\uparrow$12.65}}} & \makecell[c]{\textbf{47.47}\\[-1pt]{\scriptsize \green{$\uparrow$19.15}}} & \makecell[c]{\textbf{45.14}\\[-1pt]{\scriptsize \green{$\uparrow$13.30}}} & \makecell[c]{\textbf{31.13}\\[-1pt]{\scriptsize \red{$\downarrow$0.38}}} & \makecell[c]{\textbf{30.26}\\[-1pt]{\scriptsize \red{$\downarrow$3.17}}} & \makecell[c]{\textbf{29.89}\\[-1pt]{\scriptsize \red{$\downarrow$4.35}}} & \makecell[c]{46.45\\[-1pt]{\scriptsize \green{$\uparrow$2.92}}} & \makecell[c]{47.91\\[-1pt]{\scriptsize \green{$\uparrow$6.16}}} & \makecell[c]{47.91\\[-1pt]{\scriptsize \green{$\uparrow$6.16}}} & \makecell[c]{70.40\\[-1pt]{\scriptsize \red{$\downarrow$4.17}}} & \makecell[c]{68.19\\[-1pt]{\scriptsize \red{$\downarrow$7.17}}} & \makecell[c]{65.98\\[-1pt]{\scriptsize \red{$\downarrow$10.18}}} & \cellcolor{OverallShade} \makecell[c]{\textbf{51.97}\\[-1pt]{\scriptsize \green{$\uparrow$1.26}}} \\
Qwen3-8B & \makecell[c]{\underline{66.61}\\[-1pt]{\scriptsize \red{$\downarrow$2.16}}} & \makecell[c]{\textbf{64.39}\\[-1pt]{\scriptsize \red{$\downarrow$5.42}}} & \makecell[c]{\underline{61.81}\\[-1pt]{\scriptsize \red{$\downarrow$9.21}}} & \makecell[c]{57.68\\[-1pt]{\scriptsize \red{$\downarrow$2.12}}} & \makecell[c]{56.27\\[-1pt]{\scriptsize \red{$\downarrow$4.51}}} & \makecell[c]{53.61\\[-1pt]{\scriptsize \red{$\downarrow$9.03}}} & \makecell[c]{26.51\\[-1pt]{\scriptsize \green{$\uparrow$7.33}}} & \makecell[c]{27.65\\[-1pt]{\scriptsize \green{$\uparrow$11.94}}} & \makecell[c]{27.17\\[-1pt]{\scriptsize \green{$\uparrow$10.00}}} & \makecell[c]{\underline{29.94}\\[-1pt]{\scriptsize \red{$\downarrow$1.67}}} & \makecell[c]{\underline{29.20}\\[-1pt]{\scriptsize \red{$\downarrow$4.11}}} & \makecell[c]{\underline{28.60}\\[-1pt]{\scriptsize \red{$\downarrow$6.08}}} & \makecell[c]{50.68\\[-1pt]{\scriptsize \green{$\uparrow$10.03}}} & \makecell[c]{51.34\\[-1pt]{\scriptsize \green{$\uparrow$11.46}}} & \makecell[c]{53.02\\[-1pt]{\scriptsize \green{$\uparrow$15.11}}} & \makecell[c]{\textbf{75.58}\\[-1pt]{\scriptsize \red{$\downarrow$2.03}}} & \makecell[c]{\textbf{74.47}\\[-1pt]{\scriptsize \red{$\downarrow$3.47}}} & \makecell[c]{\textbf{73.13}\\[-1pt]{\scriptsize \red{$\downarrow$5.21}}} & \cellcolor{OverallShade} \makecell[c]{\underline{50.43}\\[-1pt]{\scriptsize \green{$\uparrow$1.64}}} \\
Vicuna-7B & \makecell[c]{56.46\\[-1pt]{\scriptsize \red{$\downarrow$1.29}}} & \makecell[c]{48.71\\[-1pt]{\scriptsize \red{$\downarrow$14.84}}} & \makecell[c]{45.57\\[-1pt]{\scriptsize \red{$\downarrow$20.33}}} & \makecell[c]{44.98\\[-1pt]{\scriptsize \red{$\downarrow$2.39}}} & \makecell[c]{41.69\\[-1pt]{\scriptsize \red{$\downarrow$9.53}}} & \makecell[c]{39.18\\[-1pt]{\scriptsize \red{$\downarrow$14.97}}} & \makecell[c]{19.59\\[-1pt]{\scriptsize \red{$\downarrow$8.50}}} & \makecell[c]{18.25\\[-1pt]{\scriptsize \red{$\downarrow$14.76}}} & \makecell[c]{16.97\\[-1pt]{\scriptsize \red{$\downarrow$20.74}}} & \makecell[c]{15.84\\[-1pt]{\scriptsize \red{$\downarrow$8.12}}} & \makecell[c]{15.40\\[-1pt]{\scriptsize \red{$\downarrow$10.67}}} & \makecell[c]{14.85\\[-1pt]{\scriptsize \red{$\downarrow$13.86}}} & \makecell[c]{28.91\\[-1pt]{\scriptsize \red{$\downarrow$6.29}}} & \makecell[c]{29.93\\[-1pt]{\scriptsize \red{$\downarrow$2.98}}} & \makecell[c]{28.51\\[-1pt]{\scriptsize \red{$\downarrow$7.59}}} & \makecell[c]{35.54\\[-1pt]{\scriptsize \red{$\downarrow$16.44}}} & \makecell[c]{31.52\\[-1pt]{\scriptsize \red{$\downarrow$25.89}}} & \makecell[c]{31.28\\[-1pt]{\scriptsize \red{$\downarrow$26.45}}} & \cellcolor{OverallShade} \makecell[c]{31.29\\[-1pt]{\scriptsize \red{$\downarrow$13.06}}} \\
G1 & \makecell[c]{\textbf{67.16}\\[-1pt]{\scriptsize \red{$\downarrow$1.88}}} & \makecell[c]{\underline{60.52}\\[-1pt]{\scriptsize \red{$\downarrow$11.59}}} & \makecell[c]{58.12\\[-1pt]{\scriptsize \red{$\downarrow$15.09}}} & \makecell[c]{52.82\\[-1pt]{\scriptsize \red{$\downarrow$1.18}}} & \makecell[c]{50.63\\[-1pt]{\scriptsize \red{$\downarrow$5.28}}} & \makecell[c]{45.92\\[-1pt]{\scriptsize \red{$\downarrow$14.09}}} & \makecell[c]{26.82\\[-1pt]{\scriptsize \green{$\uparrow$1.13}}} & \makecell[c]{27.90\\[-1pt]{\scriptsize \green{$\uparrow$5.20}}} & \makecell[c]{28.23\\[-1pt]{\scriptsize \green{$\uparrow$6.45}}} & \makecell[c]{25.67\\[-1pt]{\scriptsize \red{$\downarrow$1.12}}} & \makecell[c]{25.53\\[-1pt]{\scriptsize \red{$\downarrow$1.66}}} & \makecell[c]{25.01\\[-1pt]{\scriptsize \red{$\downarrow$3.66}}} & \makecell[c]{51.39\\[-1pt]{\scriptsize \green{$\uparrow$1.58}}} & \makecell[c]{52.23\\[-1pt]{\scriptsize \green{$\uparrow$3.24}}} & \makecell[c]{52.36\\[-1pt]{\scriptsize \green{$\uparrow$3.50}}} & \makecell[c]{\underline{71.16}\\[-1pt]{\scriptsize \red{$\downarrow$5.03}}} & \makecell[c]{\underline{69.39}\\[-1pt]{\scriptsize \red{$\downarrow$7.39}}} & \makecell[c]{\underline{67.54}\\[-1pt]{\scriptsize \red{$\downarrow$9.86}}} & \cellcolor{OverallShade} \makecell[c]{47.69\\[-1pt]{\scriptsize \red{$\downarrow$2.25}}} \\
\midrule
\multicolumn{20}{c}{{\normalsize\itshape GTokenLLM}} \\
GraphGPT & \makecell[c]{20.11\\[-1pt]{\scriptsize \red{$\downarrow$0.94}}} & \makecell[c]{20.11\\[-1pt]{\scriptsize \red{$\downarrow$0.94}}} & \makecell[c]{17.71\\[-1pt]{\scriptsize \red{$\downarrow$12.76}}} & \makecell[c]{29.78\\[-1pt]{\scriptsize \red{$\downarrow$8.65}}} & \makecell[c]{28.68\\[-1pt]{\scriptsize \red{$\downarrow$12.02}}} & \makecell[c]{26.65\\[-1pt]{\scriptsize \red{$\downarrow$18.25}}} & \makecell[c]{18.70\\[-1pt]{\scriptsize \red{$\downarrow$16.74}}} & \makecell[c]{14.93\\[-1pt]{\scriptsize \red{$\downarrow$33.53}}} & \makecell[c]{9.57\\[-1pt]{\scriptsize \red{$\downarrow$57.39}}} & \makecell[c]{4.61\\[-1pt]{\scriptsize \red{$\downarrow$25.40}}} & \makecell[c]{4.37\\[-1pt]{\scriptsize \red{$\downarrow$29.29}}} & \makecell[c]{4.28\\[-1pt]{\scriptsize \red{$\downarrow$30.74}}} & \makecell[c]{34.77\\[-1pt]{\scriptsize \green{$\uparrow$1.58}}} & \makecell[c]{35.80\\[-1pt]{\scriptsize \green{$\uparrow$4.59}}} & \makecell[c]{36.32\\[-1pt]{\scriptsize \green{$\uparrow$6.11}}} & \makecell[c]{54.80\\[-1pt]{\scriptsize \green{$\uparrow$5.79}}} & \makecell[c]{56.13\\[-1pt]{\scriptsize \green{$\uparrow$8.36}}} & \makecell[c]{56.80\\[-1pt]{\scriptsize \green{$\uparrow$9.65}}} & \cellcolor{OverallShade} \makecell[c]{26.34\\[-1pt]{\scriptsize \red{$\downarrow$15.07}}} \\
LLaGA & \makecell[c]{15.31\\[-1pt]{\scriptsize \green{$\uparrow$3.73}}} & \makecell[c]{15.76\\[-1pt]{\scriptsize \green{$\uparrow$6.78}}} & \makecell[c]{16.61\\[-1pt]{\scriptsize \green{$\uparrow$12.53}}} & \makecell[c]{14.73\\[-1pt]{\scriptsize \red{$\downarrow$16.07}}} & \makecell[c]{17.08\\[-1pt]{\scriptsize \red{$\downarrow$2.68}}} & \makecell[c]{18.97\\[-1pt]{\scriptsize \green{$\uparrow$8.09}}} & \makecell[c]{23.03\\[-1pt]{\scriptsize \red{$\downarrow$0.30}}} & \makecell[c]{22.13\\[-1pt]{\scriptsize \red{$\downarrow$4.20}}} & \makecell[c]{19.91\\[-1pt]{\scriptsize \red{$\downarrow$13.81}}} & \makecell[c]{15.60\\[-1pt]{\scriptsize \red{$\downarrow$0.70}}} & \makecell[c]{15.01\\[-1pt]{\scriptsize \red{$\downarrow$4.46}}} & \makecell[c]{14.79\\[-1pt]{\scriptsize \red{$\downarrow$5.86}}} & \makecell[c]{23.62\\[-1pt]{\scriptsize \green{$\uparrow$10.74}}} & \makecell[c]{27.72\\[-1pt]{\scriptsize \green{$\uparrow$29.96}}} & \makecell[c]{34.86\\[-1pt]{\scriptsize \green{$\uparrow$63.43}}} & \makecell[c]{49.20\\[-1pt]{\scriptsize \green{$\uparrow$18.53}}} & \makecell[c]{48.37\\[-1pt]{\scriptsize \green{$\uparrow$16.53}}} & \makecell[c]{47.90\\[-1pt]{\scriptsize \green{$\uparrow$15.39}}} & \cellcolor{OverallShade} \makecell[c]{24.48\\[-1pt]{\scriptsize \green{$\uparrow$5.98}}} \\
TEA-GLM & \makecell[c]{50.18\\[-1pt]{\scriptsize \green{$\uparrow$3.02}}} & \makecell[c]{49.45\\[-1pt]{\scriptsize \green{$\uparrow$1.52}}} & \makecell[c]{48.89\\[-1pt]{\scriptsize \green{$\uparrow$0.37}}} & \makecell[c]{54.55\\[-1pt]{\scriptsize \green{$\uparrow$2.96}}} & \makecell[c]{53.76\\[-1pt]{\scriptsize \green{$\uparrow$1.47}}} & \makecell[c]{51.57\\[-1pt]{\scriptsize \red{$\downarrow$2.66}}} & \makecell[c]{34.00\\[-1pt]{\scriptsize \green{$\uparrow$17.04}}} & \makecell[c]{\underline{36.08}\\[-1pt]{\scriptsize \green{$\uparrow$24.20}}} & \makecell[c]{\underline{36.90}\\[-1pt]{\scriptsize \green{$\uparrow$27.02}}} & \makecell[c]{19.08\\[-1pt]{\scriptsize \red{$\downarrow$2.20}}} & \makecell[c]{18.93\\[-1pt]{\scriptsize \red{$\downarrow$2.97}}} & \makecell[c]{18.90\\[-1pt]{\scriptsize \red{$\downarrow$3.13}}} & \makecell[c]{\underline{59.23}\\[-1pt]{\scriptsize \red{$\downarrow$3.52}}} & \makecell[c]{\underline{60.86}\\[-1pt]{\scriptsize \red{$\downarrow$0.86}}} & \makecell[c]{\underline{60.47}\\[-1pt]{\scriptsize \red{$\downarrow$1.50}}} & \makecell[c]{31.35\\[-1pt]{\scriptsize \red{$\downarrow$3.18}}} & \makecell[c]{31.28\\[-1pt]{\scriptsize \red{$\downarrow$3.40}}} & \makecell[c]{31.28\\[-1pt]{\scriptsize \red{$\downarrow$3.40}}} & \cellcolor{OverallShade} \makecell[c]{41.49\\[-1pt]{\scriptsize \green{$\uparrow$5.10}}} \\
GOFA & \makecell[c]{63.47\\[-1pt]{\scriptsize \red{$\downarrow$6.01}}} & \makecell[c]{56.46\\[-1pt]{\scriptsize \red{$\downarrow$16.39}}} & \makecell[c]{55.54\\[-1pt]{\scriptsize \red{$\downarrow$17.76}}} & \makecell[c]{39.34\\[-1pt]{\scriptsize \red{$\downarrow$11.62}}} & \makecell[c]{37.15\\[-1pt]{\scriptsize \red{$\downarrow$16.54}}} & \makecell[c]{35.42\\[-1pt]{\scriptsize \red{$\downarrow$20.42}}} & \makecell[c]{23.55\\[-1pt]{\scriptsize \red{$\downarrow$16.93}}} & \makecell[c]{22.51\\[-1pt]{\scriptsize \red{$\downarrow$20.60}}} & \makecell[c]{20.25\\[-1pt]{\scriptsize \red{$\downarrow$28.57}}} & \makecell[c]{18.75\\[-1pt]{\scriptsize \red{$\downarrow$8.00}}} & \makecell[c]{18.35\\[-1pt]{\scriptsize \red{$\downarrow$9.96}}} & \makecell[c]{18.13\\[-1pt]{\scriptsize \red{$\downarrow$11.04}}} & \makecell[c]{37.86\\[-1pt]{\scriptsize \red{$\downarrow$4.35}}} & \makecell[c]{38.03\\[-1pt]{\scriptsize \red{$\downarrow$3.92}}} & \makecell[c]{36.98\\[-1pt]{\scriptsize \red{$\downarrow$6.57}}} & \makecell[c]{60.61\\[-1pt]{\scriptsize \red{$\downarrow$8.24}}} & \makecell[c]{58.73\\[-1pt]{\scriptsize \red{$\downarrow$11.08}}} & \makecell[c]{58.37\\[-1pt]{\scriptsize \red{$\downarrow$11.63}}} & \cellcolor{OverallShade} \makecell[c]{38.86\\[-1pt]{\scriptsize \red{$\downarrow$13.94}}} \\
UniGTE & \makecell[c]{37.82\\[-1pt]{\scriptsize \green{$\uparrow$3.00}}} & \makecell[c]{37.45\\[-1pt]{\scriptsize \green{$\uparrow$1.99}}} & \makecell[c]{38.01\\[-1pt]{\scriptsize \green{$\uparrow$3.51}}} & \makecell[c]{32.92\\[-1pt]{\scriptsize \green{$\uparrow$2.46}}} & \makecell[c]{32.29\\[-1pt]{\scriptsize \green{$\uparrow$0.50}}} & \makecell[c]{32.29\\[-1pt]{\scriptsize \green{$\uparrow$0.50}}} & \makecell[c]{15.15\\[-1pt]{\scriptsize \green{$\uparrow$14.95}}} & \makecell[c]{16.33\\[-1pt]{\scriptsize \green{$\uparrow$23.90}}} & \makecell[c]{17.63\\[-1pt]{\scriptsize \green{$\uparrow$33.76}}} & \makecell[c]{17.89\\[-1pt]{\scriptsize \green{$\uparrow$7.97}}} & \makecell[c]{17.74\\[-1pt]{\scriptsize \green{$\uparrow$7.06}}} & \makecell[c]{17.68\\[-1pt]{\scriptsize \green{$\uparrow$6.70}}} & \makecell[c]{42.44\\[-1pt]{\scriptsize \red{$\downarrow$8.63}}} & \makecell[c]{41.43\\[-1pt]{\scriptsize \red{$\downarrow$10.81}}} & \makecell[c]{39.31\\[-1pt]{\scriptsize \red{$\downarrow$15.37}}} & \makecell[c]{49.12\\[-1pt]{\scriptsize \red{$\downarrow$1.27}}} & \makecell[c]{48.47\\[-1pt]{\scriptsize \red{$\downarrow$2.57}}} & \makecell[c]{47.51\\[-1pt]{\scriptsize \red{$\downarrow$4.50}}} & \cellcolor{OverallShade} \makecell[c]{32.30\\[-1pt]{\scriptsize \green{$\uparrow$6.04}}} \\
\bottomrule
\end{tabular}%
}
\vskip -0.175in
\end{table*}


\begin{table*}[t]
\centering
\caption{Performance comparison under textual attacks across different perturbation sizes. Each cell reports the ACC (top) and RDA (\%, bottom). Best scores are bolded and runner-up scores are underlined.}
\vskip -0.125in
\label{tab:textual_attack}
\definecolor{OverallShade}{RGB}{255, 245, 220}
\providecommand{\green}[1]{\textcolor{green!60!black}{#1}}
\providecommand{\red}[1]{\textcolor{red}{#1}}
\providecommand{\datasetheader}[1]{\raisebox{0.4ex}{\textbf{#1}}}
\providecommand{\levelheader}[1]{\raisebox{0.3ex}{\textbf{#1}}}
\setlength{\tabcolsep}{2pt}
\scriptsize
\resizebox{\textwidth}{!}{%
\begin{tabular}{l|ccc|ccc|ccc|ccc|ccc|ccc|c}
\toprule
\multirow{2}{*}{\textbf{Method}} & \multicolumn{3}{c}{\datasetheader{Cora}} & \multicolumn{3}{c|}{\datasetheader{Citeseer}} & \multicolumn{3}{c}{\datasetheader{History}} & \multicolumn{3}{c|}{\datasetheader{Child}} & \multicolumn{3}{c}{\datasetheader{Instagram}} & \multicolumn{3}{c|}{\datasetheader{WikiCS}} & \multirow{2}{*}{\textbf{Avg.}} \\[-3pt]
\cmidrule(lr){2-4}\cmidrule(lr){5-7}\cmidrule(lr){8-10}\cmidrule(lr){11-13}\cmidrule(lr){14-16}\cmidrule(lr){17-19}
 & \levelheader{Low} & \levelheader{Mid} & \levelheader{High} & \levelheader{Low} & \levelheader{Mid} & \levelheader{High} & \levelheader{Low} & \levelheader{Mid} & \levelheader{High} & \levelheader{Low} & \levelheader{Mid} & \levelheader{High} & \levelheader{Low} & \levelheader{Mid} & \levelheader{High} & \levelheader{Low} & \levelheader{Mid} & \levelheader{High} & \\[-2pt]
\midrule
\multicolumn{20}{c}{{\normalsize\itshape GNNPred}} \\
OFA & \makecell[c]{24.91\\[-1pt]{\scriptsize \red{$\downarrow$4.23}}} & \makecell[c]{23.06\\[-1pt]{\scriptsize \red{$\downarrow$11.34}}} & \makecell[c]{23.99\\[-1pt]{\scriptsize \red{$\downarrow$7.77}}} & \makecell[c]{42.32\\[-1pt]{\scriptsize \red{$\downarrow$6.58}}} & \makecell[c]{37.46\\[-1pt]{\scriptsize \red{$\downarrow$17.31}}} & \makecell[c]{28.21\\[-1pt]{\scriptsize \red{$\downarrow$37.73}}} & \makecell[c]{5.86\\[-1pt]{\scriptsize \red{$\downarrow$2.17}}} & \makecell[c]{5.96\\[-1pt]{\scriptsize \red{$\downarrow$0.50}}} & \makecell[c]{5.85\\[-1pt]{\scriptsize \red{$\downarrow$2.34}}} & \makecell[c]{0.26\\[-1pt]{\scriptsize \red{$\downarrow$3.70}}} & \makecell[c]{0.25\\[-1pt]{\scriptsize \red{$\downarrow$7.41}}} & \makecell[c]{0.24\\[-1pt]{\scriptsize \red{$\downarrow$11.11}}} & \makecell[c]{45.70\\[-1pt]{\scriptsize \red{$\downarrow$6.10}}} & \makecell[c]{44.69\\[-1pt]{\scriptsize \red{$\downarrow$8.18}}} & \makecell[c]{41.96\\[-1pt]{\scriptsize \red{$\downarrow$13.79}}} & \makecell[c]{9.65\\[-1pt]{\scriptsize \red{$\downarrow$3.40}}} & \makecell[c]{9.41\\[-1pt]{\scriptsize \red{$\downarrow$5.81}}} & \makecell[c]{9.32\\[-1pt]{\scriptsize \red{$\downarrow$6.71}}} & \cellcolor{OverallShade} \makecell[c]{19.95\\[-1pt]{\scriptsize \red{$\downarrow$7.95}}} \\
G2P2 & \makecell[c]{45.53\\[-1pt]{\scriptsize \red{$\downarrow$0.50}}} & \makecell[c]{45.02\\[-1pt]{\scriptsize \red{$\downarrow$1.62}}} & \makecell[c]{44.10\\[-1pt]{\scriptsize \red{$\downarrow$3.63}}} & \makecell[c]{51.10\\[-1pt]{\scriptsize \red{$\downarrow$4.95}}} & \makecell[c]{48.75\\[-1pt]{\scriptsize \red{$\downarrow$9.32}}} & \makecell[c]{\underline{47.34}\\[-1pt]{\scriptsize \red{$\downarrow$11.94}}} & \makecell[c]{12.84\\[-1pt]{\scriptsize \red{$\downarrow$1.83}}} & \makecell[c]{12.57\\[-1pt]{\scriptsize \red{$\downarrow$3.90}}} & \makecell[c]{12.30\\[-1pt]{\scriptsize \red{$\downarrow$5.96}}} & \makecell[c]{11.33\\[-1pt]{\scriptsize \red{$\downarrow$3.16}}} & \makecell[c]{11.23\\[-1pt]{\scriptsize \red{$\downarrow$4.02}}} & \makecell[c]{10.77\\[-1pt]{\scriptsize \red{$\downarrow$7.95}}} & \makecell[c]{44.34\\[-1pt]{\scriptsize \red{$\downarrow$2.51}}} & \makecell[c]{43.68\\[-1pt]{\scriptsize \red{$\downarrow$3.96}}} & \makecell[c]{42.44\\[-1pt]{\scriptsize \red{$\downarrow$6.68}}} & \makecell[c]{41.19\\[-1pt]{\scriptsize \red{$\downarrow$0.24}}} & \makecell[c]{41.05\\[-1pt]{\scriptsize \red{$\downarrow$0.58}}} & \makecell[c]{39.03\\[-1pt]{\scriptsize \red{$\downarrow$5.47}}} & \cellcolor{OverallShade} \makecell[c]{33.59\\[-1pt]{\scriptsize \red{$\downarrow$4.40}}} \\
ZeroG & \makecell[c]{52.01\\[-1pt]{\scriptsize \red{$\downarrow$0.04}}} & \makecell[c]{51.85\\[-1pt]{\scriptsize \red{$\downarrow$0.35}}} & \makecell[c]{\textbf{51.73}\\[-1pt]{\scriptsize \red{$\downarrow$0.58}}} & \makecell[c]{52.04\\[-1pt]{\scriptsize \red{$\downarrow$1.18}}} & \makecell[c]{\underline{49.84}\\[-1pt]{\scriptsize \red{$\downarrow$5.36}}} & \makecell[c]{\textbf{49.22}\\[-1pt]{\scriptsize \red{$\downarrow$6.53}}} & \makecell[c]{34.87\\[-1pt]{\scriptsize \red{$\downarrow$7.21}}} & \makecell[c]{32.62\\[-1pt]{\scriptsize \red{$\downarrow$13.20}}} & \makecell[c]{\underline{28.02}\\[-1pt]{\scriptsize \red{$\downarrow$25.44}}} & \makecell[c]{12.74\\[-1pt]{\scriptsize \red{$\downarrow$1.16}}} & \makecell[c]{12.55\\[-1pt]{\scriptsize \red{$\downarrow$2.64}}} & \makecell[c]{12.48\\[-1pt]{\scriptsize \red{$\downarrow$3.18}}} & \makecell[c]{55.44\\[-1pt]{\scriptsize \red{$\downarrow$1.88}}} & \makecell[c]{53.02\\[-1pt]{\scriptsize \red{$\downarrow$6.16}}} & \makecell[c]{51.34\\[-1pt]{\scriptsize \red{$\downarrow$9.13}}} & \makecell[c]{35.54\\[-1pt]{\scriptsize \red{$\downarrow$5.55}}} & \makecell[c]{32.63\\[-1pt]{\scriptsize \red{$\downarrow$13.29}}} & \makecell[c]{27.76\\[-1pt]{\scriptsize \red{$\downarrow$26.23}}} & \cellcolor{OverallShade} \makecell[c]{38.65\\[-1pt]{\scriptsize \red{$\downarrow$8.39}}} \\
GraphCLIP & \makecell[c]{59.78\\[-1pt]{\scriptsize \red{$\downarrow$5.26}}} & \makecell[c]{\underline{55.54}\\[-1pt]{\scriptsize \red{$\downarrow$11.98}}} & \makecell[c]{\underline{48.52}\\[-1pt]{\scriptsize \red{$\downarrow$23.11}}} & \makecell[c]{\textbf{63.79}\\[-1pt]{\scriptsize \red{$\downarrow$5.57}}} & \makecell[c]{\textbf{57.52}\\[-1pt]{\scriptsize \red{$\downarrow$14.85}}} & \makecell[c]{44.51\\[-1pt]{\scriptsize \red{$\downarrow$34.11}}} & \makecell[c]{\textbf{55.69}\\[-1pt]{\scriptsize \red{$\downarrow$2.04}}} & \makecell[c]{\textbf{54.55}\\[-1pt]{\scriptsize \red{$\downarrow$4.05}}} & \makecell[c]{\textbf{50.86}\\[-1pt]{\scriptsize \red{$\downarrow$10.54}}} & \makecell[c]{19.49\\[-1pt]{\scriptsize \red{$\downarrow$7.67}}} & \makecell[c]{18.06\\[-1pt]{\scriptsize \red{$\downarrow$14.45}}} & \makecell[c]{14.30\\[-1pt]{\scriptsize \red{$\downarrow$32.26}}} & \makecell[c]{\textbf{63.46}\\[-1pt]{\scriptsize \red{$\downarrow$2.10}}} & \makecell[c]{\textbf{63.42}\\[-1pt]{\scriptsize \red{$\downarrow$2.16}}} & \makecell[c]{\textbf{62.01}\\[-1pt]{\scriptsize \red{$\downarrow$4.34}}} & \makecell[c]{64.73\\[-1pt]{\scriptsize \red{$\downarrow$6.80}}} & \makecell[c]{58.77\\[-1pt]{\scriptsize \red{$\downarrow$15.38}}} & \makecell[c]{44.24\\[-1pt]{\scriptsize \red{$\downarrow$36.30}}} & \cellcolor{OverallShade} \makecell[c]{\textbf{49.96}\\[-1pt]{\scriptsize \red{$\downarrow$12.38}}} \\
\midrule
\multicolumn{20}{c}{{\normalsize\itshape GTextLLM}} \\
Llama3.1-8B & \makecell[c]{60.15\\[-1pt]{\scriptsize \red{$\downarrow$8.42}}} & \makecell[c]{53.32\\[-1pt]{\scriptsize \red{$\downarrow$18.82}}} & \makecell[c]{40.22\\[-1pt]{\scriptsize \red{$\downarrow$38.76}}} & \makecell[c]{\underline{52.98}\\[-1pt]{\scriptsize \red{$\downarrow$15.07}}} & \makecell[c]{46.55\\[-1pt]{\scriptsize \red{$\downarrow$25.38}}} & \makecell[c]{34.64\\[-1pt]{\scriptsize \red{$\downarrow$44.47}}} & \makecell[c]{\underline{36.83}\\[-1pt]{\scriptsize \red{$\downarrow$7.56}}} & \makecell[c]{\underline{33.52}\\[-1pt]{\scriptsize \red{$\downarrow$15.86}}} & \makecell[c]{26.81\\[-1pt]{\scriptsize \red{$\downarrow$32.71}}} & \makecell[c]{\textbf{27.39}\\[-1pt]{\scriptsize \red{$\downarrow$12.35}}} & \makecell[c]{\textbf{23.67}\\[-1pt]{\scriptsize \red{$\downarrow$24.26}}} & \makecell[c]{\textbf{17.10}\\[-1pt]{\scriptsize \red{$\downarrow$45.28}}} & \makecell[c]{41.34\\[-1pt]{\scriptsize \red{$\downarrow$8.40}}} & \makecell[c]{36.27\\[-1pt]{\scriptsize \red{$\downarrow$19.63}}} & \makecell[c]{30.85\\[-1pt]{\scriptsize \red{$\downarrow$31.64}}} & \makecell[c]{65.59\\[-1pt]{\scriptsize \red{$\downarrow$10.71}}} & \makecell[c]{58.34\\[-1pt]{\scriptsize \red{$\downarrow$20.58}}} & \makecell[c]{42.19\\[-1pt]{\scriptsize \red{$\downarrow$42.57}}} & \cellcolor{OverallShade} \makecell[c]{\underline{40.43}\\[-1pt]{\scriptsize \red{$\downarrow$23.55}}} \\
Qwen3-8B & \makecell[c]{\underline{61.81}\\[-1pt]{\scriptsize \red{$\downarrow$9.21}}} & \makecell[c]{\textbf{56.46}\\[-1pt]{\scriptsize \red{$\downarrow$17.07}}} & \makecell[c]{41.51\\[-1pt]{\scriptsize \red{$\downarrow$39.03}}} & \makecell[c]{52.51\\[-1pt]{\scriptsize \red{$\downarrow$10.89}}} & \makecell[c]{44.67\\[-1pt]{\scriptsize \red{$\downarrow$24.20}}} & \makecell[c]{31.19\\[-1pt]{\scriptsize \red{$\downarrow$47.07}}} & \makecell[c]{22.84\\[-1pt]{\scriptsize \red{$\downarrow$7.53}}} & \makecell[c]{21.18\\[-1pt]{\scriptsize \red{$\downarrow$14.25}}} & \makecell[c]{17.01\\[-1pt]{\scriptsize \red{$\downarrow$31.13}}} & \makecell[c]{\underline{26.78}\\[-1pt]{\scriptsize \red{$\downarrow$12.05}}} & \makecell[c]{\underline{23.40}\\[-1pt]{\scriptsize \red{$\downarrow$23.15}}} & \makecell[c]{\underline{16.87}\\[-1pt]{\scriptsize \red{$\downarrow$44.60}}} & \makecell[c]{43.76\\[-1pt]{\scriptsize \red{$\downarrow$4.99}}} & \makecell[c]{36.80\\[-1pt]{\scriptsize \red{$\downarrow$20.10}}} & \makecell[c]{30.41\\[-1pt]{\scriptsize \red{$\downarrow$33.98}}} & \makecell[c]{\textbf{68.91}\\[-1pt]{\scriptsize \red{$\downarrow$10.68}}} & \makecell[c]{\underline{60.65}\\[-1pt]{\scriptsize \red{$\downarrow$21.39}}} & \makecell[c]{\underline{44.71}\\[-1pt]{\scriptsize \red{$\downarrow$42.05}}} & \cellcolor{OverallShade} \makecell[c]{38.97\\[-1pt]{\scriptsize \red{$\downarrow$22.94}}} \\
Vicuna-7B & \makecell[c]{51.48\\[-1pt]{\scriptsize \red{$\downarrow$10.00}}} & \makecell[c]{47.42\\[-1pt]{\scriptsize \red{$\downarrow$17.10}}} & \makecell[c]{35.24\\[-1pt]{\scriptsize \red{$\downarrow$38.39}}} & \makecell[c]{40.44\\[-1pt]{\scriptsize \red{$\downarrow$12.24}}} & \makecell[c]{39.50\\[-1pt]{\scriptsize \red{$\downarrow$14.28}}} & \makecell[c]{29.94\\[-1pt]{\scriptsize \red{$\downarrow$35.03}}} & \makecell[c]{21.22\\[-1pt]{\scriptsize \red{$\downarrow$0.89}}} & \makecell[c]{20.01\\[-1pt]{\scriptsize \red{$\downarrow$6.54}}} & \makecell[c]{19.25\\[-1pt]{\scriptsize \red{$\downarrow$10.09}}} & \makecell[c]{16.26\\[-1pt]{\scriptsize \red{$\downarrow$5.68}}} & \makecell[c]{14.61\\[-1pt]{\scriptsize \red{$\downarrow$15.26}}} & \makecell[c]{12.57\\[-1pt]{\scriptsize \red{$\downarrow$27.09}}} & \makecell[c]{29.44\\[-1pt]{\scriptsize \red{$\downarrow$4.57}}} & \makecell[c]{28.03\\[-1pt]{\scriptsize \red{$\downarrow$9.14}}} & \makecell[c]{25.12\\[-1pt]{\scriptsize \red{$\downarrow$18.57}}} & \makecell[c]{38.09\\[-1pt]{\scriptsize \red{$\downarrow$10.44}}} & \makecell[c]{32.85\\[-1pt]{\scriptsize \red{$\downarrow$22.76}}} & \makecell[c]{22.49\\[-1pt]{\scriptsize \red{$\downarrow$47.12}}} & \cellcolor{OverallShade} \makecell[c]{29.11\\[-1pt]{\scriptsize \red{$\downarrow$16.09}}} \\
G1 & \makecell[c]{58.67\\[-1pt]{\scriptsize \red{$\downarrow$14.29}}} & \makecell[c]{54.61\\[-1pt]{\scriptsize \red{$\downarrow$20.22}}} & \makecell[c]{40.22\\[-1pt]{\scriptsize \red{$\downarrow$41.24}}} & \makecell[c]{46.71\\[-1pt]{\scriptsize \red{$\downarrow$12.61}}} & \makecell[c]{41.85\\[-1pt]{\scriptsize \red{$\downarrow$21.70}}} & \makecell[c]{28.21\\[-1pt]{\scriptsize \red{$\downarrow$47.22}}} & \makecell[c]{24.81\\[-1pt]{\scriptsize \red{$\downarrow$6.45}}} & \makecell[c]{23.52\\[-1pt]{\scriptsize \red{$\downarrow$11.31}}} & \makecell[c]{19.49\\[-1pt]{\scriptsize \red{$\downarrow$26.51}}} & \makecell[c]{23.02\\[-1pt]{\scriptsize \red{$\downarrow$11.33}}} & \makecell[c]{21.13\\[-1pt]{\scriptsize \red{$\downarrow$18.61}}} & \makecell[c]{16.38\\[-1pt]{\scriptsize \red{$\downarrow$36.90}}} & \makecell[c]{45.04\\[-1pt]{\scriptsize \red{$\downarrow$10.97}}} & \makecell[c]{39.53\\[-1pt]{\scriptsize \red{$\downarrow$21.86}}} & \makecell[c]{30.67\\[-1pt]{\scriptsize \red{$\downarrow$39.38}}} & \makecell[c]{\underline{68.09}\\[-1pt]{\scriptsize \red{$\downarrow$9.13}}} & \makecell[c]{\textbf{61.67}\\[-1pt]{\scriptsize \red{$\downarrow$17.70}}} & \makecell[c]{\textbf{46.11}\\[-1pt]{\scriptsize \red{$\downarrow$38.46}}} & \cellcolor{OverallShade} \makecell[c]{38.32\\[-1pt]{\scriptsize \red{$\downarrow$22.19}}} \\
\midrule
\multicolumn{20}{c}{{\normalsize\itshape GTokenLLM}} \\
GraphGPT & \makecell[c]{16.24\\[-1pt]{\scriptsize \red{$\downarrow$20.00}}} & \makecell[c]{15.50\\[-1pt]{\scriptsize \red{$\downarrow$23.65}}} & \makecell[c]{10.70\\[-1pt]{\scriptsize \red{$\downarrow$47.29}}} & \makecell[c]{29.31\\[-1pt]{\scriptsize \red{$\downarrow$10.09}}} & \makecell[c]{25.39\\[-1pt]{\scriptsize \red{$\downarrow$22.12}}} & \makecell[c]{17.71\\[-1pt]{\scriptsize \red{$\downarrow$45.67}}} & \makecell[c]{18.93\\[-1pt]{\scriptsize \red{$\downarrow$15.72}}} & \makecell[c]{15.93\\[-1pt]{\scriptsize \red{$\downarrow$29.07}}} & \makecell[c]{10.66\\[-1pt]{\scriptsize \red{$\downarrow$52.54}}} & \makecell[c]{4.99\\[-1pt]{\scriptsize \red{$\downarrow$19.26}}} & \makecell[c]{4.21\\[-1pt]{\scriptsize \red{$\downarrow$31.88}}} & \makecell[c]{3.53\\[-1pt]{\scriptsize \red{$\downarrow$42.88}}} & \makecell[c]{33.22\\[-1pt]{\scriptsize \red{$\downarrow$2.95}}} & \makecell[c]{31.93\\[-1pt]{\scriptsize \red{$\downarrow$6.72}}} & \makecell[c]{31.16\\[-1pt]{\scriptsize \red{$\downarrow$8.97}}} & \makecell[c]{46.14\\[-1pt]{\scriptsize \red{$\downarrow$10.93}}} & \makecell[c]{42.04\\[-1pt]{\scriptsize \red{$\downarrow$18.84}}} & \makecell[c]{31.52\\[-1pt]{\scriptsize \red{$\downarrow$39.15}}} & \cellcolor{OverallShade} \makecell[c]{21.62\\[-1pt]{\scriptsize \red{$\downarrow$26.47}}} \\
LLaGA & \makecell[c]{12.18\\[-1pt]{\scriptsize \red{$\downarrow$17.48}}} & \makecell[c]{11.07\\[-1pt]{\scriptsize \red{$\downarrow$25.00}}} & \makecell[c]{8.30\\[-1pt]{\scriptsize \red{$\downarrow$43.77}}} & \makecell[c]{16.46\\[-1pt]{\scriptsize \red{$\downarrow$6.21}}} & \makecell[c]{14.89\\[-1pt]{\scriptsize \red{$\downarrow$15.16}}} & \makecell[c]{9.25\\[-1pt]{\scriptsize \red{$\downarrow$47.29}}} & \makecell[c]{21.49\\[-1pt]{\scriptsize \red{$\downarrow$6.97}}} & \makecell[c]{19.61\\[-1pt]{\scriptsize \red{$\downarrow$15.11}}} & \makecell[c]{15.35\\[-1pt]{\scriptsize \red{$\downarrow$33.55}}} & \makecell[c]{14.43\\[-1pt]{\scriptsize \red{$\downarrow$8.15}}} & \makecell[c]{13.35\\[-1pt]{\scriptsize \red{$\downarrow$15.02}}} & \makecell[c]{10.82\\[-1pt]{\scriptsize \red{$\downarrow$31.13}}} & \makecell[c]{20.67\\[-1pt]{\scriptsize \red{$\downarrow$3.09}}} & \makecell[c]{19.83\\[-1pt]{\scriptsize \red{$\downarrow$7.03}}} & \makecell[c]{19.66\\[-1pt]{\scriptsize \red{$\downarrow$7.83}}} & \makecell[c]{37.95\\[-1pt]{\scriptsize \red{$\downarrow$8.58}}} & \makecell[c]{35.33\\[-1pt]{\scriptsize \red{$\downarrow$14.89}}} & \makecell[c]{27.06\\[-1pt]{\scriptsize \red{$\downarrow$34.81}}} & \cellcolor{OverallShade} \makecell[c]{18.21\\[-1pt]{\scriptsize \red{$\downarrow$19.49}}} \\
TEA-GLM & \makecell[c]{43.54\\[-1pt]{\scriptsize \red{$\downarrow$10.61}}} & \makecell[c]{39.67\\[-1pt]{\scriptsize \red{$\downarrow$18.56}}} & \makecell[c]{29.15\\[-1pt]{\scriptsize \red{$\downarrow$40.16}}} & \makecell[c]{48.43\\[-1pt]{\scriptsize \red{$\downarrow$8.59}}} & \makecell[c]{41.38\\[-1pt]{\scriptsize \red{$\downarrow$21.90}}} & \makecell[c]{27.59\\[-1pt]{\scriptsize \red{$\downarrow$47.92}}} & \makecell[c]{26.58\\[-1pt]{\scriptsize \red{$\downarrow$8.50}}} & \makecell[c]{24.57\\[-1pt]{\scriptsize \red{$\downarrow$15.42}}} & \makecell[c]{19.06\\[-1pt]{\scriptsize \red{$\downarrow$34.39}}} & \makecell[c]{17.67\\[-1pt]{\scriptsize \red{$\downarrow$9.43}}} & \makecell[c]{16.09\\[-1pt]{\scriptsize \red{$\downarrow$17.53}}} & \makecell[c]{13.50\\[-1pt]{\scriptsize \red{$\downarrow$30.80}}} & \makecell[c]{\underline{60.29}\\[-1pt]{\scriptsize \red{$\downarrow$1.79}}} & \makecell[c]{\underline{56.63}\\[-1pt]{\scriptsize \red{$\downarrow$7.75}}} & \makecell[c]{\underline{51.87}\\[-1pt]{\scriptsize \red{$\downarrow$15.51}}} & \makecell[c]{28.60\\[-1pt]{\scriptsize \red{$\downarrow$11.67}}} & \makecell[c]{25.21\\[-1pt]{\scriptsize \red{$\downarrow$22.14}}} & \makecell[c]{18.06\\[-1pt]{\scriptsize \red{$\downarrow$44.22}}} & \cellcolor{OverallShade} \makecell[c]{32.66\\[-1pt]{\scriptsize \red{$\downarrow$20.84}}} \\
GOFA & \makecell[c]{\textbf{62.36}\\[-1pt]{\scriptsize \red{$\downarrow$7.66}}} & \makecell[c]{54.24\\[-1pt]{\scriptsize \red{$\downarrow$19.68}}} & \makecell[c]{44.46\\[-1pt]{\scriptsize \red{$\downarrow$34.16}}} & \makecell[c]{39.50\\[-1pt]{\scriptsize \red{$\downarrow$11.26}}} & \makecell[c]{34.48\\[-1pt]{\scriptsize \red{$\downarrow$22.53}}} & \makecell[c]{23.98\\[-1pt]{\scriptsize \red{$\downarrow$46.12}}} & \makecell[c]{26.41\\[-1pt]{\scriptsize \red{$\downarrow$6.84}}} & \makecell[c]{24.76\\[-1pt]{\scriptsize \red{$\downarrow$12.66}}} & \makecell[c]{20.74\\[-1pt]{\scriptsize \red{$\downarrow$26.84}}} & \makecell[c]{18.84\\[-1pt]{\scriptsize \red{$\downarrow$7.56}}} & \makecell[c]{17.84\\[-1pt]{\scriptsize \red{$\downarrow$12.46}}} & \makecell[c]{15.56\\[-1pt]{\scriptsize \red{$\downarrow$23.65}}} & \makecell[c]{36.10\\[-1pt]{\scriptsize \red{$\downarrow$8.79}}} & \makecell[c]{32.66\\[-1pt]{\scriptsize \red{$\downarrow$17.48}}} & \makecell[c]{29.00\\[-1pt]{\scriptsize \red{$\downarrow$26.73}}} & \makecell[c]{59.16\\[-1pt]{\scriptsize \red{$\downarrow$10.43}}} & \makecell[c]{52.61\\[-1pt]{\scriptsize \red{$\downarrow$20.35}}} & \makecell[c]{38.21\\[-1pt]{\scriptsize \red{$\downarrow$42.15}}} & \cellcolor{OverallShade} \makecell[c]{35.05\\[-1pt]{\scriptsize \red{$\downarrow$19.84}}} \\
UniGTE & \makecell[c]{33.03\\[-1pt]{\scriptsize \red{$\downarrow$10.05}}} & \makecell[c]{28.23\\[-1pt]{\scriptsize \red{$\downarrow$23.12}}} & \makecell[c]{22.14\\[-1pt]{\scriptsize \red{$\downarrow$39.71}}} & \makecell[c]{28.53\\[-1pt]{\scriptsize \red{$\downarrow$11.20}}} & \makecell[c]{25.08\\[-1pt]{\scriptsize \red{$\downarrow$21.94}}} & \makecell[c]{17.40\\[-1pt]{\scriptsize \red{$\downarrow$45.85}}} & \makecell[c]{12.93\\[-1pt]{\scriptsize \red{$\downarrow$1.90}}} & \makecell[c]{12.80\\[-1pt]{\scriptsize \red{$\downarrow$2.88}}} & \makecell[c]{12.00\\[-1pt]{\scriptsize \red{$\downarrow$8.95}}} & \makecell[c]{15.69\\[-1pt]{\scriptsize \red{$\downarrow$5.31}}} & \makecell[c]{14.61\\[-1pt]{\scriptsize \red{$\downarrow$11.83}}} & \makecell[c]{11.79\\[-1pt]{\scriptsize \red{$\downarrow$28.85}}} & \makecell[c]{46.01\\[-1pt]{\scriptsize \red{$\downarrow$0.95}}} & \makecell[c]{45.48\\[-1pt]{\scriptsize \red{$\downarrow$2.09}}} & \makecell[c]{45.34\\[-1pt]{\scriptsize \red{$\downarrow$2.39}}} & \makecell[c]{45.66\\[-1pt]{\scriptsize \red{$\downarrow$8.22}}} & \makecell[c]{40.74\\[-1pt]{\scriptsize \red{$\downarrow$18.11}}} & \makecell[c]{30.15\\[-1pt]{\scriptsize \red{$\downarrow$39.40}}} & \cellcolor{OverallShade} \makecell[c]{27.09\\[-1pt]{\scriptsize \red{$\downarrow$14.73}}} \\
\bottomrule
\end{tabular}%
}
\vskip -0.175in
\end{table*}

\noindent \textbf{\ding{172} Within GNNPreds, parameter-free similarity prediction shows stronger zero-shot transferability than training a set of binary classifiers.} As reported in Table~\ref{tab:clean_acc}, G2P2, ZeroG, and GraphCLIP achieve average accuracies of 35.18\%, 41.55\%, and 57.15\%, respectively, all outperforming OFA, with GraphCLIP ranking first overall. One possible reason is that similarity-based methods use a consistent objective during training and inference, whereas a parametric classifier may overfit the classification patterns of the source graphs. GraphCLIP further benefits from the semantic information provided by LLM-generated graph summaries.


\noindent \textbf{\ding{173} GTextLLMs generally achieve stronger clean zero-shot performance than GTokenLLMs.} Llama3.1-8B, Qwen3-8B, and G1 achieve average ranks of 3.50, 3.67, and 3.50, respectively, whereas the evaluated GTokenLLMs rank between 6.33 and 10.83 (Table~\ref{tab:clean_acc}). This is because existing GTokenLLMs use the relatively weaker Vicuna-7B as LLM backbone. It suggests that graph tokens provide useful information, but their benefits are generally insufficient to close the gap with GTextLLMs built on stronger LLM backbones.

\subsection{Robustness against Structural Attacks}
\noindent \textbf{\ding{174} Structural attacks are more effective against GNNPreds than GTextLLMs and GTokenLLMs.} In Table~\ref{tab:structural_attack}, GNNPreds suffer an average RDA of 18.37\%, compared with only 3.10\% and 2.45\% for GTextLLMs and GTokenLLMs. 
Moreover, as the attack strength increases, the accuracy of GTextLLMs and GTokenLLMs even improves in some cases. On the one hand, GNNPreds rely mainly on GNNs to aggregate neighborhood information over the perturbed graph and thus directly inherit their vulnerability to structure changes. On the other hand, GTextLLMs and GTokenLLMs use LLMs for prediction and typically input the clean text of the central node. When neighborhood information is corrupted, they may rely more on clean text from the central node, reducing the effect of structural perturbations.

\noindent \textbf{\ding{175} GTextLLMs equipped with stronger LLM backbones exhibit greater robustness to structural attacks.}
Llama3.1-8B, Qwen3-8B, and G1 show almost no average performance degradation under structural attacks, whereas Vicuna-7B suffers an average drop of 13.06\% (Table~\ref{tab:structural_attack}). This suggests that, when neighborhood information is perturbed, stronger LLMs can better leverage the clean textual semantics of the central node and candidate labels, thereby reducing their dependence on the corrupted graph context.

\begin{table*}[t]
\centering
\caption{Performance comparison under node injection attacks across different perturbation sizes. Each cell reports the ACC (top) and RDA (\%, bottom). Best scores are bolded and runner-up scores are underlined.}
\vskip -0.125in
\label{tab:hybrid_attack}
\definecolor{OverallShade}{RGB}{255, 245, 220}
\providecommand{\green}[1]{\textcolor{green!60!black}{#1}}
\providecommand{\red}[1]{\textcolor{red}{#1}}
\providecommand{\datasetheader}[1]{\raisebox{0.4ex}{\textbf{#1}}}
\providecommand{\levelheader}[1]{\raisebox{0.3ex}{\textbf{#1}}}
\setlength{\tabcolsep}{2pt}
\scriptsize
\resizebox{\textwidth}{!}{%
\begin{tabular}{l|ccc|ccc|ccc|ccc|ccc|ccc|c}
\toprule
\multirow{2}{*}{\textbf{Method}} & \multicolumn{3}{c}{\datasetheader{Cora}} & \multicolumn{3}{c|}{\datasetheader{Citeseer}} & \multicolumn{3}{c}{\datasetheader{History}} & \multicolumn{3}{c|}{\datasetheader{Child}} & \multicolumn{3}{c}{\datasetheader{Instagram}} & \multicolumn{3}{c|}{\datasetheader{WikiCS}} & \multirow{2}{*}{\textbf{Avg.}} \\[-3pt]
\cmidrule(lr){2-4}\cmidrule(lr){5-7}\cmidrule(lr){8-10}\cmidrule(lr){11-13}\cmidrule(lr){14-16}\cmidrule(lr){17-19}
 & \levelheader{Low} & \levelheader{Mid} & \levelheader{High} & \levelheader{Low} & \levelheader{Mid} & \levelheader{High} & \levelheader{Low} & \levelheader{Mid} & \levelheader{High} & \levelheader{Low} & \levelheader{Mid} & \levelheader{High} & \levelheader{Low} & \levelheader{Mid} & \levelheader{High} & \levelheader{Low} & \levelheader{Mid} & \levelheader{High} & \\[-2pt]
\midrule
\multicolumn{20}{c}{{\normalsize\itshape GNNPred}} \\
OFA & \makecell[c]{25.83\\[-1pt]{\scriptsize \red{$\downarrow$0.69}}} & \makecell[c]{25.38\\[-1pt]{\scriptsize \red{$\downarrow$2.42}}} & \makecell[c]{25.01\\[-1pt]{\scriptsize \red{$\downarrow$3.84}}} & \makecell[c]{43.73\\[-1pt]{\scriptsize \red{$\downarrow$3.47}}} & \makecell[c]{43.10\\[-1pt]{\scriptsize \red{$\downarrow$4.86}}} & \makecell[c]{40.60\\[-1pt]{\scriptsize \red{$\downarrow$10.38}}} & \makecell[c]{5.95\\[-1pt]{\scriptsize \red{$\downarrow$0.67}}} & \makecell[c]{5.72\\[-1pt]{\scriptsize \red{$\downarrow$4.51}}} & \makecell[c]{5.05\\[-1pt]{\scriptsize \red{$\downarrow$15.69}}} & \makecell[c]{0.12\\[-1pt]{\scriptsize \red{$\downarrow$55.56}}} & \makecell[c]{0.12\\[-1pt]{\scriptsize \red{$\downarrow$55.56}}} & \makecell[c]{0.12\\[-1pt]{\scriptsize \red{$\downarrow$55.56}}} & \makecell[c]{50.42\\[-1pt]{\scriptsize \green{$\uparrow$3.60}}} & \makecell[c]{51.74\\[-1pt]{\scriptsize \green{$\uparrow$6.31}}} & \makecell[c]{53.55\\[-1pt]{\scriptsize \green{$\uparrow$10.03}}} & \makecell[c]{8.64\\[-1pt]{\scriptsize \red{$\downarrow$13.51}}} & \makecell[c]{8.36\\[-1pt]{\scriptsize \red{$\downarrow$16.32}}} & \makecell[c]{7.82\\[-1pt]{\scriptsize \red{$\downarrow$21.72}}} & \cellcolor{OverallShade} \makecell[c]{22.29\\[-1pt]{\scriptsize \red{$\downarrow$13.25}}} \\
G2P2 & \makecell[c]{44.28\\[-1pt]{\scriptsize \red{$\downarrow$3.23}}} & \makecell[c]{42.80\\[-1pt]{\scriptsize \red{$\downarrow$6.47}}} & \makecell[c]{40.59\\[-1pt]{\scriptsize \red{$\downarrow$11.30}}} & \makecell[c]{48.28\\[-1pt]{\scriptsize \red{$\downarrow$10.19}}} & \makecell[c]{47.34\\[-1pt]{\scriptsize \red{$\downarrow$11.94}}} & \makecell[c]{44.36\\[-1pt]{\scriptsize \red{$\downarrow$17.49}}} & \makecell[c]{11.14\\[-1pt]{\scriptsize \red{$\downarrow$14.83}}} & \makecell[c]{11.88\\[-1pt]{\scriptsize \red{$\downarrow$9.17}}} & \makecell[c]{12.57\\[-1pt]{\scriptsize \red{$\downarrow$3.90}}} & \makecell[c]{9.95\\[-1pt]{\scriptsize \red{$\downarrow$14.96}}} & \makecell[c]{9.33\\[-1pt]{\scriptsize \red{$\downarrow$20.26}}} & \makecell[c]{7.78\\[-1pt]{\scriptsize \red{$\downarrow$33.50}}} & \makecell[c]{45.29\\[-1pt]{\scriptsize \red{$\downarrow$0.42}}} & \makecell[c]{45.21\\[-1pt]{\scriptsize \red{$\downarrow$0.59}}} & \makecell[c]{45.16\\[-1pt]{\scriptsize \red{$\downarrow$0.70}}} & \makecell[c]{40.74\\[-1pt]{\scriptsize \red{$\downarrow$1.33}}} & \makecell[c]{40.36\\[-1pt]{\scriptsize \red{$\downarrow$2.25}}} & \makecell[c]{38.96\\[-1pt]{\scriptsize \red{$\downarrow$5.64}}} & \cellcolor{OverallShade} \makecell[c]{32.56\\[-1pt]{\scriptsize \red{$\downarrow$9.06}}} \\
ZeroG & \makecell[c]{50.00\\[-1pt]{\scriptsize \red{$\downarrow$3.90}}} & \makecell[c]{48.89\\[-1pt]{\scriptsize \red{$\downarrow$6.03}}} & \makecell[c]{38.75\\[-1pt]{\scriptsize \red{$\downarrow$25.52}}} & \makecell[c]{47.96\\[-1pt]{\scriptsize \red{$\downarrow$8.93}}} & \makecell[c]{40.75\\[-1pt]{\scriptsize \red{$\downarrow$22.62}}} & \makecell[c]{31.50\\[-1pt]{\scriptsize \red{$\downarrow$40.18}}} & \makecell[c]{12.75\\[-1pt]{\scriptsize \red{$\downarrow$66.07}}} & \makecell[c]{6.88\\[-1pt]{\scriptsize \red{$\downarrow$81.69}}} & \makecell[c]{4.43\\[-1pt]{\scriptsize \red{$\downarrow$88.21}}} & \makecell[c]{7.69\\[-1pt]{\scriptsize \red{$\downarrow$40.34}}} & \makecell[c]{4.70\\[-1pt]{\scriptsize \red{$\downarrow$63.54}}} & \makecell[c]{3.02\\[-1pt]{\scriptsize \red{$\downarrow$76.57}}} & \makecell[c]{41.82\\[-1pt]{\scriptsize \red{$\downarrow$25.98}}} & \makecell[c]{36.23\\[-1pt]{\scriptsize \red{$\downarrow$35.88}}} & \makecell[c]{35.43\\[-1pt]{\scriptsize \red{$\downarrow$37.29}}} & \makecell[c]{33.83\\[-1pt]{\scriptsize \red{$\downarrow$10.10}}} & \makecell[c]{36.00\\[-1pt]{\scriptsize \red{$\downarrow$4.33}}} & \makecell[c]{28.37\\[-1pt]{\scriptsize \red{$\downarrow$24.61}}} & \cellcolor{OverallShade} \makecell[c]{28.28\\[-1pt]{\scriptsize \red{$\downarrow$41.59}}} \\
GraphCLIP & \makecell[c]{60.33\\[-1pt]{\scriptsize \red{$\downarrow$4.39}}} & \makecell[c]{57.38\\[-1pt]{\scriptsize \red{$\downarrow$9.06}}} & \makecell[c]{45.94\\[-1pt]{\scriptsize \red{$\downarrow$27.19}}} & \makecell[c]{\textbf{65.52}\\[-1pt]{\scriptsize \red{$\downarrow$3.01}}} & \makecell[c]{\textbf{62.23}\\[-1pt]{\scriptsize \red{$\downarrow$7.88}}} & \makecell[c]{\underline{60.82}\\[-1pt]{\scriptsize \red{$\downarrow$9.96}}} & \makecell[c]{\underline{46.65}\\[-1pt]{\scriptsize \red{$\downarrow$17.94}}} & \makecell[c]{\underline{41.41}\\[-1pt]{\scriptsize \red{$\downarrow$27.16}}} & \makecell[c]{\underline{36.16}\\[-1pt]{\scriptsize \red{$\downarrow$36.39}}} & \makecell[c]{20.24\\[-1pt]{\scriptsize \red{$\downarrow$4.12}}} & \makecell[c]{20.46\\[-1pt]{\scriptsize \red{$\downarrow$3.08}}} & \makecell[c]{20.22\\[-1pt]{\scriptsize \red{$\downarrow$4.22}}} & \makecell[c]{\textbf{62.85}\\[-1pt]{\scriptsize \red{$\downarrow$3.04}}} & \makecell[c]{\textbf{62.98}\\[-1pt]{\scriptsize \red{$\downarrow$2.84}}} & \makecell[c]{\textbf{62.14}\\[-1pt]{\scriptsize \red{$\downarrow$4.13}}} & \makecell[c]{63.95\\[-1pt]{\scriptsize \red{$\downarrow$7.92}}} & \makecell[c]{62.48\\[-1pt]{\scriptsize \red{$\downarrow$10.04}}} & \makecell[c]{60.03\\[-1pt]{\scriptsize \red{$\downarrow$13.56}}} & \cellcolor{OverallShade} \makecell[c]{\underline{50.66}\\[-1pt]{\scriptsize \red{$\downarrow$12.97}}} \\
\midrule
\multicolumn{20}{c}{{\normalsize\itshape GTextLLM}} \\
Llama3.1-8B & \makecell[c]{\underline{67.53}\\[-1pt]{\scriptsize \green{$\uparrow$2.82}}} & \makecell[c]{\underline{66.05}\\[-1pt]{\scriptsize \green{$\uparrow$0.56}}} & \makecell[c]{64.94\\[-1pt]{\scriptsize \red{$\downarrow$1.13}}} & \makecell[c]{\underline{60.82}\\[-1pt]{\scriptsize \red{$\downarrow$2.50}}} & \makecell[c]{\underline{59.72}\\[-1pt]{\scriptsize \red{$\downarrow$4.26}}} & \makecell[c]{\textbf{62.07}\\[-1pt]{\scriptsize \red{$\downarrow$0.50}}} & \makecell[c]{\textbf{46.94}\\[-1pt]{\scriptsize \green{$\uparrow$17.82}}} & \makecell[c]{\textbf{47.60}\\[-1pt]{\scriptsize \green{$\uparrow$19.48}}} & \makecell[c]{\textbf{44.82}\\[-1pt]{\scriptsize \green{$\uparrow$12.50}}} & \makecell[c]{\textbf{30.73}\\[-1pt]{\scriptsize \red{$\downarrow$1.66}}} & \makecell[c]{\textbf{30.87}\\[-1pt]{\scriptsize \red{$\downarrow$1.22}}} & \makecell[c]{\textbf{30.51}\\[-1pt]{\scriptsize \red{$\downarrow$2.37}}} & \makecell[c]{46.94\\[-1pt]{\scriptsize \green{$\uparrow$4.01}}} & \makecell[c]{46.41\\[-1pt]{\scriptsize \green{$\uparrow$2.84}}} & \makecell[c]{46.36\\[-1pt]{\scriptsize \green{$\uparrow$2.73}}} & \makecell[c]{72.62\\[-1pt]{\scriptsize \red{$\downarrow$1.14}}} & \makecell[c]{72.04\\[-1pt]{\scriptsize \red{$\downarrow$1.93}}} & \makecell[c]{\underline{71.16}\\[-1pt]{\scriptsize \red{$\downarrow$3.13}}} & \cellcolor{OverallShade} \makecell[c]{\textbf{53.79}\\[-1pt]{\scriptsize \green{$\uparrow$3.74}}} \\
Qwen3-8B & \makecell[c]{\textbf{68.63}\\[-1pt]{\scriptsize \green{$\uparrow$0.81}}} & \makecell[c]{\textbf{66.79}\\[-1pt]{\scriptsize \red{$\downarrow$1.89}}} & \makecell[c]{\textbf{66.42}\\[-1pt]{\scriptsize \red{$\downarrow$2.44}}} & \makecell[c]{56.74\\[-1pt]{\scriptsize \red{$\downarrow$3.72}}} & \makecell[c]{57.68\\[-1pt]{\scriptsize \red{$\downarrow$2.12}}} & \makecell[c]{58.46\\[-1pt]{\scriptsize \red{$\downarrow$0.80}}} & \makecell[c]{25.56\\[-1pt]{\scriptsize \green{$\uparrow$3.48}}} & \makecell[c]{25.89\\[-1pt]{\scriptsize \green{$\uparrow$4.82}}} & \makecell[c]{26.19\\[-1pt]{\scriptsize \green{$\uparrow$6.03}}} & \makecell[c]{\underline{28.96}\\[-1pt]{\scriptsize \red{$\downarrow$4.89}}} & \makecell[c]{\underline{28.53}\\[-1pt]{\scriptsize \red{$\downarrow$6.31}}} & \makecell[c]{\underline{28.44}\\[-1pt]{\scriptsize \red{$\downarrow$6.60}}} & \makecell[c]{47.20\\[-1pt]{\scriptsize \green{$\uparrow$2.48}}} & \makecell[c]{45.39\\[-1pt]{\scriptsize \red{$\downarrow$1.45}}} & \makecell[c]{47.47\\[-1pt]{\scriptsize \green{$\uparrow$3.06}}} & \makecell[c]{\textbf{76.26}\\[-1pt]{\scriptsize \red{$\downarrow$1.15}}} & \makecell[c]{\textbf{75.41}\\[-1pt]{\scriptsize \red{$\downarrow$2.26}}} & \makecell[c]{\textbf{73.75}\\[-1pt]{\scriptsize \red{$\downarrow$4.41}}} & \cellcolor{OverallShade} \makecell[c]{50.21\\[-1pt]{\scriptsize \red{$\downarrow$0.33}}} \\
Vicuna-7B & \makecell[c]{57.01\\[-1pt]{\scriptsize \red{$\downarrow$0.33}}} & \makecell[c]{52.40\\[-1pt]{\scriptsize \red{$\downarrow$8.39}}} & \makecell[c]{45.57\\[-1pt]{\scriptsize \red{$\downarrow$20.33}}} & \makecell[c]{42.01\\[-1pt]{\scriptsize \red{$\downarrow$8.83}}} & \makecell[c]{42.79\\[-1pt]{\scriptsize \red{$\downarrow$7.14}}} & \makecell[c]{35.27\\[-1pt]{\scriptsize \red{$\downarrow$23.46}}} & \makecell[c]{21.12\\[-1pt]{\scriptsize \red{$\downarrow$1.35}}} & \makecell[c]{22.15\\[-1pt]{\scriptsize \green{$\uparrow$3.46}}} & \makecell[c]{21.10\\[-1pt]{\scriptsize \red{$\downarrow$1.45}}} & \makecell[c]{14.84\\[-1pt]{\scriptsize \red{$\downarrow$13.92}}} & \makecell[c]{14.98\\[-1pt]{\scriptsize \red{$\downarrow$13.11}}} & \makecell[c]{14.80\\[-1pt]{\scriptsize \red{$\downarrow$14.15}}} & \makecell[c]{31.12\\[-1pt]{\scriptsize \green{$\uparrow$0.88}}} & \makecell[c]{30.81\\[-1pt]{\scriptsize \red{$\downarrow$0.13}}} & \makecell[c]{29.18\\[-1pt]{\scriptsize \red{$\downarrow$5.41}}} & \makecell[c]{38.55\\[-1pt]{\scriptsize \red{$\downarrow$9.36}}} & \makecell[c]{36.02\\[-1pt]{\scriptsize \red{$\downarrow$15.31}}} & \makecell[c]{34.02\\[-1pt]{\scriptsize \red{$\downarrow$20.01}}} & \cellcolor{OverallShade} \makecell[c]{32.43\\[-1pt]{\scriptsize \red{$\downarrow$7.82}}} \\
G1 & \makecell[c]{65.87\\[-1pt]{\scriptsize \red{$\downarrow$3.77}}} & \makecell[c]{64.21\\[-1pt]{\scriptsize \red{$\downarrow$6.19}}} & \makecell[c]{60.89\\[-1pt]{\scriptsize \red{$\downarrow$11.04}}} & \makecell[c]{50.94\\[-1pt]{\scriptsize \red{$\downarrow$4.70}}} & \makecell[c]{50.63\\[-1pt]{\scriptsize \red{$\downarrow$5.28}}} & \makecell[c]{50.94\\[-1pt]{\scriptsize \red{$\downarrow$4.70}}} & \makecell[c]{31.31\\[-1pt]{\scriptsize \green{$\uparrow$18.06}}} & \makecell[c]{32.09\\[-1pt]{\scriptsize \green{$\uparrow$21.00}}} & \makecell[c]{29.78\\[-1pt]{\scriptsize \green{$\uparrow$12.29}}} & \makecell[c]{25.82\\[-1pt]{\scriptsize \red{$\downarrow$0.54}}} & \makecell[c]{25.74\\[-1pt]{\scriptsize \red{$\downarrow$0.85}}} & \makecell[c]{25.63\\[-1pt]{\scriptsize \red{$\downarrow$1.27}}} & \makecell[c]{51.74\\[-1pt]{\scriptsize \green{$\uparrow$2.27}}} & \makecell[c]{48.83\\[-1pt]{\scriptsize \red{$\downarrow$3.48}}} & \makecell[c]{48.74\\[-1pt]{\scriptsize \red{$\downarrow$3.66}}} & \makecell[c]{\underline{74.84}\\[-1pt]{\scriptsize \red{$\downarrow$0.12}}} & \makecell[c]{\underline{72.77}\\[-1pt]{\scriptsize \red{$\downarrow$2.88}}} & \makecell[c]{71.15\\[-1pt]{\scriptsize \red{$\downarrow$5.04}}} & \cellcolor{OverallShade} \makecell[c]{49.00\\[-1pt]{\scriptsize \green{$\uparrow$1.67}}} \\
\midrule
\multicolumn{20}{c}{{\normalsize\itshape GTokenLLM}} \\
GraphGPT & \makecell[c]{14.21\\[-1pt]{\scriptsize \red{$\downarrow$30.00}}} & \makecell[c]{9.41\\[-1pt]{\scriptsize \red{$\downarrow$53.65}}} & \makecell[c]{4.24\\[-1pt]{\scriptsize \red{$\downarrow$79.11}}} & \makecell[c]{27.43\\[-1pt]{\scriptsize \red{$\downarrow$15.86}}} & \makecell[c]{22.57\\[-1pt]{\scriptsize \red{$\downarrow$30.77}}} & \makecell[c]{17.71\\[-1pt]{\scriptsize \red{$\downarrow$45.67}}} & \makecell[c]{12.62\\[-1pt]{\scriptsize \red{$\downarrow$43.81}}} & \makecell[c]{9.17\\[-1pt]{\scriptsize \red{$\downarrow$59.17}}} & \makecell[c]{9.23\\[-1pt]{\scriptsize \red{$\downarrow$58.90}}} & \makecell[c]{2.96\\[-1pt]{\scriptsize \red{$\downarrow$52.10}}} & \makecell[c]{3.97\\[-1pt]{\scriptsize \red{$\downarrow$35.76}}} & \makecell[c]{4.60\\[-1pt]{\scriptsize \red{$\downarrow$25.57}}} & \makecell[c]{35.30\\[-1pt]{\scriptsize \green{$\uparrow$3.13}}} & \makecell[c]{35.17\\[-1pt]{\scriptsize \green{$\uparrow$2.75}}} & \makecell[c]{35.21\\[-1pt]{\scriptsize \green{$\uparrow$2.86}}} & \makecell[c]{49.10\\[-1pt]{\scriptsize \red{$\downarrow$5.21}}} & \makecell[c]{47.27\\[-1pt]{\scriptsize \red{$\downarrow$8.75}}} & \makecell[c]{46.01\\[-1pt]{\scriptsize \red{$\downarrow$11.18}}} & \cellcolor{OverallShade} \makecell[c]{21.45\\[-1pt]{\scriptsize \red{$\downarrow$33.24}}} \\
LLaGA & \makecell[c]{12.55\\[-1pt]{\scriptsize \red{$\downarrow$14.97}}} & \makecell[c]{14.58\\[-1pt]{\scriptsize \red{$\downarrow$1.22}}} & \makecell[c]{18.82\\[-1pt]{\scriptsize \green{$\uparrow$27.51}}} & \makecell[c]{14.73\\[-1pt]{\scriptsize \red{$\downarrow$16.07}}} & \makecell[c]{14.11\\[-1pt]{\scriptsize \red{$\downarrow$19.60}}} & \makecell[c]{12.54\\[-1pt]{\scriptsize \red{$\downarrow$28.55}}} & \makecell[c]{20.62\\[-1pt]{\scriptsize \red{$\downarrow$10.74}}} & \makecell[c]{19.06\\[-1pt]{\scriptsize \red{$\downarrow$17.49}}} & \makecell[c]{16.80\\[-1pt]{\scriptsize \red{$\downarrow$27.27}}} & \makecell[c]{16.08\\[-1pt]{\scriptsize \green{$\uparrow$2.36}}} & \makecell[c]{15.64\\[-1pt]{\scriptsize \red{$\downarrow$0.45}}} & \makecell[c]{15.31\\[-1pt]{\scriptsize \red{$\downarrow$2.55}}} & \makecell[c]{21.90\\[-1pt]{\scriptsize \green{$\uparrow$2.67}}} & \makecell[c]{29.22\\[-1pt]{\scriptsize \green{$\uparrow$36.99}}} & \makecell[c]{27.32\\[-1pt]{\scriptsize \green{$\uparrow$28.08}}} & \makecell[c]{44.98\\[-1pt]{\scriptsize \green{$\uparrow$8.36}}} & \makecell[c]{51.03\\[-1pt]{\scriptsize \green{$\uparrow$22.93}}} & \makecell[c]{51.96\\[-1pt]{\scriptsize \green{$\uparrow$25.17}}} & \cellcolor{OverallShade} \makecell[c]{23.18\\[-1pt]{\scriptsize \red{$\downarrow$1.48}}} \\
TEA-GLM & \makecell[c]{48.71\\[-1pt]{\scriptsize 0.00}} & \makecell[c]{50.37\\[-1pt]{\scriptsize \green{$\uparrow$3.41}}} & \makecell[c]{49.08\\[-1pt]{\scriptsize \green{$\uparrow$0.76}}} & \makecell[c]{53.13\\[-1pt]{\scriptsize \green{$\uparrow$0.28}}} & \makecell[c]{54.23\\[-1pt]{\scriptsize \green{$\uparrow$2.36}}} & \makecell[c]{53.61\\[-1pt]{\scriptsize \green{$\uparrow$1.19}}} & \makecell[c]{28.64\\[-1pt]{\scriptsize \red{$\downarrow$1.41}}} & \makecell[c]{27.70\\[-1pt]{\scriptsize \red{$\downarrow$4.65}}} & \makecell[c]{27.72\\[-1pt]{\scriptsize \red{$\downarrow$4.58}}} & \makecell[c]{19.11\\[-1pt]{\scriptsize \red{$\downarrow$2.05}}} & \makecell[c]{19.03\\[-1pt]{\scriptsize \red{$\downarrow$2.46}}} & \makecell[c]{18.95\\[-1pt]{\scriptsize \red{$\downarrow$2.87}}} & \makecell[c]{\underline{57.87}\\[-1pt]{\scriptsize \red{$\downarrow$5.73}}} & \makecell[c]{\underline{56.46}\\[-1pt]{\scriptsize \red{$\downarrow$8.03}}} & \makecell[c]{\underline{55.88}\\[-1pt]{\scriptsize \red{$\downarrow$8.98}}} & \makecell[c]{33.44\\[-1pt]{\scriptsize \green{$\uparrow$3.27}}} & \makecell[c]{34.14\\[-1pt]{\scriptsize \green{$\uparrow$5.44}}} & \makecell[c]{34.56\\[-1pt]{\scriptsize \green{$\uparrow$6.73}}} & \cellcolor{OverallShade} \makecell[c]{40.15\\[-1pt]{\scriptsize \red{$\downarrow$1.33}}} \\
GOFA & \makecell[c]{66.61\\[-1pt]{\scriptsize \red{$\downarrow$1.36}}} & \makecell[c]{63.84\\[-1pt]{\scriptsize \red{$\downarrow$5.46}}} & \makecell[c]{\underline{65.31}\\[-1pt]{\scriptsize \red{$\downarrow$3.29}}} & \makecell[c]{42.48\\[-1pt]{\scriptsize \red{$\downarrow$4.56}}} & \makecell[c]{42.63\\[-1pt]{\scriptsize \red{$\downarrow$4.22}}} & \makecell[c]{41.69\\[-1pt]{\scriptsize \red{$\downarrow$6.34}}} & \makecell[c]{25.54\\[-1pt]{\scriptsize \red{$\downarrow$9.91}}} & \makecell[c]{24.05\\[-1pt]{\scriptsize \red{$\downarrow$15.17}}} & \makecell[c]{21.99\\[-1pt]{\scriptsize \red{$\downarrow$22.43}}} & \makecell[c]{19.62\\[-1pt]{\scriptsize \red{$\downarrow$3.73}}} & \makecell[c]{19.32\\[-1pt]{\scriptsize \red{$\downarrow$5.20}}} & \makecell[c]{19.11\\[-1pt]{\scriptsize \red{$\downarrow$6.23}}} & \makecell[c]{38.34\\[-1pt]{\scriptsize \red{$\downarrow$3.13}}} & \makecell[c]{37.15\\[-1pt]{\scriptsize \red{$\downarrow$6.14}}} & \makecell[c]{37.42\\[-1pt]{\scriptsize \red{$\downarrow$5.46}}} & \makecell[c]{65.02\\[-1pt]{\scriptsize \red{$\downarrow$1.56}}} & \makecell[c]{64.67\\[-1pt]{\scriptsize \red{$\downarrow$2.09}}} & \makecell[c]{64.12\\[-1pt]{\scriptsize \red{$\downarrow$2.92}}} & \cellcolor{OverallShade} \makecell[c]{42.16\\[-1pt]{\scriptsize \red{$\downarrow$7.34}}} \\
UniGTE & \makecell[c]{37.45\\[-1pt]{\scriptsize \green{$\uparrow$1.99}}} & \makecell[c]{36.35\\[-1pt]{\scriptsize \red{$\downarrow$1.01}}} & \makecell[c]{34.13\\[-1pt]{\scriptsize \red{$\downarrow$7.05}}} & \makecell[c]{31.66\\[-1pt]{\scriptsize \red{$\downarrow$1.46}}} & \makecell[c]{31.35\\[-1pt]{\scriptsize \red{$\downarrow$2.43}}} & \makecell[c]{30.56\\[-1pt]{\scriptsize \red{$\downarrow$4.89}}} & \makecell[c]{16.27\\[-1pt]{\scriptsize \green{$\uparrow$23.44}}} & \makecell[c]{17.45\\[-1pt]{\scriptsize \green{$\uparrow$32.40}}} & \makecell[c]{17.58\\[-1pt]{\scriptsize \green{$\uparrow$33.38}}} & \makecell[c]{18.98\\[-1pt]{\scriptsize \green{$\uparrow$14.54}}} & \makecell[c]{19.07\\[-1pt]{\scriptsize \green{$\uparrow$15.09}}} & \makecell[c]{19.02\\[-1pt]{\scriptsize \green{$\uparrow$14.79}}} & \makecell[c]{42.84\\[-1pt]{\scriptsize \red{$\downarrow$7.77}}} & \makecell[c]{42.09\\[-1pt]{\scriptsize \red{$\downarrow$9.39}}} & \makecell[c]{40.15\\[-1pt]{\scriptsize \red{$\downarrow$13.56}}} & \makecell[c]{51.12\\[-1pt]{\scriptsize \green{$\uparrow$2.75}}} & \makecell[c]{51.65\\[-1pt]{\scriptsize \green{$\uparrow$3.82}}} & \makecell[c]{52.71\\[-1pt]{\scriptsize \green{$\uparrow$5.95}}} & \cellcolor{OverallShade} \makecell[c]{32.80\\[-1pt]{\scriptsize \green{$\uparrow$8.32}}} \\
\bottomrule
\end{tabular}%
}
\vskip -0.175in
\end{table*}

\subsection{Robustness against Textual Attacks}
\noindent \textbf{\ding{176} Textual attacks are more effective against GTextLLMs and GTokenLLMs than GNNPreds.} As reported in Table~\ref{tab:textual_attack}, GTextLLMs and GTokenLLMs suffer average relative accuracy drops of 21.19\% and 20.27\%, respectively, compared with 8.28\% for GNNPreds. 
This is because LLM-based models mainly rely on node text for prediction, whereas GNNPreds can use the unperturbed neighborhood and graph structure to mitigate the effect of corrupted node text.


\noindent \textbf{\ding{177} GTextLLMs equipped with stronger LLM backbones are more vulnerable to textual attacks.}
Llama3.1-8B, Qwen3-8B, and G1 all suffer average relative accuracy drops of over 22\%, compared with 16.09\% for Vicuna-7B (Table~\ref{tab:textual_attack}). One possible reason is that LLMs with stronger text understanding rely more on textual semantics, which also makes them more sensitive to semantically misleading text.

\noindent \textbf{\ding{178} GTokenLLMs are more vulnerable to textual attacks than structural attacks.}
Their average relative performance drop increases from 2.45\% under structural attacks to 20.27\% under textual attacks (Table~\ref{tab:textual_attack}). 
One possible explanation is that the inputs to GTokenLLMs typically include the central-node text. Therefore, textual perturbations not only corrupt the textual information directly provided to the LLM, but also affect the graph tokens generated from the perturbed node text. In contrast, structural perturbations mainly affect the graph-token representations while preserving the clean central-node text, which remains available for prediction.




\begin{figure}[t]
  \centering
  \includegraphics[width=0.97\linewidth]{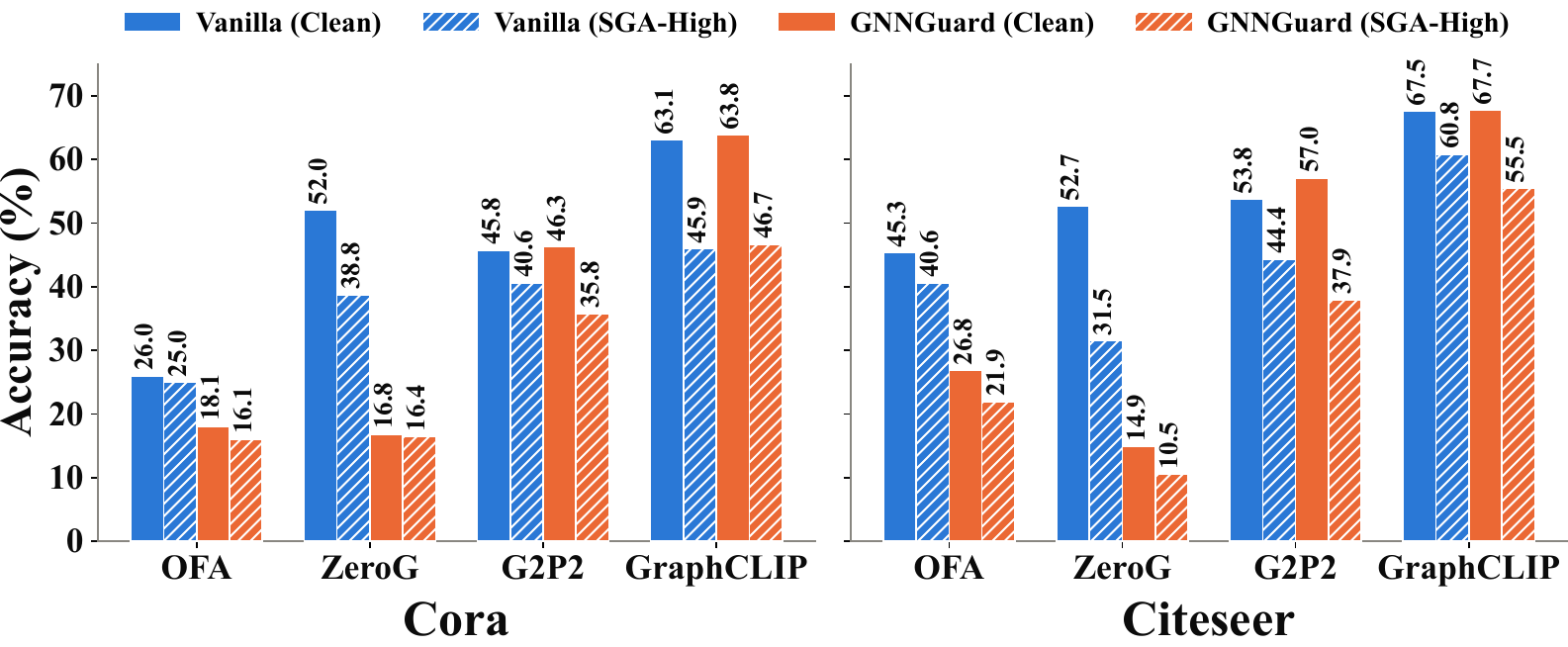}
  \vskip -0.125in
  \caption{Effectiveness of model-centric defenses that replace the Vanilla GNN with GNNGuard under structural attacks.}
  \vskip -0.225in
  \label{fig:model_centric_defense}
\end{figure}

\begin{figure}[t]
  \centering
  \includegraphics[width=\linewidth]{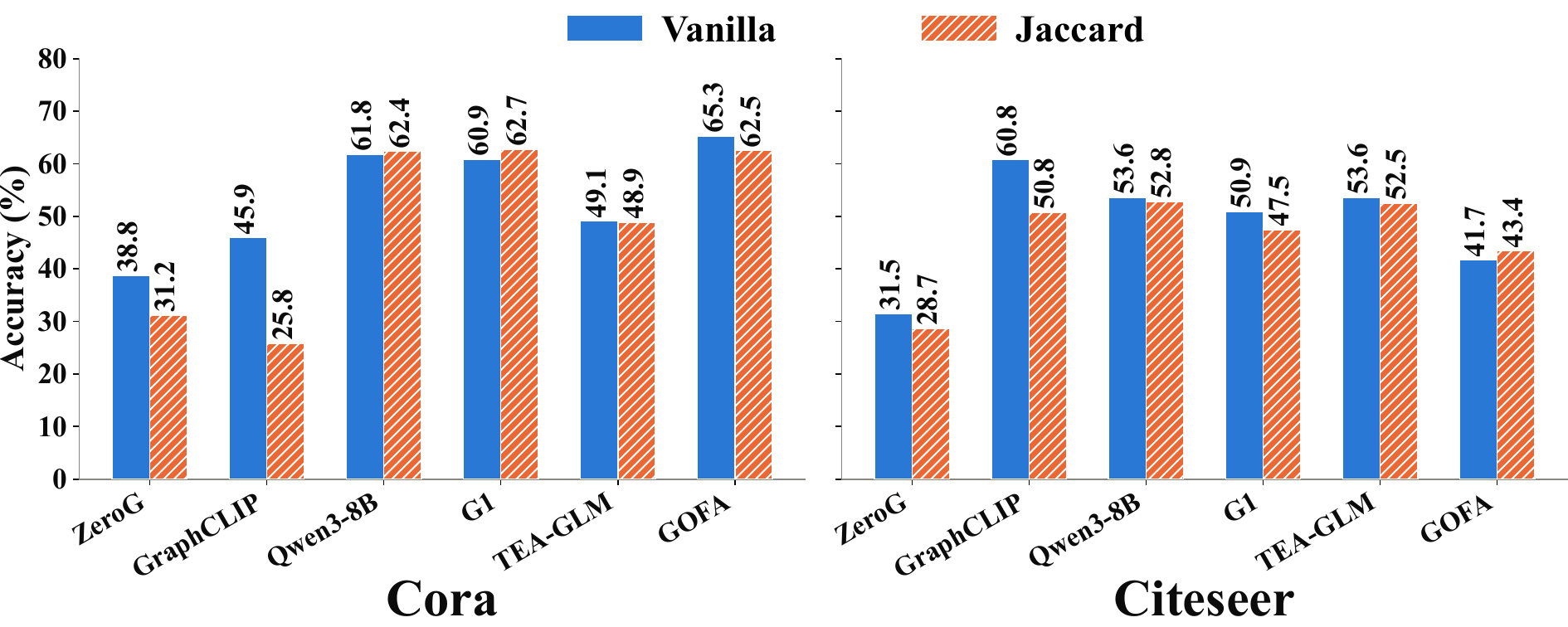}
  \vskip -0.125in
  \caption{Effectiveness of data-centric defenses that purify graph structures perturbed by SGA-High.}
  \label{fig:data_centric_defense}
  \vskip -0.225in
\end{figure}
\subsection{Robustness against Node Injection Attacks}
\textbf{\ding{179} Node injection attacks show a similar vulnerability pattern to structural attacks.} As shown in Table~\ref{tab:hybrid_attack}, GNNPreds, GTextLLMs, and GTokenLLMs exhibit average relative accuracy drops of 19.22\%, 0.69\%, and 7.01\% under node injection attacks, respectively, which are closer to those under structural attacks (18.37\%, 3.10\%, and 2.45\%) than to those under textual attacks (8.28\%, 21.19\%, and 20.27\%). A possible explanation is that node injection perturbs neighborhood information while preserving the central-node text. It therefore strongly affects GNNPreds, has little effect on GTextLLMs, and moderately affects GTokenLLMs that use graph tokens as auxiliary context.

\subsection{Effectiveness of Existing Graph Defenses}
To evaluate existing defenses in zero-shot settings, we consider model-centric and data-centric structural defenses on Cora and Citeseer under SGA-High. The former replaces the original GNN backbone with robust GNNGuard~\cite{zhang2020gnnguard}, while the latter applies Jaccard~\cite{wu2019adversarial} to purify the attacked structure with a threshold of 0.5.

\noindent \textbf{\ding{180} Existing graph defenses do not consistently improve zero-shot adversarial robustness and may even degrade model performance.} 
Replacing the original GNNs with GNNGuard generally fails to improve robustness and substantially reduces the clean performance of GNNPreds (Figure~\ref{fig:model_centric_defense}). Similarly, Jaccard provides no consistent improvement and even degrades model performance (Figure~\ref{fig:data_centric_defense}). One possible explanation is that both defenses identify suspicious edges using fixed feature-similarity rules, which may not transfer across unseen target graphs with distinct similarity patterns.

\subsection{Summary and Future Work}
Based on the above experimental observations, we summarize 3 key conclusions:
(1) Despite promising performance on clean graphs, all ZGMs suffer substantial degradation under adversarial attacks. GNNPreds are more vulnerable to structural and node injection attacks due to their reliance on neighborhood aggregation, whereas GTextLLMs and GTokenLLMs are more vulnerable to textual attacks because their predictions depend more heavily on node text.
(2) Stronger LLM backbones improve robustness to corrupted structural context but increase sensitivity to misleading textual semantics, revealing a structure--text robustness trade-off.
(3) Existing graph defenses do not consistently improve robustness and may even compromise clean zero-shot performance. 

These observations raise alarms about the secure deployment of ZGMs in real-world applications and point to several future directions:
For GNNPreds, robustness should be improved at the neighborhood aggregation stage by reducing the influence of adversarial edges and injected nodes. For LLM-based models, future methods should verify textual evidence against graph context to reduce their sensitivity to misleading semantics. 
Moreover, the observed structure--text trade-off calls for methods that dynamically balance structural and textual evidence.
Finally, since existing supervised defenses may remove information essential for zero-shot transfer, 
new defenses should identify adversarial perturbations without target labels while preserving clean zero-shot performance.

\vskip -0.09in
\section{Conclusion}
\vskip -0.01in
In this work, we introduced ZeroGAR, the first systematic benchmark for evaluating the adversarial robustness of zero-shot graph models on unseen target graphs.
ZeroGAR provides a unified evaluation across 13 representative models, diverse text-attributed graphs, in-domain and cross-domain transfer settings, and structural, textual, and node injection attacks, while also examining existing graph defenses.
Our experiments show that strong clean zero-shot performance does not guarantee adversarial robustness. GNN-based methods are more vulnerable to structural and node injection attacks, whereas LLM-based methods are more vulnerable to textual attacks;
stronger LLMs introduce a structure–text robustness trade-off; and existing graph defenses do not consistently improve zero-shot robustness and may compromise clean performance.
We hope ZeroGAR will provide a standardized testbed for developing robustness methods specifically designed for zero-shot deployment on unseen graphs.


\bibliography{aaai2027}
\clearpage
\section{Related Work}
\subsection{Zero-Shot Graph Models}
Based on their prediction mechanism, existing zero-shot graph models can be broadly divided into 3 categories: GNNPreds, GTextLLMs, and GTokenLLMs.
GNNPreds train a TAG encoder on source graphs, to encode the textual and structural information of target nodes, while an off-the-shelf text encoder encodes target class descriptions.
For zero-shot inference, G2P2~\cite{wen2023augmenting}, ZeroG~\cite{li2024zerog}, and GraphCLIP~\cite{zhu2025graphclip} compute contrastive similarities between the representation of each node and those of the candidate classes, whereas OFA~\cite{liu2024one} concatenates their representations and trains a set of binary classifiers to determine whether the node belongs to each candidate class.
Unlike GNNPreds, GTextLLMs and GTokenLLMs use GPT-style LLMs as predictors, but differ in how they provide graph information to the LLM~\cite{yu2026graph2text, zhang2026revisiting}.
GTextLLMs convert node attributes and graph structure into textual descriptions and prompt an LLM to predict node labels~\cite{chen2024exploring, tan2026gabench, yan2025data}.
To better model graph structure, existing methods use structured representations such as syntax trees~\cite{zhao2023graphtext}, random walks~\cite{tan2024walklm}, and code-like formats~\cite{wang2024instructgraph}.
Some studies further fine-tune open-source LLMs to improve their performance on graph tasks~\cite{wang2026exploring}.
In contrast, GTokenLLMs encode node text and graph structure into graph tokens, project them into the input space of an LLM, and combine them with textual task instructions for prediction~\cite{zhang2026toward}.
This design provides a more compact way to pass graph information to LLMs. Representative methods include LLaGA~\cite{chenllaga}, TEA-GLM~\cite{wang2024llms}, and GOFA~\cite{konggofa}.
Although these models achieve promising zero-shot performance on clean graphs, their robustness against structure, text, and node injection attacks remains unexplored.

\subsection{Graph Robustness Benchmark}
Existing studies have evaluated graph robustness across an increasingly broad range of models and attacks. For example, GRB~\cite{zheng2graph} provides a unified benchmark for evaluating GNNs and robust GNNs under graph injection attacks.  Guo et al.~\cite{guo2024learning} and LLM4RGNN~\cite{DBLP:conf/kdd/Zhang0Z0Z0025} extend the evaluation to LLM-enhanced graph learning, considering both LLMs as feature enhancers and predictors under structural and textual perturbations. 
Guo et al.~\cite{ma2026llm} further investigate the poisoning robustness of LLM-enhanced GNNs, evaluating different language model encoders and GNN backbones under structural and textual attacks.
TrustGLM~\cite{zhang2025trustglm} evaluates GraphLLMs against prompt, textual, and structural attacks, while Olatunji et al.~\cite{olatunji2025adversarial} further examine graph-aware LLMs under poisoning and evasion attacks. More recently, Lei et al.~\cite{lei2026robustness} compare GNNs, robust GNNs, and GraphLLMs under textual and structural attacks. However, existing evaluations mainly focus on supervised settings, where models are trained and evaluated on the same graph or task. They therefore provide limited insight into whether ZGMs remain robust when transferred to unseen target graphs without target-domain supervision. 

\section{Additional Details on ZeroGAR}

\subsection{Datasets}
All datasets used in ZeroGAR are publicly available and cover multiple domains. For each dataset, we store the graph data in PyTorch .pt format, including sentence bert embeddings commonly used in existing methods, original node texts, edge indices, node labels, label names, and train, validation, and test masks. Unless otherwise specified, all datasets are released under the MIT license. Detailed descriptions of these datasets are provided below.

\noindent $\bullet$ \textbf{Cora.}
The Cora dataset is a citation network in the computer science domain, where nodes represent research papers and edges denote citation relationships between papers. The raw text attributes are collected from the GitHub repository provided in Chen et al.~\cite{chen2024exploring}. Each node contains the paper title and abstract as its text feature, and the node label corresponds to the paper category.

\noindent $\bullet$ \textbf{Citeseer.}
The Citeseer dataset is a citation network in the computer science domain. Each node represents a research paper, and each edge indicates a citation relationship between two papers. The TAG version contains text attributes for 3,186 nodes, where the raw texts are collected from the GitHub repository provided in Chen et al.~\cite{chen2024exploring}.

\noindent $\bullet$ \textbf{OGBN-Products.} OGBN-Products~\cite{hu2020open} contains over 2 million nodes and 61 million edges. Each node represents an Amazon product, and each edge indicates a co-purchase relationship between two products. The task is to classify each product into one of 47 top-level categories.

\noindent $\bullet$ \textbf{Ogbn-Arxiv.}
The Ogbn-Arxiv dataset is a citation network collected from the arXiv platform, where nodes represent papers and edges denote citation relationships. The raw text attributes are collected from the GitHub repository provided in OFA~\cite{liu2024one}.

\noindent $\bullet$ \textbf{Child.} The Child dataset~\cite{yan2023comprehensive} is an Amazon co-purchase and co-view graph of children’s books. Each node represents a book, and each edge connects two books that are frequently purchased or viewed together. Each node is associated with textual attributes, including the book title and descriptive metadata.

\noindent $\bullet$ \textbf{History.}
The History dataset~\cite{yan2023comprehensive} is a subset of Amazon-Books containing books in the ``History'' category. Each node represents a book, and each edge connects two books that are frequently purchased or viewed together. The title and description of each book serve as its textual attributes. Each node is assigned one of 12 labels derived from the three-level book classification system.
%

\noindent $\bullet$ \textbf{WikiCS.}
The WikiCS dataset is an internet link network, where nodes represent Wikipedia pages and edges denote reference links between pages. The raw text attributes are collected from OFA~\cite{liu2024one}, including the page name and content of each Wikipedia entry. Each node is labeled according to the category of the corresponding entry.

\noindent $\bullet$ \textbf{Instagram.}
The Instagram is a social network, where nodes represent users and edges indicate following relationships. The raw text attributes are collected from GraphAdapter~\cite{huang2024can}, consisting of user personal introductions. Each node is labeled as either commercial or normal.

\subsection{Details of Victim Models} \label{sec:model_detail}
In ZeroGAR, we benchmark 13 representative ZGMs, comprising four GNN-based Predictors, four Graph-Textualizing LLMs, and five Graph-Tokenizing LLMs. Detailed descriptions of these benchmarked methods are provided below.

\noindent $\bullet$ \textbf{GNN-based Predictors}

\begin{itemize}
    \setlength{\itemsep}{0.3em}
    \setlength{\parskip}{0pt}

    \item[--] \textbf{G2P2}~\cite{wen2023augmenting} jointly pre-trains graph and text encoders using graph-grounded contrastive objectives over node texts, node representations, and neighborhood summaries. During inference, it compares node representations with prompted class descriptions to perform zero-shot node classification.

    \item[--] \textbf{ZeroG}~\cite{li2024zerog} uses a language model to encode node attributes and class descriptions into a shared semantic space. It combines prompt-based subgraph sampling with lightweight fine-tuning and predicts unseen classes by measuring the similarity between node and class representations.

    \item[--] \textbf{GraphCLIP}~\cite{zhu2025graphclip} constructs graph-summary pairs with an LLM and contrastively aligns graph representations with summary embeddings. It further introduces invariant learning to improve cross-domain transfer and performs zero-shot prediction by matching target subgraphs with textual class descriptions.

    \item[--] \textbf{OFA}~\cite{liu2024one} represents nodes, edges, and task descriptions as natural-language text and encodes them into a unified feature space. It introduces nodes of interest and graph prompting substructures to support node-, link-, and graph-level classification across domains.
\end{itemize}

\noindent $\bullet$ \textbf{Graph-Textualizing LLMs}

\begin{itemize}
    \setlength{\itemsep}{0.3em}
    \setlength{\parskip}{0pt}

    \item[--] \textbf{Llama3.1-8B}~\cite{grattafiori2024llama} directly processes the target node and its local neighborhood after they are converted into a natural-language prompt. Without graph-specific training, it predicts the node label from candidate classes using its pretrained semantic knowledge and instruction-following ability.

    \item[--] \textbf{Qwen3-8B}~\cite{yang2025qwen3} follows the same graph-textualization setting, where node attributes and neighborhood information are represented in natural language. It directly reasons over the resulting prompt and selects the predicted label from the candidate classes without target-graph adaptation.

    \item[--] \textbf{Vicuna-7B}~\cite{chiang2023vicuna} receives textual descriptions of the target node and its neighborhood under the same prompt template. It serves as an instruction-tuned LLM baseline that directly produces node predictions without explicitly encoding the graph into specialized representations.

    \item[--] \textbf{G1}~\cite{guo2025g1} improves the graph reasoning ability of an LLM through reinforcement learning on synthetic graph-theoretic tasks with verifiable rewards. It directly reasons over textualized graph inputs and transfers the learned reasoning ability to unseen real-world node classification tasks.
\end{itemize}

\noindent $\bullet$ \textbf{Graph-Tokenizing LLMs}

\begin{itemize}
    \setlength{\itemsep}{0.3em}
    \setlength{\parskip}{0pt}

    \item[--] \textbf{GraphGPT}~\cite{tang2024graphgpt} first aligns a graph encoder with language semantics through text-graph grounding and then maps graph representations into the LLM token space using a projector. Its two-stage graph instruction tuning enables natural-language prediction across graph datasets and tasks.

    \item[--] \textbf{LLaGA}~\cite{chenllaga} reorganizes local graph structures into structure-aware node sequences using node-level templates and maps these sequences into the LLM token space through a versatile projector. This design enables the LLM to generalize across previously unseen graph datasets and tasks.

    \item[--] \textbf{TEA-GLM}~\cite{wang2024llms} pre-trains a GNN and aligns its representations with the LLM token embedding space. A lightweight linear projector converts each graph representation into a fixed number of graph tokens, which a frozen LLM processes under unified instructions for zero-shot prediction.

    \item[--] \textbf{GOFA}~\cite{konggofa} interleaves trainable GNN layers with a frozen pre-trained LLM to jointly model graph structure and textual semantics. It is pre-trained with graph-level language modeling, question-answering, and structural tasks, enabling generative zero-shot inference across different graph tasks.

    \item[--] \textbf{UniGTE}~\cite{wang2026unigte} adopts an instruction-tuned encoder--decoder architecture that jointly processes graph tokens and natural-language task prompts. Learnable alignment tokens produce compact task-aware graph representations, while a frozen LLM decoder predicts task answers and reconstructs graph prompts to preserve structural and semantic information.
\end{itemize}

\section{Additional Details on Implementations} \label{app:model_settings}
\subsection{Hyperparameter Settings}
All implementations strictly follow the settings in the original papers and official repositories. We use a unified random seed of 42 for all experiments. The implementation details are provided below:

\noindent $\bullet$ \textbf{Structure attack}
  We use SGA as the structural attack. The attack is a direct targeted evasion attack with structure perturbation enabled and feature perturbation disabled. The SGA surrogate is an SGC model with $K=3$ and learning rate $0.01$; the auxiliary GCN used for evasion evaluation has hidden size 64, dropout 0.5, and early-stopping patience 30. 

  \noindent $\bullet$ \textbf{Textual attack.} We evaluate an LLM-based textual rewriting attack using Qwen3.6-27B; API generation uses temperature 0.7 and max tokens 1000.

  \noindent $\bullet$ \textbf{Node injection attack}
  We use WTGIA with the ATDGIA injection strategy. Injected node features are initialized as zeros and optimized with FGSM-style binary feature flipping. We use sparsity level 0.15, sequential injection step 0.2, epsilon 0.01, seed 42, and BoW features. The feature-flipping batch size is 1 for Cora/Citeseer and 50 for the other datasets.

\noindent $\bullet$ \textbf{G2P2}: We use SBERT-384 node features and train for 2 epochs with batch size 64 and learning rate $2\times10^{-5}$. The graph encoder has input dimension 384, hidden dimension 128, and output dimension 128. The text encoder uses context length 128, transformer width 512, 12 layers, and 8 attention heads. For OGBN-Arxiv, the edge coefficient is 0.1 and full-graph training is used. For OGBN-Products and the Arxiv+Products setting, the edge coefficient is 10 and 2-hop neighbor sampling with fanout $[25,10]$ is used.

\noindent $\bullet$ \textbf{ZeroG}: We train ZeroG for 50 epochs using SentenceBERT textual features, learning rate $10^{-4}$, weight decay 0.1, gradient accumulation steps 4, and gradient clipping 1.0. The feature propagation round is $R=10$, normalization is enabled, and the maximum sampled subgraph size is 100 nodes. Checkpoints are saved after each epoch.

\noindent $\bullet$ \textbf{GraphCLIP}: We use BERT-384 graph inputs, a GPS graph encoder with hidden/output dimension 1024 and 12 layers, and a frozen MiniLM text encoder. Training uses AdamW with learning rate $10^{-5}$, weight decay $10^{-5}$, batch size 1024, and 30 epochs. Graph augmentation drops 30\% of feature dimensions and 20\% of edges. The adversarial augmentation step size is $10^{-2}$ with 3 update steps. The contrastive logit scale is initialized as $1/0.07$.

\noindent $\bullet$ \textbf{OFA}: We use SBERT-384 node features projected to hidden dimension 768. The OFA model uses 7 GNN layers, dropout 0.15, JK set to ``none'', MiniLM textual embeddings, batch size 512, learning rate $10^{-3}$, validation interval 50, and 50 training epochs. The same settings are used for Arxiv, Products, and Arxiv+Products source configurations.

\noindent $\bullet$ \textbf{Llama3.1-8B}: Evaluated with vLLM using the 2-hop without-label prompt format. The prompt includes full target-node text, at most 20 hop-1 neighbors, and at most 5 hop-2 neighbors. Generation uses temperature 0.6, top-$p$
0.95, top-$k$ 30, max new tokens 4096, and seed 42.

\noindent $\bullet$ \textbf{Qwen3-8B}: Uses the same vLLM inference settings as Llama3.1-8B. Thinking is disabled by default.

\noindent $\bullet$ \textbf{Vicuna-7B}: Uses the same sampling settings. Since Vicuna has no built-in chat template in the current code, the evaluation falls back to the Vicuna single-turn chat format. The effective context length is treated as 16K.

\noindent $\bullet$ \textbf{G1}: We evaluate G1-7B with the same 2-hop without-label prompts and the same vLLM sampling settings: temperature 0.6, top-$p$ 0.95, top-$k$ 30, max new tokens 4096, and seed 42.

\noindent $\bullet$ \textbf{GraphGPT}: We use Vicuna-7B-v1.5-16K as the base LLM. The per-GPU batch size is set to 2, and the learning rate is $2\times10^{-3}$. We use a warmup ratio of $3\times10^{-2}$ and set the maximum LLM input length to 2048. The model is trained for 3 epochs.

\noindent $\bullet$ \textbf{LLaGA}: We use Vicuna-7B-v1.5-16K as the frozen language backbone and SBERT graph embeddings. The model uses the ND template, 2-hop graph context, neighbor sample size 10, and a linear graph projector by default. Only the
multimodal projector is tuned. Training uses bf16/tf32, model max length 4096, 1 epoch, per-device batch size 4, gradient accumulation 1, learning rate $2\times10^{-3}$, weight decay 0, cosine scheduler, warmup ratio 0.03. Inference uses temperature 0.2, and max new tokens 1024.

\noindent $\bullet$ \textbf{TEA-GLM}: TEA-GLM uses a two-stage training pipeline. Stage 1 pretrains the GNN for 60 epochs with learning rate 0.02, batch size 512, hidden size 2048, output size 4096, 2 layers, fanout $[25,10]$, edge-drop rates 0.3/0.4, feature-drop rates 0.0/0.1, and temperature $\tau=0.4$. Stage 2 freezes the Vicuna-7B-v1.5-16K backbone and aligns the graph encoder for 1 epoch with batch size 2, gradient
steps 1, learning rate $10^{-3}$, weight decay 0, max text length 700, 5 graph tokens, attention dimension 2048, and GNN output dimension 4096. Inference uses batch size 4 and max text length 700.

\noindent $\bullet$ \textbf{GOFA}: We use Mistral-7B-Instruct-v0.2 with decoder LoRA enabled. Fine-tuning runs for 1 epoch with batch size 1, gradient accumulation 64, learning rate $10^{-4}$, weight decay 0.1, gradient clipping 0.5, LLM max length 256, and 6 GNN layers. The graph context uses 3 hops with at most 5 nodes per hop. Node-classification training uses 10-way sampled choices, and link prediction uses binary choices.

\noindent $\bullet$ \textbf{UniGTE}: We use Vicuna-7B-v1.5-16K as both graph-text encoder and LLM backbone, with 64 memory tokens and node dimension 768. PEFT/LoRA is enabled with rank 128, LoRA alpha 32, dropout 0.05, and target modules \texttt{q\_proj} and \texttt{v\_proj}. Training uses bf16, 1 epoch, per-device batch size 2, gradient accumulation 2, base/projector learning rate $2\times10^{-4}$, graph learning rate $2\times10^{-3}$, cosine scheduler, warmup ratio 0.03, max gradient norm 10. Evaluation uses per-device batch size 1; generation uses temperature 0.7 and max new tokens 80.






















\subsection{Prompt Template for the Text-based Attack} \label{subsec:text_attack}
In ZeroGAR, we adopt an LLM-based attack~\cite{lei2026robustness} to generate fluent text perturbations. The detailed prompt template is provided below:
\begin{figure}[h]
\vskip -0.1in
\begin{tcolorbox}[top=0pt, bottom=2pt, left=2pt, right=2pt, halign title=center, title={Prompt Template (instantiated per target node)}]

\textbf{Graph node classification task}\\
Available classes: \{classes\_str\}\\
Target node \{node\_id\}:\\
Original text: ``\{text\}''\\
Original label: ``\{current\_label\}''\\
\{neighbor\_info\} \quad \textit{(e.g., ``Neighbor labels: [...] (counts: \{...\})'' or ``No neighbors found'')}\\

\textbf{Task.} Rewrite the text to be as different as possible from the original while keeping a similar length.\\

\textbf{Requirements:}
\begin{itemize}
    \item Must \textit{not} belong to the original class: ``\{current\_label\}''.
    
    \item Should \textit{not} belong to neighbor classes: \{unique\_neighbor\_labels\} or None.
    
    \item \{target\_instruction\} \textit{(e.g., prefer a class from the allowed set, or the least frequent neighbor class if all are forbidden)}.
    
    \item Make the content maximally dissimilar from the original semantics.
    
    \item Keep the word count roughly similar.
    
    \item Produce content that is most unlikely under the node/neighbor context for the target class.
\end{itemize}

Goal: create text that is jointly inconsistent with the original content and its local graph context.\\
Return \textit{only} the modified text, with no explanations or notes.\\

\textbf{Modified text:}

\end{tcolorbox}
\vskip -0.15in
\end{figure}

\subsection{Text Prompt for GTextLLM}
For all GTextLLMs, we adopt the \textit{2-hop w/o Label} prompt template introduced by Wang et al.~\cite{wang2026exploring}, as shown below:
\begin{figure}[h]
\vskip -0.1in
\begin{tcolorbox}[
    top=2pt,
    bottom=2pt,
    left=2pt,
    right=2pt,
    halign title=center,
    title={Prompt Template for 2-hop w/o Label (instantiated per target node)}
]

\textbf{``Context''}: ``You are a good graph reasoner. Give you a graph language that describes a graph structure and node information from the \{dataset\_name\} dataset. You need to understand the graph and the task definition and answer the question. 
(\texttt{<Target node>}, \texttt{<Node attributes>}), 
(\texttt{<1-hop neighbors>}, \texttt{<Node attributes>}), 
(\texttt{<2-hop neighbors>}, \texttt{<Node attributes>}).''\\[2pt]

\textbf{``Question''}: ``Please predict the most appropriate category for the Target node. Choose from the following categories: \texttt{<Categories>}. Do not provide your reasoning. Answer:''\\[2pt]

\textbf{``Answer''}: ``\texttt{<Correct answer>}''\\[4pt]

\texttt{(<Target node>, <Node attributes>)}:\\
\#\# Target node:\\
Paper id: \{target\_node\_id\}\\
Title: \{target\_node\_text\}\\[3pt]

\texttt{(<1-hop neighbors>, <Node attributes>)}:\\
Known neighbor papers at hop 1 (partial, may be incomplete):\\
\{one\_hop\_neighbor\_information\}\\[3pt]

\texttt{(<2-hop neighbors>, <Node attributes>)}:\\
Known neighbor papers at hop 2 (partial, may be incomplete):\\
\{two\_hop\_neighbor\_information\}\\[3pt]

\texttt{<Categories>}:\\
\{categories\}\\[3pt]

\texttt{<Correct answer>}: \{correct\_answer\}

\end{tcolorbox}
\vskip -0.15in
\end{figure}

\subsection{Computing Environment and Resources}
The publicly available source code is provided in the supplementary material to support future research on the adversarial robustness of zero-shot graph models. ZeroGAR is implemented using PyTorch Geometric (PyG). All experiments are conducted in the following computing environment:
\begin{itemize}
\item \textbf{OS}: Ubuntu Linux 5.15.0-102-generic;
\item \textbf{CPU}: Intel(R) Xeon(R) Platinum 8358 CPU @ 2.60GHz;
\item \textbf{GPU}: NVIDIA A800 80GB.
\end{itemize}

\section{Additional Evaluation Results for ZeroGAR}

\subsection{Results for the ASR Metric} \label{app:asr_results}
We also report the attack success rate (ASR), defined as:
\begin{align}
\mathrm{ASR}
&=
\frac{
\sum_{v_i\in\mathcal{V}_{\mathrm{test}}}
\mathbb{I}
\left(
\hat{y}^{\mathrm{clean}}i=y_i
\land
\hat{y}^{\mathrm{adv}}i\neq y_i
\right)
}{
\sum{v_i\in\mathcal{V}{\mathrm{test}}}
\mathbb{I}
\left(
\hat{y}^{\mathrm{clean}}i=y_i
\right)
},
\label{eq:asr}
\end{align}
where $\mathcal{V}{\mathrm{test}}$ denotes the set of test nodes, $y_i$ is the ground-truth label of node $v_i$, and $\hat{y}^{\mathrm{clean}}_i$ and $\hat{y}^{\mathrm{adv}}_i$ denote its predictions on the clean and adversarial graphs, respectively. $\mathbb{I}(\cdot)$ is the indicator function. ASR measures the proportion of initially correctly classified test nodes that are misclassified after the attack. The ASR results for structural, textual, and node injection attacks are reported in Table~\ref{tab:structure_asr}, Table~\ref{tab:textual_asr}, and Table~\ref{tab:hybrid_asr}, respectively. The results are consistent with the observations reported in the main paper.

\begin{table*}[t]
\centering
\caption{Attack success rate (ASR, \%) under structural attacks across different perturbation sizes. Lower values indicate stronger robustness. Best results are bolded and runner-up results are underlined.}
\vskip -0.125in
\label{tab:structure_asr}
\definecolor{OverallShade}{RGB}{255, 245, 220}
\providecommand{\datasetheader}[1]{\raisebox{0.4ex}{\textbf{#1}}}
\providecommand{\levelheader}[1]{\raisebox{0.3ex}{\textbf{#1}}}
\setlength{\tabcolsep}{2pt}
\scriptsize
\resizebox{\textwidth}{!}{%
\begin{tabular}{l|ccc|ccc|ccc|ccc|ccc|ccc|c}
\toprule
\multirow{2}{*}{\textbf{Method}} & \multicolumn{3}{c}{\datasetheader{Cora}} & \multicolumn{3}{c|}{\datasetheader{Citeseer}} & \multicolumn{3}{c}{\datasetheader{History}} & \multicolumn{3}{c|}{\datasetheader{Child}} & \multicolumn{3}{c}{\datasetheader{Instagram}} & \multicolumn{3}{c|}{\datasetheader{WikiCS}} & \multirow{2}{*}{\textbf{Avg.}} \\[-3pt]
\cmidrule(lr){2-4}\cmidrule(lr){5-7}\cmidrule(lr){8-10}\cmidrule(lr){11-13}\cmidrule(lr){14-16}\cmidrule(lr){17-19}
 & \levelheader{Low} & \levelheader{Mid} & \levelheader{High}
 & \levelheader{Low} & \levelheader{Mid} & \levelheader{High}
 & \levelheader{Low} & \levelheader{Mid} & \levelheader{High}
 & \levelheader{Low} & \levelheader{Mid} & \levelheader{High}
 & \levelheader{Low} & \levelheader{Mid} & \levelheader{High}
 & \levelheader{Low} & \levelheader{Mid} & \levelheader{High} & \\[-2pt]
\midrule
\multicolumn{20}{c}{{\normalsize\itshape GNNPred}} \\
OFA       & 7.80  & 11.35 & 12.77 & 11.66 & 15.19 & 17.31 & 19.68 & \underline{17.87} & 22.09 & 38.10 & 45.24 & 45.24 & \underline{4.00} & 7.67 & 13.22 & 13.49 & 16.08 & 18.67 & \cellcolor{OverallShade} 18.75 \\
G2P2      & 6.45  & 20.56 & 27.82 & 13.39 & 27.38 & 30.95 & \underline{17.94} & 20.70 & 23.92 & 16.45 & 23.18 & 27.35 & 7.54 & 7.78 & 12.44 & \textbf{3.15} & \underline{5.75} & \underline{8.23} & \cellcolor{OverallShade} 16.72 \\
ZeroG     & 16.67 & 41.49 & 59.22 & 12.95 & 23.80 & 33.13 & 46.97 & 57.06 & 69.64 & 40.97 & 52.02 & 58.68 & 39.63 & 52.35 & 76.13 & 12.63 & 27.63 & 36.57 & \cellcolor{OverallShade} 42.09 \\
GraphCLIP & 15.79 & 27.19 & 31.09 & 8.64  & 8.88  & 13.08 & 39.24 & 42.77 & 51.17 & 32.45 & 35.35 & 36.30 & \textbf{1.61} & \textbf{1.61} & \textbf{1.53} & 16.90 & 22.60 & 25.98 & \cellcolor{OverallShade} 22.90 \\
\midrule
\multicolumn{20}{c}{{\normalsize\itshape GTextLLM}} \\
Llama3.1-8B & 10.67 & 14.89 & 15.21 & 16.32 & 19.43 & 20.98 & 31.56 & 30.41 & 33.01 & 29.68 & 33.22 & 33.83 & 27.26 & 26.43 & 25.12 & 12.38 & 14.82 & 16.98 & \cellcolor{OverallShade} 22.90 \\
Qwen3-8B    & 8.40  & 10.30 & 14.40 & 12.09 & 13.74 & 18.96 & 28.49 & 28.69 & 30.74 & 22.73 & 25.66 & 28.05 & 17.68 & 19.20 & 18.03 & 5.94  & 7.63  & 9.08  & \cellcolor{OverallShade} 17.77 \\
Vicuna-7B   & 19.35 & 30.32 & 31.39 & 26.76 & 28.17 & 32.75 & 47.33 & 49.75 & 54.58 & 50.87 & 52.42 & 53.74 & 59.40 & 57.55 & 61.41 & 43.88 & 49.96 & 50.79 & \cellcolor{OverallShade} 44.47 \\
G1          & 8.63  & 14.56 & 18.92 & 14.55 & 14.24 & 22.73 & 28.99 & 28.09 & 30.85 & 24.98 & 25.38 & 28.36 & 24.56 & 25.08 & 25.39 & 8.62  & 10.82 & 12.70 & \cellcolor{OverallShade} 20.41 \\
\midrule
\multicolumn{20}{c}{{\normalsize\itshape GTokenLLM}} \\
GraphGPT & 26.36 & 33.64 & 47.71 & 16.67 & 28.79 & 35.86 & 39.21 & 53.51 & 72.90 & 57.37 & 63.37 & 66.74 & 8.90  & 10.21 & 9.42  & \underline{3.93} & \textbf{3.79} & \textbf{4.48} & \cellcolor{OverallShade} 32.38 \\
LLaGA    & 22.50 & 25.00 & 25.00 & 38.46 & 27.88 & 25.96 & 23.80 & 27.45 & 33.49 & 33.17 & 36.11 & 37.52 & 49.85 & 48.35 & 43.84 & 14.74 & 16.85 & 17.57 & \cellcolor{OverallShade} 30.42 \\
TEA-GLM  & \textbf{2.27} & \textbf{3.03} & \textbf{3.03} & \textbf{1.52} & \textbf{3.03} & \textbf{5.45} & 19.51 & 19.80 & \underline{20.26} & \underline{11.67} & \underline{12.14} & \underline{13.60} & 7.48 & \underline{6.04} & \underline{6.89} & 10.11 & 10.99 & 11.37 & \cellcolor{OverallShade} \textbf{9.34} \\
GOFA     & 7.10  & 16.67 & 18.36 & 13.77 & 17.39 & 21.74 & 20.54 & 24.07 & 31.49 & 14.14 & 15.35 & 16.85 & 8.65  & 9.20  & 12.23 & 10.85 & 13.46 & 13.99 & \cellcolor{OverallShade} 15.88 \\
UniGTE   & \underline{3.52} & \underline{6.53} & \underline{7.07} & \underline{4.55} & \underline{8.08} & \underline{10.61} & \textbf{3.93} & \textbf{4.75} & \textbf{6.03} & \textbf{7.62} & \textbf{8.95} & \textbf{9.42} & 16.45 & 21.00 & 26.95 & 6.29 & 8.01 & 10.49 & \cellcolor{OverallShade} \underline{9.46} \\
\bottomrule
\end{tabular}%
}
\vskip -0.175in
\end{table*}
\begin{table*}[t]
\centering
\caption{Attack success rate (ASR, \%) under textual attacks across different perturbation sizes. Lower values indicate stronger robustness. Best results are bolded and runner-up results are underlined.}
\vskip -0.125in
\label{tab:textual_asr}
\definecolor{OverallShade}{RGB}{255, 245, 220}
\providecommand{\datasetheader}[1]{\raisebox{0.4ex}{\textbf{#1}}}
\providecommand{\levelheader}[1]{\raisebox{0.3ex}{\textbf{#1}}}
\setlength{\tabcolsep}{2pt}
\scriptsize
\resizebox{\textwidth}{!}{%
\begin{tabular}{l|ccc|ccc|ccc|ccc|ccc|ccc|c}
\toprule
\multirow{2}{*}{\textbf{Method}} & \multicolumn{3}{c}{\datasetheader{Cora}} & \multicolumn{3}{c|}{\datasetheader{Citeseer}} & \multicolumn{3}{c}{\datasetheader{History}} & \multicolumn{3}{c|}{\datasetheader{Child}} & \multicolumn{3}{c}{\datasetheader{Instagram}} & \multicolumn{3}{c|}{\datasetheader{WikiCS}} & \multirow{2}{*}{\textbf{Avg.}} \\[-3pt]
\cmidrule(lr){2-4}\cmidrule(lr){5-7}\cmidrule(lr){8-10}\cmidrule(lr){11-13}\cmidrule(lr){14-16}\cmidrule(lr){17-19}
 & \levelheader{Low} & \levelheader{Mid} & \levelheader{High}
 & \levelheader{Low} & \levelheader{Mid} & \levelheader{High}
 & \levelheader{Low} & \levelheader{Mid} & \levelheader{High}
 & \levelheader{Low} & \levelheader{Mid} & \levelheader{High}
 & \levelheader{Low} & \levelheader{Mid} & \levelheader{High}
 & \levelheader{Low} & \levelheader{Mid} & \levelheader{High} & \\[-2pt]
\midrule
\multicolumn{20}{c}{{\normalsize\itshape GNNPred}} \\
OFA       & 5.67 & 13.48 & 14.18 & 7.27 & 18.34 & 40.48 & 7.03 & 10.84 & 21.69 & 11.90 & 23.81 & 30.95 & 5.48 & 11.80 & 24.17 & 10.96 & \underline{15.58} & \underline{25.00} & \cellcolor{OverallShade} 16.59 \\
G2P2      & \underline{2.02} & \underline{5.65} & \underline{9.68} & \underline{5.54} & \underline{10.50} & \underline{13.39} & 3.77 & 7.82 & \textbf{12.88} & \underline{6.28} & \textbf{10.12} & \underline{16.84} & 5.14 & 8.33 & \underline{13.08} & \textbf{4.35} & \textbf{7.79} & \textbf{14.29} & \cellcolor{OverallShade} \textbf{8.75} \\
ZeroG     & \textbf{1.77} & \textbf{4.26} & \textbf{4.96} & \textbf{3.57} & \textbf{8.33} & \textbf{12.20} & 10.28 & 18.09 & 31.32 & 6.81 & \underline{10.60} & \textbf{16.35} & 10.30 & 24.57 & 31.20 & 9.95 & 22.77 & 37.00 & \cellcolor{OverallShade} 14.68 \\
GraphCLIP & 5.56 & 12.57 & 24.27 & 5.80 & 15.08 & 34.57 & \textbf{3.01} & \textbf{5.84} & \underline{13.42} & 8.38 & 15.59 & 34.18 & \textbf{0.55} & \textbf{0.69} & \textbf{2.83} & \underline{7.21} & 16.03 & 37.28 & \cellcolor{OverallShade} \underline{13.49} \\
\midrule
\multicolumn{20}{c}{{\normalsize\itshape GTextLLM}} \\
Llama3.1-8B & 18.54 & 26.12 & 43.54 & 24.87 & 34.67 & 53.77 & 40.95 & 47.27 & 57.35 & 36.01 & 45.17 & 61.20 & 32.32 & 41.02 & 51.56 & 18.74 & 28.06 & 48.08 & \cellcolor{OverallShade} 39.40 \\
Qwen3-8B    & 15.72 & 23.04 & 42.55 & 20.48 & 33.78 & 55.59 & 34.24 & 39.45 & 51.49 & 30.04 & 39.07 & 57.12 & 25.93 & 37.32 & 49.57 & 14.52 & 24.78 & 45.00 & \cellcolor{OverallShade} 35.54 \\
Vicuna-7B   & 28.06 & 33.55 & 50.97 & 31.97 & 36.05 & 51.02 & 42.95 & 47.11 & 52.28 & 48.04 & 54.15 & 60.34 & 58.71 & 59.43 & 62.71 & 40.45 & 49.46 & 65.94 & \cellcolor{OverallShade} 48.51 \\
G1          & 19.68 & 25.61 & 43.94 & 19.06 & 29.91 & 51.32 & 32.12 & 35.30 & 46.91 & 30.82 & 37.28 & 52.04 & 32.49 & 40.42 & 55.66 & 13.06 & 21.23 & 41.18 & \cellcolor{OverallShade} 34.89 \\
\midrule
\multicolumn{20}{c}{{\normalsize\itshape GTokenLLM}} \\
GraphGPT & 32.73 & 33.64 & 61.82 & 17.31 & 29.81 & 52.40 & 25.82 & 38.89 & 61.60 & 30.00 & 41.16 & 52.42 & 10.22 & 13.04 & 20.00 & 12.45 & 22.94 & 42.23 & \cellcolor{OverallShade} 33.25 \\
LLaGA    & 23.75 & 28.75 & 46.25 & 25.00 & 31.25 & 57.14 & 23.44 & 32.50 & 47.76 & 28.86 & 35.69 & 48.57 & 31.20 & 32.85 & 37.19 & 18.54 & 24.64 & 42.32 & \cellcolor{OverallShade} 34.21 \\
TEA-GLM  & 14.02 & 23.11 & 42.42 & 15.09 & 26.63 & 52.66 & 24.28 & 32.15 & 50.70 & 15.41 & 22.37 & 35.91 & 6.17 & 12.35 & 21.18 & 17.54 & 28.21 & 49.34 & \cellcolor{OverallShade} 27.20 \\
GOFA     & 10.11 & 21.31 & 36.07 & 11.62 & 22.89 & 46.48 & 7.26 & 14.26 & 29.50 & 8.17 & 13.62 & 25.97 & 9.35 & 17.71 & 28.62 & 10.54 & 20.61 & 42.67 & \cellcolor{OverallShade} 20.93 \\
UniGTE   & 10.05 & 24.12 & 41.21 & 11.71 & 24.88 & 48.78 & \underline{3.56} & \underline{7.58} & 17.63 & \textbf{6.05} & 13.39 & 31.17 & \underline{3.32} & \underline{6.93} & 13.66 & 9.32 & 19.77 & 41.87 & \cellcolor{OverallShade} 18.61 \\
\bottomrule
\end{tabular}%
}
\vskip -0.175in
\end{table*}
\begin{table*}[t]
\centering
\caption{Attack success rate (ASR, \%) under node injection attacks across different perturbation sizes. Lower values indicate stronger robustness. Best results are bolded and runner-up results are underlined.}
\vskip -0.125in
\label{tab:hybrid_asr}
\definecolor{OverallShade}{RGB}{255, 245, 220}
\providecommand{\datasetheader}[1]{\raisebox{0.4ex}{\textbf{#1}}}
\providecommand{\levelheader}[1]{\raisebox{0.3ex}{\textbf{#1}}}
\setlength{\tabcolsep}{2pt}
\scriptsize
\resizebox{\textwidth}{!}{%
\begin{tabular}{l|ccc|ccc|ccc|ccc|ccc|ccc|c}
\toprule
\multirow{2}{*}{\textbf{Method}} & \multicolumn{3}{c}{\datasetheader{Cora}} & \multicolumn{3}{c|}{\datasetheader{Citeseer}} & \multicolumn{3}{c}{\datasetheader{History}} & \multicolumn{3}{c|}{\datasetheader{Child}} & \multicolumn{3}{c}{\datasetheader{Instagram}} & \multicolumn{3}{c|}{\datasetheader{WikiCS}} & \multirow{2}{*}{\textbf{Avg.}} \\[-3pt]
\cmidrule(lr){2-4}\cmidrule(lr){5-7}\cmidrule(lr){8-10}\cmidrule(lr){11-13}\cmidrule(lr){14-16}\cmidrule(lr){17-19}
 & \levelheader{Low} & \levelheader{Mid} & \levelheader{High}
 & \levelheader{Low} & \levelheader{Mid} & \levelheader{High}
 & \levelheader{Low} & \levelheader{Mid} & \levelheader{High}
 & \levelheader{Low} & \levelheader{Mid} & \levelheader{High}
 & \levelheader{Low} & \levelheader{Mid} & \levelheader{High}
 & \levelheader{Low} & \levelheader{Mid} & \levelheader{High} & \\[-2pt]
\midrule
\multicolumn{20}{c}{{\normalsize\itshape GNNPred}} \\
OFA       & 3.55 & 5.67 & 8.51 & 7.27 & 12.11 & 17.30 & 24.30 & 30.72 & 40.16 & 54.76 & 57.14 & 57.14 & \underline{8.50} & 14.64 & 19.26 & 17.98 & 22.26 & 27.74 & \cellcolor{OverallShade} 23.83 \\
G2P2      & 8.06 & 14.52 & 19.76 & 12.24 & 16.62 & 21.28 & 23.64 & 27.14 & 31.46 & 28.46 & 37.08 & 51.36 & 14.92 & 16.96 & 24.03 & 6.38 & 11.89 & 21.04 & \cellcolor{OverallShade} 21.49 \\
ZeroG     & 9.57 & 13.83 & 39.01 & 12.80 & 29.46 & 49.40 & 68.68 & 83.89 & 91.26 & 45.51 & 68.21 & 81.84 & 46.18 & 65.52 & 67.86 & 18.05 & 26.77 & 51.50 & \cellcolor{OverallShade} 48.30 \\
GraphCLIP & 5.85 & 10.82 & 29.53 & 5.80 & 11.37 & 15.55 & 24.32 & 33.06 & 41.97 & 26.44 & 27.83 & 30.60 & \textbf{4.14} & \textbf{6.08} & \underline{12.50} & 13.17 & 16.55 & 19.72 & \cellcolor{OverallShade} 18.63 \\
\midrule
\multicolumn{20}{c}{{\normalsize\itshape GTextLLM}} \\
Llama3.1-8B & 8.71 & 10.96 & 11.80 & 15.83 & 16.83 & 15.58 & 29.84 & 30.05 & 33.43 & 29.68 & 30.16 & 30.81 & 28.61 & 28.42 & 28.32 & 10.22 & 11.85 & 12.48 & \cellcolor{OverallShade} 21.31 \\
Qwen3-8B    & 7.32 & 7.05 & 10.30 & 14.36 & 15.16 & 13.30 & 29.13 & 31.17 & 30.59 & 25.51 & 27.11 & 27.07 & 22.78 & 26.89 & 25.55 & 6.01 & 7.83 & 10.04 & \cellcolor{OverallShade} 18.73 \\
Vicuna-7B   & 20.00 & 22.26 & 32.58 & 29.59 & 30.27 & 44.22 & 47.44 & 47.72 & 49.63 & 53.36 & 53.06 & 54.34 & 56.00 & 55.00 & 56.57 & 40.01 & 44.91 & 49.34 & \cellcolor{OverallShade} 43.68 \\
G1          & 9.97 & 9.97 & 16.17 & 14.66 & 14.66 & 19.06 & 24.73 & 23.82 & \underline{27.77} & 23.33 & 24.05 & 24.58 & 26.74 & 29.97 & 31.88 & 5.50 & 8.70 & 10.89 & \cellcolor{OverallShade} 19.25 \\
\midrule
\multicolumn{20}{c}{{\normalsize\itshape GTokenLLM}} \\
GraphGPT & 48.18 & 70.00 & 91.82 & 28.85 & 46.15 & 60.10 & 71.02 & 85.16 & 85.97 & 79.47 & 80.42 & 80.95 & 10.00 & \underline{10.00} & \textbf{10.00} & 10.00 & 13.83 & 16.34 & \cellcolor{OverallShade} 49.90 \\
LLaGA    & 28.75 & 23.75 & 27.50 & 33.04 & 33.04 & 48.21 & 32.66 & 37.55 & 42.81 & 34.29 & 37.52 & 38.88 & 59.09 & 51.65 & 52.89 & 15.20 & 13.02 & 13.10 & \cellcolor{OverallShade} 34.61 \\
TEA-GLM  & \textbf{2.65} & \textbf{2.27} & \textbf{2.27} & \textbf{3.85} & \textbf{2.66} & \textbf{3.55} & 22.99 & 27.26 & 29.16 & 10.77 & 12.44 & 14.50 & 11.99 & 18.16 & 20.67 & 6.08 & \textbf{5.49} & \textbf{6.44} & \cellcolor{OverallShade} 11.29 \\
GOFA     & \underline{3.28} & 7.92 & \underline{6.56} & 6.34 & 8.80 & \underline{9.15} & \underline{15.07} & \underline{20.54} & 27.89 & \underline{9.13} & \underline{10.98} & \underline{12.09} & 11.02 & 13.59 & 13.70 & \underline{5.10} & 6.47 & 7.07 & \cellcolor{OverallShade} \textbf{10.82} \\
UniGTE   & 3.52 & \underline{5.53} & 12.06 & \underline{4.39} & \underline{5.85} & 9.76 & \textbf{7.03} & \textbf{8.86} & \textbf{9.13} & \textbf{6.20} & \textbf{6.56} & \textbf{7.11} & 27.32 & 32.92 & 38.14 & \textbf{4.92} & \underline{5.95} & \underline{6.88} & \cellcolor{OverallShade} \underline{11.23} \\
\bottomrule
\end{tabular}%
}
\vskip -0.175in
\end{table*}


\subsection{Results against TextFooler Attack} \label{subsec:textfooler}
In ZeroGAR, we primarily use LLM-based text generation to perform textual attacks. However, the generated replacements may not always be sufficiently imperceptible. To address this concern, we additionally evaluate four word-level attack methods implemented in the TextAttack library: TextFooler, BAE, PWWS, and HotFlip. We use TextFooler as the primary method and treat the other three methods as additional baselines. TextFooler replaces selected words while preserving the original semantics, producing subtle lexical changes that are suitable for evaluating attacks with stronger imperceptibility requirements.

For all target datasets, the perturbed-node ratios are set to $\{0.1, 0.2, 0.4\}$ under low, medium, and high attack budgets, respectively. We fix the text encoder as MiniLM and attack only correctly classified test nodes selected through degree-weighted sampling. For MiniLM-based attacks, the input text is truncated to at most 256 tokens. TextAttack is executed in batch mode with a batch size of 4. The experimental results are reported in Table~\ref{tab:textfooler}. These results suggest that text-based adversarial attacks still overfit to the surrogate model and embedding type, resulting in limited attack effectiveness. To ensure consistently strong attacks and evaluate worst-case robustness, we therefore adopt the LLM-based textual attack, which generally degrades the performance of all backbones across different text encoders.


\begin{table*}[t]
\centering
\caption{Performance comparison under TextFooler attacks across different perturbation sizes. Each cell reports the ACC (top) and RDA (\%, bottom). Best scores are bolded and runner-up scores are underlined.}
\vskip -0.125in
\label{tab:textfooler}
\definecolor{OverallShade}{RGB}{255, 245, 220}
\providecommand{\green}[1]{\textcolor{green!60!black}{#1}}
\providecommand{\red}[1]{\textcolor{red}{#1}}
\providecommand{\datasetheader}[1]{\raisebox{0.4ex}{\textbf{#1}}}
\providecommand{\levelheader}[1]{\raisebox{0.3ex}{\textbf{#1}}}
\setlength{\tabcolsep}{2pt}
\scriptsize
\resizebox{\textwidth}{!}{%
\begin{tabular}{l|ccc|ccc|ccc|ccc|ccc|ccc|c}
\toprule
\multirow{2}{*}{\textbf{Method}} & \multicolumn{3}{c}{\datasetheader{Cora}} & \multicolumn{3}{c|}{\datasetheader{Citeseer}} & \multicolumn{3}{c}{\datasetheader{History}} & \multicolumn{3}{c|}{\datasetheader{Child}} & \multicolumn{3}{c}{\datasetheader{Instagram}} & \multicolumn{3}{c|}{\datasetheader{WikiCS}} & \multirow{2}{*}{\textbf{Avg.}} \\[-3pt]
\cmidrule(lr){2-4}\cmidrule(lr){5-7}\cmidrule(lr){8-10}\cmidrule(lr){11-13}\cmidrule(lr){14-16}\cmidrule(lr){17-19}
 & \levelheader{Low} & \levelheader{Mid} & \levelheader{High} & \levelheader{Low} & \levelheader{Mid} & \levelheader{High} & \levelheader{Low} & \levelheader{Mid} & \levelheader{High} & \levelheader{Low} & \levelheader{Mid} & \levelheader{High} & \levelheader{Low} & \levelheader{Mid} & \levelheader{High} & \levelheader{Low} & \levelheader{Mid} & \levelheader{High} & \\[-2pt]
\midrule
\multicolumn{20}{c}{{\normalsize\itshape GNNPred}} \\
OFA & \makecell[c]{24.72\\[-1pt]{\scriptsize \red{$\downarrow$4.96}}} & \makecell[c]{23.62\\[-1pt]{\scriptsize \red{$\downarrow$9.19}}} & \makecell[c]{23.06\\[-1pt]{\scriptsize \red{$\downarrow$11.34}}} & \makecell[c]{43.89\\[-1pt]{\scriptsize \red{$\downarrow$3.11}}} & \makecell[c]{42.48\\[-1pt]{\scriptsize \red{$\downarrow$6.23}}} & \makecell[c]{38.56\\[-1pt]{\scriptsize \red{$\downarrow$14.88}}} & \makecell[c]{6.12\\[-1pt]{\scriptsize \green{$\uparrow$2.17}}} & \makecell[c]{6.20\\[-1pt]{\scriptsize \green{$\uparrow$3.51}}} & \makecell[c]{6.27\\[-1pt]{\scriptsize \green{$\uparrow$4.67}}} & \makecell[c]{0.27\\[-1pt]{\scriptsize 0.00}} & \makecell[c]{0.29\\[-1pt]{\scriptsize \green{$\uparrow$7.41}}} & \makecell[c]{0.28\\[-1pt]{\scriptsize \green{$\uparrow$3.70}}} & \makecell[c]{46.76\\[-1pt]{\scriptsize \red{$\downarrow$3.92}}} & \makecell[c]{46.19\\[-1pt]{\scriptsize \red{$\downarrow$5.10}}} & \makecell[c]{45.84\\[-1pt]{\scriptsize \red{$\downarrow$5.81}}} & \makecell[c]{9.65\\[-1pt]{\scriptsize \red{$\downarrow$3.40}}} & \makecell[c]{9.47\\[-1pt]{\scriptsize \red{$\downarrow$5.21}}} & \makecell[c]{8.79\\[-1pt]{\scriptsize \red{$\downarrow$12.01}}} & \cellcolor{OverallShade} \makecell[c]{21.25\\[-1pt]{\scriptsize \red{$\downarrow$2.78}}} \\
G2P2 & \makecell[c]{45.39\\[-1pt]{\scriptsize \red{$\downarrow$0.81}}} & \makecell[c]{44.65\\[-1pt]{\scriptsize \red{$\downarrow$2.43}}} & \makecell[c]{43.17\\[-1pt]{\scriptsize \red{$\downarrow$5.66}}} & \makecell[c]{51.72\\[-1pt]{\scriptsize \red{$\downarrow$3.79}}} & \makecell[c]{52.51\\[-1pt]{\scriptsize \red{$\downarrow$2.33}}} & \makecell[c]{50.00\\[-1pt]{\scriptsize \red{$\downarrow$6.99}}} & \makecell[c]{12.89\\[-1pt]{\scriptsize \red{$\downarrow$1.45}}} & \makecell[c]{12.83\\[-1pt]{\scriptsize \red{$\downarrow$1.91}}} & \makecell[c]{12.63\\[-1pt]{\scriptsize \red{$\downarrow$3.44}}} & \makecell[c]{11.47\\[-1pt]{\scriptsize \red{$\downarrow$1.97}}} & \makecell[c]{11.29\\[-1pt]{\scriptsize \red{$\downarrow$3.50}}} & \makecell[c]{10.51\\[-1pt]{\scriptsize \red{$\downarrow$10.17}}} & \makecell[c]{45.57\\[-1pt]{\scriptsize \green{$\uparrow$0.20}}} & \makecell[c]{45.48\\[-1pt]{\scriptsize 0.00}} & \makecell[c]{44.78\\[-1pt]{\scriptsize \red{$\downarrow$1.54}}} & \makecell[c]{41.29\\[-1pt]{\scriptsize 0.00}} & \makecell[c]{41.06\\[-1pt]{\scriptsize \red{$\downarrow$0.56}}} & \makecell[c]{39.97\\[-1pt]{\scriptsize \red{$\downarrow$3.20}}} & \cellcolor{OverallShade} \makecell[c]{34.29\\[-1pt]{\scriptsize \red{$\downarrow$2.75}}} \\
ZeroG & \makecell[c]{51.29\\[-1pt]{\scriptsize \red{$\downarrow$1.42}}} & \makecell[c]{51.29\\[-1pt]{\scriptsize \red{$\downarrow$1.42}}} & \makecell[c]{51.11\\[-1pt]{\scriptsize \red{$\downarrow$1.77}}} & \makecell[c]{52.35\\[-1pt]{\scriptsize \red{$\downarrow$0.59}}} & \makecell[c]{53.45\\[-1pt]{\scriptsize \green{$\uparrow$1.50}}} & \makecell[c]{53.13\\[-1pt]{\scriptsize \green{$\uparrow$0.89}}} & \makecell[c]{37.25\\[-1pt]{\scriptsize \red{$\downarrow$0.88}}} & \makecell[c]{36.60\\[-1pt]{\scriptsize \red{$\downarrow$2.61}}} & \makecell[c]{36.54\\[-1pt]{\scriptsize \red{$\downarrow$2.77}}} & \makecell[c]{13.03\\[-1pt]{\scriptsize \green{$\uparrow$1.09}}} & \makecell[c]{13.14\\[-1pt]{\scriptsize \green{$\uparrow$1.94}}} & \makecell[c]{11.99\\[-1pt]{\scriptsize \red{$\downarrow$6.98}}} & \makecell[c]{56.54\\[-1pt]{\scriptsize \green{$\uparrow$0.07}}} & \makecell[c]{55.80\\[-1pt]{\scriptsize \red{$\downarrow$1.24}}} & \makecell[c]{55.84\\[-1pt]{\scriptsize \red{$\downarrow$1.17}}} & \makecell[c]{37.64\\[-1pt]{\scriptsize \green{$\uparrow$0.03}}} & \makecell[c]{37.40\\[-1pt]{\scriptsize \red{$\downarrow$0.61}}} & \makecell[c]{37.18\\[-1pt]{\scriptsize \red{$\downarrow$1.20}}} & \cellcolor{OverallShade} \makecell[c]{41.20\\[-1pt]{\scriptsize \red{$\downarrow$1.13}}} \\
GraphCLIP & \makecell[c]{60.70\\[-1pt]{\scriptsize \red{$\downarrow$3.80}}} & \makecell[c]{57.01\\[-1pt]{\scriptsize \red{$\downarrow$9.65}}} & \makecell[c]{53.69\\[-1pt]{\scriptsize \red{$\downarrow$14.91}}} & \makecell[c]{\textbf{65.05}\\[-1pt]{\scriptsize \red{$\downarrow$3.70}}} & \makecell[c]{\textbf{64.11}\\[-1pt]{\scriptsize \red{$\downarrow$5.09}}} & \makecell[c]{\textbf{61.29}\\[-1pt]{\scriptsize \red{$\downarrow$9.27}}} & \makecell[c]{\textbf{54.12}\\[-1pt]{\scriptsize \red{$\downarrow$4.80}}} & \makecell[c]{\textbf{51.14}\\[-1pt]{\scriptsize \red{$\downarrow$10.04}}} & \makecell[c]{\textbf{44.30}\\[-1pt]{\scriptsize \red{$\downarrow$22.08}}} & \makecell[c]{20.16\\[-1pt]{\scriptsize \red{$\downarrow$4.50}}} & \makecell[c]{19.00\\[-1pt]{\scriptsize \red{$\downarrow$10.00}}} & \makecell[c]{16.29\\[-1pt]{\scriptsize \red{$\downarrow$22.83}}} & \makecell[c]{\textbf{63.82}\\[-1pt]{\scriptsize \red{$\downarrow$1.54}}} & \makecell[c]{\textbf{63.86}\\[-1pt]{\scriptsize \red{$\downarrow$1.48}}} & \makecell[c]{\textbf{63.55}\\[-1pt]{\scriptsize \red{$\downarrow$1.96}}} & \makecell[c]{66.84\\[-1pt]{\scriptsize \red{$\downarrow$3.76}}} & \makecell[c]{62.70\\[-1pt]{\scriptsize \red{$\downarrow$9.72}}} & \makecell[c]{53.48\\[-1pt]{\scriptsize \red{$\downarrow$22.99}}} & \cellcolor{OverallShade} \makecell[c]{\textbf{52.28}\\[-1pt]{\scriptsize \red{$\downarrow$9.71}}} \\
\midrule
\multicolumn{20}{c}{{\normalsize\itshape GTextLLM}} \\
Llama3.1-8B & \makecell[c]{64.58\\[-1pt]{\scriptsize \red{$\downarrow$1.67}}} & \makecell[c]{\underline{65.50}\\[-1pt]{\scriptsize \red{$\downarrow$0.27}}} & \makecell[c]{\underline{64.39}\\[-1pt]{\scriptsize \red{$\downarrow$1.96}}} & \makecell[c]{\underline{59.09}\\[-1pt]{\scriptsize \red{$\downarrow$5.27}}} & \makecell[c]{\underline{58.31}\\[-1pt]{\scriptsize \red{$\downarrow$6.52}}} & \makecell[c]{55.64\\[-1pt]{\scriptsize \red{$\downarrow$10.80}}} & \makecell[c]{\underline{40.22}\\[-1pt]{\scriptsize \green{$\uparrow$0.95}}} & \makecell[c]{\underline{38.11}\\[-1pt]{\scriptsize \red{$\downarrow$4.34}}} & \makecell[c]{\underline{37.73}\\[-1pt]{\scriptsize \red{$\downarrow$5.30}}} & \makecell[c]{\textbf{31.04}\\[-1pt]{\scriptsize \red{$\downarrow$0.67}}} & \makecell[c]{\textbf{29.87}\\[-1pt]{\scriptsize \red{$\downarrow$4.42}}} & \makecell[c]{\textbf{28.74}\\[-1pt]{\scriptsize \red{$\downarrow$8.03}}} & \makecell[c]{45.04\\[-1pt]{\scriptsize \red{$\downarrow$0.20}}} & \makecell[c]{44.91\\[-1pt]{\scriptsize \red{$\downarrow$0.49}}} & \makecell[c]{44.60\\[-1pt]{\scriptsize \red{$\downarrow$1.17}}} & \makecell[c]{73.17\\[-1pt]{\scriptsize \red{$\downarrow$0.39}}} & \makecell[c]{72.67\\[-1pt]{\scriptsize \red{$\downarrow$1.08}}} & \makecell[c]{72.81\\[-1pt]{\scriptsize \red{$\downarrow$0.88}}} & \cellcolor{OverallShade} \makecell[c]{\underline{51.47}\\[-1pt]{\scriptsize \red{$\downarrow$3.11}}} \\
Qwen3-8B & \makecell[c]{\textbf{68.08}\\[-1pt]{\scriptsize 0.00}} & \makecell[c]{\textbf{67.71}\\[-1pt]{\scriptsize \red{$\downarrow$0.54}}} & \makecell[c]{\textbf{65.68}\\[-1pt]{\scriptsize \red{$\downarrow$3.53}}} & \makecell[c]{57.52\\[-1pt]{\scriptsize \red{$\downarrow$2.39}}} & \makecell[c]{56.11\\[-1pt]{\scriptsize \red{$\downarrow$4.79}}} & \makecell[c]{\underline{57.37}\\[-1pt]{\scriptsize \red{$\downarrow$2.65}}} & \makecell[c]{24.17\\[-1pt]{\scriptsize \red{$\downarrow$2.15}}} & \makecell[c]{24.14\\[-1pt]{\scriptsize \red{$\downarrow$2.27}}} & \makecell[c]{23.82\\[-1pt]{\scriptsize \red{$\downarrow$3.56}}} & \makecell[c]{\underline{29.89}\\[-1pt]{\scriptsize \red{$\downarrow$1.84}}} & \makecell[c]{\underline{29.64}\\[-1pt]{\scriptsize \red{$\downarrow$2.66}}} & \makecell[c]{\underline{27.97}\\[-1pt]{\scriptsize \red{$\downarrow$8.14}}} & \makecell[c]{46.06\\[-1pt]{\scriptsize 0.00}} & \makecell[c]{46.14\\[-1pt]{\scriptsize \green{$\uparrow$0.17}}} & \makecell[c]{46.36\\[-1pt]{\scriptsize \green{$\uparrow$0.65}}} & \makecell[c]{\textbf{76.77}\\[-1pt]{\scriptsize \red{$\downarrow$0.49}}} & \makecell[c]{\textbf{76.40}\\[-1pt]{\scriptsize \red{$\downarrow$0.97}}} & \makecell[c]{\textbf{75.63}\\[-1pt]{\scriptsize \red{$\downarrow$1.97}}} & \cellcolor{OverallShade} \makecell[c]{49.97\\[-1pt]{\scriptsize \red{$\downarrow$2.15}}} \\
Vicuna-7B & \makecell[c]{58.49\\[-1pt]{\scriptsize \green{$\uparrow$2.26}}} & \makecell[c]{59.41\\[-1pt]{\scriptsize \green{$\uparrow$3.86}}} & \makecell[c]{56.27\\[-1pt]{\scriptsize \red{$\downarrow$1.63}}} & \makecell[c]{42.63\\[-1pt]{\scriptsize \red{$\downarrow$7.49}}} & \makecell[c]{41.54\\[-1pt]{\scriptsize \red{$\downarrow$9.85}}} & \makecell[c]{42.95\\[-1pt]{\scriptsize \red{$\downarrow$6.79}}} & \makecell[c]{21.62\\[-1pt]{\scriptsize \green{$\uparrow$0.98}}} & \makecell[c]{21.67\\[-1pt]{\scriptsize \green{$\uparrow$1.21}}} & \makecell[c]{22.04\\[-1pt]{\scriptsize \green{$\uparrow$2.94}}} & \makecell[c]{17.28\\[-1pt]{\scriptsize \green{$\uparrow$0.23}}} & \makecell[c]{16.63\\[-1pt]{\scriptsize \red{$\downarrow$3.54}}} & \makecell[c]{16.01\\[-1pt]{\scriptsize \red{$\downarrow$7.13}}} & \makecell[c]{30.01\\[-1pt]{\scriptsize \red{$\downarrow$2.72}}} & \makecell[c]{31.42\\[-1pt]{\scriptsize \green{$\uparrow$1.85}}} & \makecell[c]{30.28\\[-1pt]{\scriptsize \red{$\downarrow$1.85}}} & \makecell[c]{41.32\\[-1pt]{\scriptsize \red{$\downarrow$2.85}}} & \makecell[c]{40.60\\[-1pt]{\scriptsize \red{$\downarrow$4.54}}} & \makecell[c]{39.66\\[-1pt]{\scriptsize \red{$\downarrow$6.75}}} & \cellcolor{OverallShade} \makecell[c]{34.99\\[-1pt]{\scriptsize \red{$\downarrow$1.88}}} \\
G1 & \makecell[c]{65.68\\[-1pt]{\scriptsize \red{$\downarrow$4.05}}} & \makecell[c]{64.58\\[-1pt]{\scriptsize \red{$\downarrow$5.65}}} & \makecell[c]{64.02\\[-1pt]{\scriptsize \red{$\downarrow$6.47}}} & \makecell[c]{51.57\\[-1pt]{\scriptsize \red{$\downarrow$3.52}}} & \makecell[c]{52.51\\[-1pt]{\scriptsize \red{$\downarrow$1.76}}} & \makecell[c]{50.63\\[-1pt]{\scriptsize \red{$\downarrow$5.28}}} & \makecell[c]{26.23\\[-1pt]{\scriptsize \red{$\downarrow$1.09}}} & \makecell[c]{26.03\\[-1pt]{\scriptsize \red{$\downarrow$1.85}}} & \makecell[c]{26.77\\[-1pt]{\scriptsize \green{$\uparrow$0.94}}} & \makecell[c]{25.40\\[-1pt]{\scriptsize \red{$\downarrow$2.16}}} & \makecell[c]{25.12\\[-1pt]{\scriptsize \red{$\downarrow$3.24}}} & \makecell[c]{23.85\\[-1pt]{\scriptsize \red{$\downarrow$8.13}}} & \makecell[c]{49.89\\[-1pt]{\scriptsize \red{$\downarrow$1.38}}} & \makecell[c]{48.96\\[-1pt]{\scriptsize \red{$\downarrow$3.22}}} & \makecell[c]{50.02\\[-1pt]{\scriptsize \red{$\downarrow$1.13}}} & \makecell[c]{\underline{74.91}\\[-1pt]{\scriptsize \red{$\downarrow$0.03}}} & \makecell[c]{\underline{75.01}\\[-1pt]{\scriptsize \green{$\uparrow$0.11}}} & \makecell[c]{\underline{73.95}\\[-1pt]{\scriptsize \red{$\downarrow$1.31}}} & \cellcolor{OverallShade} \makecell[c]{48.62\\[-1pt]{\scriptsize \red{$\downarrow$2.51}}} \\
\midrule
\multicolumn{20}{c}{{\normalsize\itshape GTokenLLM}} \\
GraphGPT & \makecell[c]{19.93\\[-1pt]{\scriptsize \red{$\downarrow$1.82}}} & \makecell[c]{20.30\\[-1pt]{\scriptsize 0.00}} & \makecell[c]{16.42\\[-1pt]{\scriptsize \red{$\downarrow$19.11}}} & \makecell[c]{31.19\\[-1pt]{\scriptsize \red{$\downarrow$4.33}}} & \makecell[c]{28.68\\[-1pt]{\scriptsize \red{$\downarrow$12.02}}} & \makecell[c]{25.86\\[-1pt]{\scriptsize \red{$\downarrow$20.67}}} & \makecell[c]{20.80\\[-1pt]{\scriptsize \red{$\downarrow$7.39}}} & \makecell[c]{19.34\\[-1pt]{\scriptsize \red{$\downarrow$13.89}}} & \makecell[c]{17.18\\[-1pt]{\scriptsize \red{$\downarrow$23.51}}} & \makecell[c]{5.84\\[-1pt]{\scriptsize \red{$\downarrow$5.50}}} & \makecell[c]{5.74\\[-1pt]{\scriptsize \red{$\downarrow$7.12}}} & \makecell[c]{5.44\\[-1pt]{\scriptsize \red{$\downarrow$11.97}}} & \makecell[c]{34.21\\[-1pt]{\scriptsize \red{$\downarrow$0.06}}} & \makecell[c]{34.15\\[-1pt]{\scriptsize \red{$\downarrow$0.23}}} & \makecell[c]{33.71\\[-1pt]{\scriptsize \red{$\downarrow$1.52}}} & \makecell[c]{51.00\\[-1pt]{\scriptsize \red{$\downarrow$1.54}}} & \makecell[c]{49.82\\[-1pt]{\scriptsize \red{$\downarrow$3.82}}} & \makecell[c]{46.35\\[-1pt]{\scriptsize \red{$\downarrow$10.52}}} & \cellcolor{OverallShade} \makecell[c]{25.89\\[-1pt]{\scriptsize \red{$\downarrow$9.12}}} \\
LLaGA & \makecell[c]{13.10\\[-1pt]{\scriptsize \red{$\downarrow$11.25}}} & \makecell[c]{13.10\\[-1pt]{\scriptsize \red{$\downarrow$11.25}}} & \makecell[c]{9.78\\[-1pt]{\scriptsize \red{$\downarrow$33.74}}} & \makecell[c]{17.55\\[-1pt]{\scriptsize 0.00}} & \makecell[c]{15.67\\[-1pt]{\scriptsize \red{$\downarrow$10.71}}} & \makecell[c]{12.70\\[-1pt]{\scriptsize \red{$\downarrow$27.64}}} & \makecell[c]{22.16\\[-1pt]{\scriptsize \red{$\downarrow$4.07}}} & \makecell[c]{21.44\\[-1pt]{\scriptsize \red{$\downarrow$7.19}}} & \makecell[c]{19.72\\[-1pt]{\scriptsize \red{$\downarrow$14.63}}} & \makecell[c]{15.06\\[-1pt]{\scriptsize \red{$\downarrow$4.14}}} & \makecell[c]{14.50\\[-1pt]{\scriptsize \red{$\downarrow$7.70}}} & \makecell[c]{13.16\\[-1pt]{\scriptsize \red{$\downarrow$16.23}}} & \makecell[c]{21.46\\[-1pt]{\scriptsize \green{$\uparrow$0.61}}} & \makecell[c]{20.85\\[-1pt]{\scriptsize \red{$\downarrow$2.25}}} & \makecell[c]{22.61\\[-1pt]{\scriptsize \green{$\uparrow$6.00}}} & \makecell[c]{40.77\\[-1pt]{\scriptsize \red{$\downarrow$1.78}}} & \makecell[c]{39.94\\[-1pt]{\scriptsize \red{$\downarrow$3.78}}} & \makecell[c]{37.69\\[-1pt]{\scriptsize \red{$\downarrow$9.20}}} & \cellcolor{OverallShade} \makecell[c]{20.63\\[-1pt]{\scriptsize \red{$\downarrow$9.04}}} \\
TEA-GLM & \makecell[c]{47.79\\[-1pt]{\scriptsize \red{$\downarrow$1.89}}} & \makecell[c]{45.02\\[-1pt]{\scriptsize \red{$\downarrow$7.58}}} & \makecell[c]{41.14\\[-1pt]{\scriptsize \red{$\downarrow$15.54}}} & \makecell[c]{52.35\\[-1pt]{\scriptsize \red{$\downarrow$1.19}}} & \makecell[c]{50.94\\[-1pt]{\scriptsize \red{$\downarrow$3.85}}} & \makecell[c]{45.30\\[-1pt]{\scriptsize \red{$\downarrow$14.50}}} & \makecell[c]{28.61\\[-1pt]{\scriptsize \red{$\downarrow$1.51}}} & \makecell[c]{26.50\\[-1pt]{\scriptsize \red{$\downarrow$8.78}}} & \makecell[c]{23.70\\[-1pt]{\scriptsize \red{$\downarrow$18.42}}} & \makecell[c]{19.08\\[-1pt]{\scriptsize \red{$\downarrow$2.20}}} & \makecell[c]{18.19\\[-1pt]{\scriptsize \red{$\downarrow$6.77}}} & \makecell[c]{16.85\\[-1pt]{\scriptsize \red{$\downarrow$13.63}}} & \makecell[c]{\underline{61.70}\\[-1pt]{\scriptsize \green{$\uparrow$0.50}}} & \makecell[c]{\underline{62.01}\\[-1pt]{\scriptsize \green{$\uparrow$1.01}}} & \makecell[c]{\underline{62.85}\\[-1pt]{\scriptsize \green{$\uparrow$2.38}}} & \makecell[c]{30.79\\[-1pt]{\scriptsize \red{$\downarrow$4.91}}} & \makecell[c]{29.02\\[-1pt]{\scriptsize \red{$\downarrow$10.38}}} & \makecell[c]{25.12\\[-1pt]{\scriptsize \red{$\downarrow$22.42}}} & \cellcolor{OverallShade} \makecell[c]{38.16\\[-1pt]{\scriptsize \red{$\downarrow$7.84}}} \\
GOFA & \makecell[c]{\underline{66.61}\\[-1pt]{\scriptsize \red{$\downarrow$1.36}}} & \makecell[c]{59.23\\[-1pt]{\scriptsize \red{$\downarrow$12.29}}} & \makecell[c]{58.86\\[-1pt]{\scriptsize \red{$\downarrow$12.84}}} & \makecell[c]{42.63\\[-1pt]{\scriptsize \red{$\downarrow$4.22}}} & \makecell[c]{42.48\\[-1pt]{\scriptsize \red{$\downarrow$4.56}}} & \makecell[c]{39.81\\[-1pt]{\scriptsize \red{$\downarrow$10.56}}} & \makecell[c]{27.58\\[-1pt]{\scriptsize \red{$\downarrow$2.72}}} & \makecell[c]{26.80\\[-1pt]{\scriptsize \red{$\downarrow$5.47}}} & \makecell[c]{25.10\\[-1pt]{\scriptsize \red{$\downarrow$11.46}}} & \makecell[c]{19.93\\[-1pt]{\scriptsize \red{$\downarrow$2.21}}} & \makecell[c]{19.49\\[-1pt]{\scriptsize \red{$\downarrow$4.37}}} & \makecell[c]{18.52\\[-1pt]{\scriptsize \red{$\downarrow$9.13}}} & \makecell[c]{38.78\\[-1pt]{\scriptsize \red{$\downarrow$2.02}}} & \makecell[c]{39.36\\[-1pt]{\scriptsize \red{$\downarrow$0.56}}} & \makecell[c]{38.61\\[-1pt]{\scriptsize \red{$\downarrow$2.45}}} & \makecell[c]{65.06\\[-1pt]{\scriptsize \red{$\downarrow$1.50}}} & \makecell[c]{64.14\\[-1pt]{\scriptsize \red{$\downarrow$2.89}}} & \makecell[c]{61.89\\[-1pt]{\scriptsize \red{$\downarrow$6.30}}} & \cellcolor{OverallShade} \makecell[c]{41.94\\[-1pt]{\scriptsize \red{$\downarrow$5.69}}} \\
UniGTE & \makecell[c]{35.79\\[-1pt]{\scriptsize \red{$\downarrow$2.53}}} & \makecell[c]{34.13\\[-1pt]{\scriptsize \red{$\downarrow$7.05}}} & \makecell[c]{34.32\\[-1pt]{\scriptsize \red{$\downarrow$6.54}}} & \makecell[c]{30.72\\[-1pt]{\scriptsize \red{$\downarrow$4.39}}} & \makecell[c]{30.56\\[-1pt]{\scriptsize \red{$\downarrow$4.89}}} & \makecell[c]{27.12\\[-1pt]{\scriptsize \red{$\downarrow$15.59}}} & \makecell[c]{13.05\\[-1pt]{\scriptsize \red{$\downarrow$0.99}}} & \makecell[c]{12.95\\[-1pt]{\scriptsize \red{$\downarrow$1.75}}} & \makecell[c]{12.60\\[-1pt]{\scriptsize \red{$\downarrow$4.40}}} & \makecell[c]{16.20\\[-1pt]{\scriptsize \red{$\downarrow$2.23}}} & \makecell[c]{15.71\\[-1pt]{\scriptsize \red{$\downarrow$5.19}}} & \makecell[c]{14.44\\[-1pt]{\scriptsize \red{$\downarrow$12.85}}} & \makecell[c]{45.97\\[-1pt]{\scriptsize \red{$\downarrow$1.03}}} & \makecell[c]{46.41\\[-1pt]{\scriptsize \red{$\downarrow$0.09}}} & \makecell[c]{45.92\\[-1pt]{\scriptsize \red{$\downarrow$1.14}}} & \makecell[c]{48.76\\[-1pt]{\scriptsize \red{$\downarrow$1.99}}} & \makecell[c]{47.58\\[-1pt]{\scriptsize \red{$\downarrow$4.36}}} & \makecell[c]{44.74\\[-1pt]{\scriptsize \red{$\downarrow$10.07}}} & \cellcolor{OverallShade} \makecell[c]{30.94\\[-1pt]{\scriptsize \red{$\downarrow$4.66}}} \\
\bottomrule
\end{tabular}%
}
\vskip -0.175in
\end{table*}


\end{document}